\documentclass{article}
\usepackage{spconf,amsmath,graphicx}

\usepackage{float}
\usepackage{hyperref}
\usepackage{amssymb}
\usepackage{multirow, multicol}
\usepackage{changepage}
\usepackage{svg}
\usepackage{caption}
\usepackage{rotating}
\usepackage{makecell}

\def\ours{SUPER}

\title{
SUPER: \underline{S}elfie \underline{U}ndistortion and Head \underline{P}ose \underline{E}diting with Identity P\underline{r}eservation
}
\name{Polina Karpikova, Andrei Spiridonov, Anna Vorontsova, Anastasia Yaschenko,}
\secondlinename{Ekaterina Radionova, Igor Medvedev, Alexander Limonov}
\address{Samsung Research}
\begin{document}

\twocolumn[{
\renewcommand\twocolumn[1][]{#1}
\maketitle

\begin{center}
    \centering
    \captionsetup{type=figure}
     \setlength{\tabcolsep}{0pt}
     \begin{tabular}{*{8}{@{\vspace{-1pt}}c}}
     \parbox{10pt}{\rotatebox[]{90}{Original}} &
     \makecell{\includegraphics[width=0.14\linewidth]{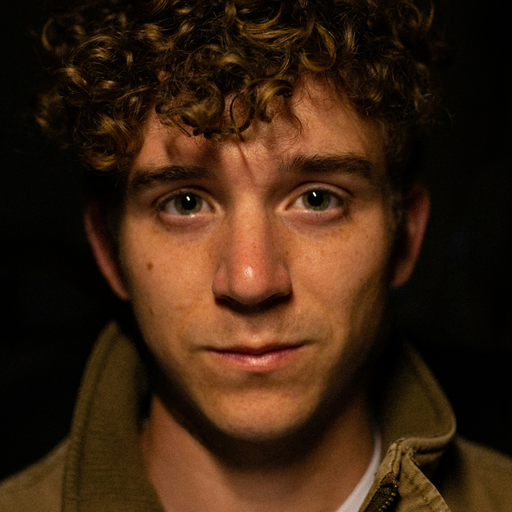}} & 
     \makecell{\includegraphics[width=0.14\linewidth]{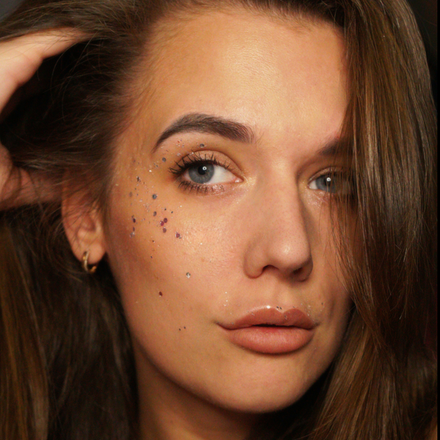}} &
     \makecell{\includegraphics[width=0.14\linewidth]{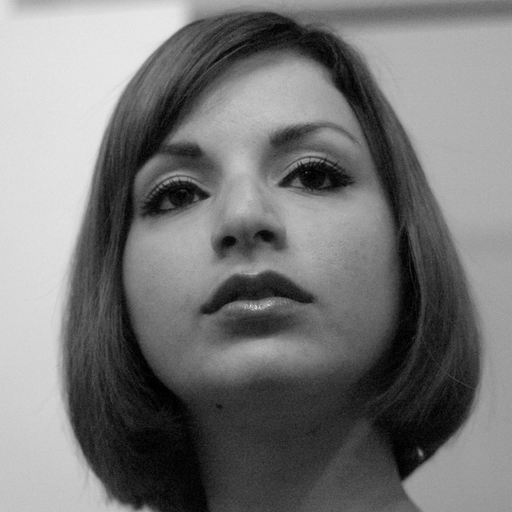}} & 
     \makecell{\includegraphics[width=0.14\linewidth]{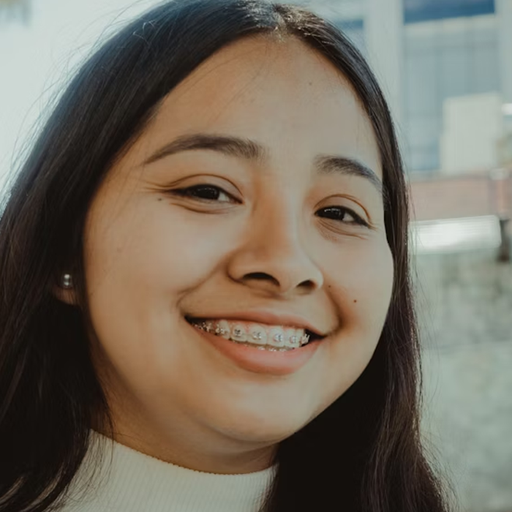}} &
     \makecell{\includegraphics[width=0.14\linewidth]{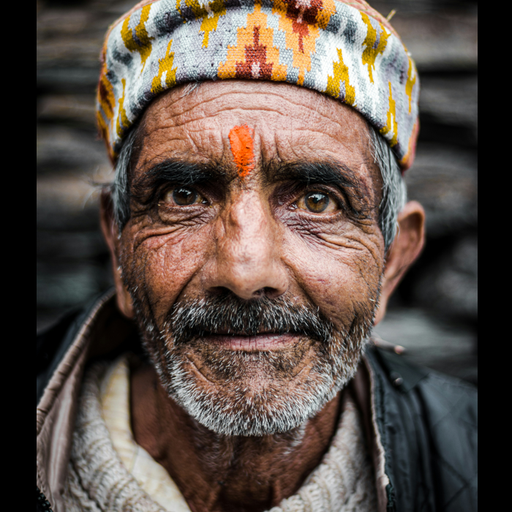}} &
     \makecell{\includegraphics[width=0.14\linewidth]{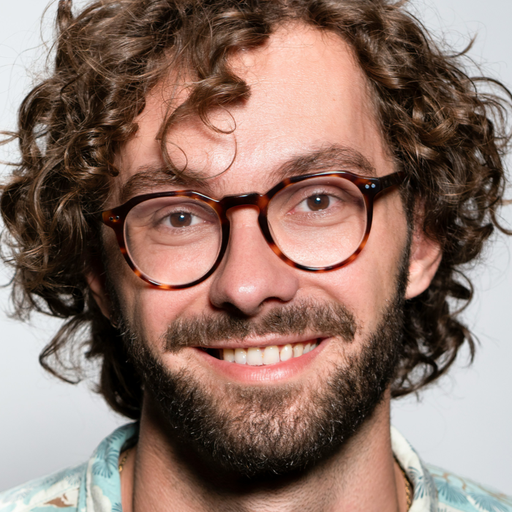}}  &
     \makecell{\includegraphics[width=0.14\linewidth]{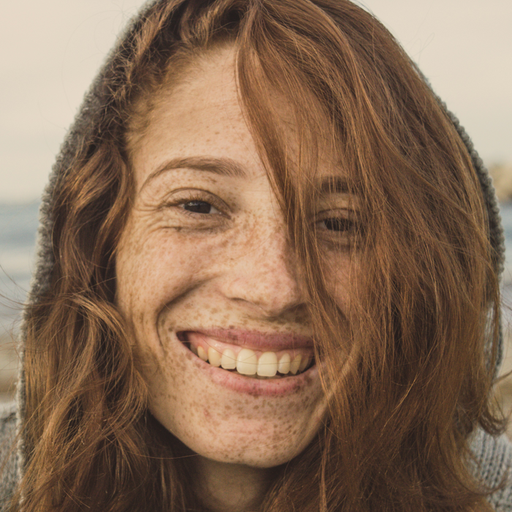}}\\
     \parbox{10pt}{\rotatebox[]{90}{Corrected}} &
     \makecell{\includegraphics[width=0.14\linewidth]{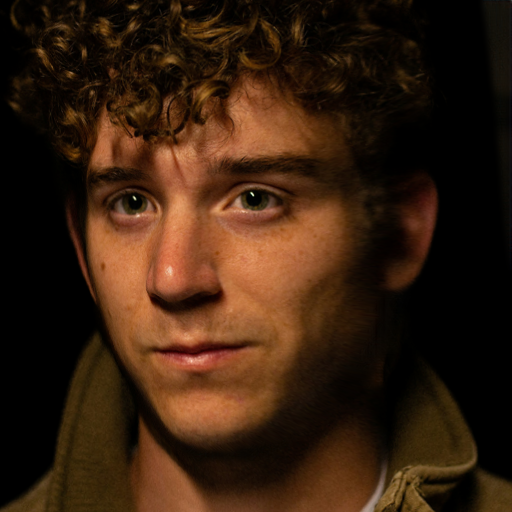}} &
     \makecell{\includegraphics[width=0.14\linewidth]{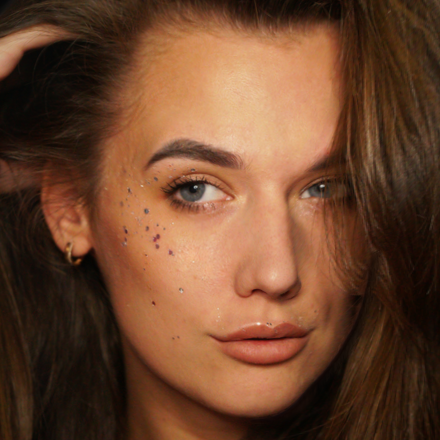}} &
     \makecell{\includegraphics[width=0.14\linewidth]{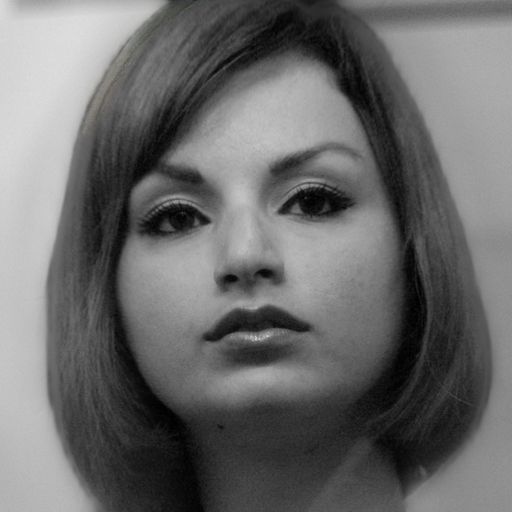}} &
     \makecell{\includegraphics[width=0.14\linewidth]{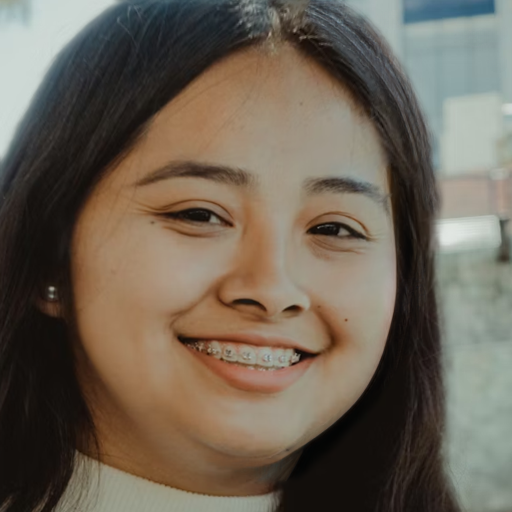}} &
     \makecell{\includegraphics[width=0.14\linewidth]{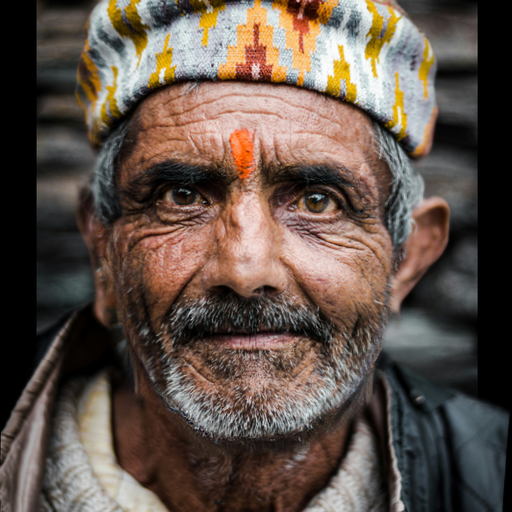}} &
     \makecell{\includegraphics[width=0.14\linewidth]{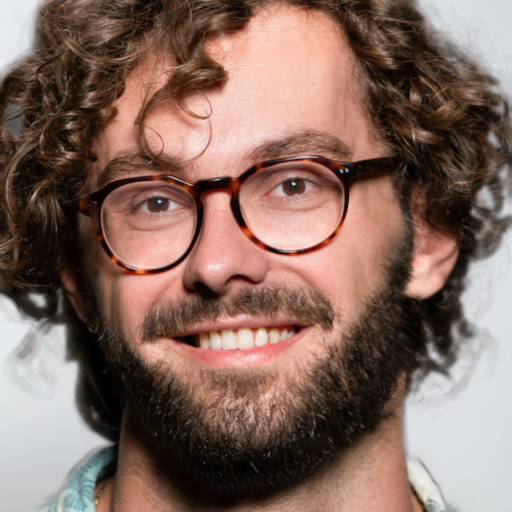}} &
     \makecell{\includegraphics[width=0.14\linewidth]{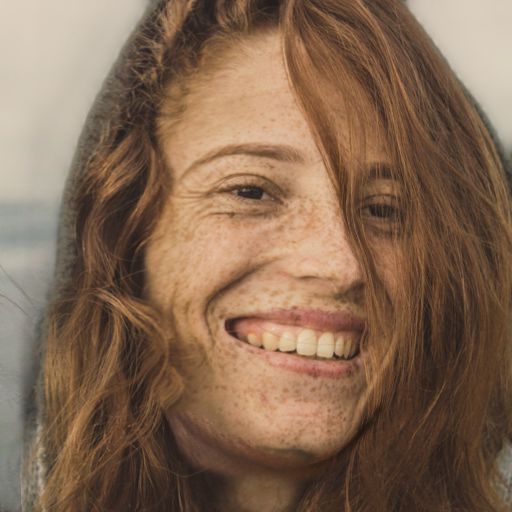}}\\
     \end{tabular}
    \captionof{figure}{A \ours{} selfie editing example. Given a selfie, we can modify a head pose and remove perspective distortion seamlessly, obtaining photorealistic and detailed corrected portraits.}
\end{center}
}]

\begin{abstract}

Self-portraits captured from a short distance might look unnatural or even unattractive due to heavy distortions making facial features malformed, and ill-placed head poses. 
In this paper, we propose \ours{}, a novel method of eliminating distortions and adjusting head pose in a close-up face crop.
We perform 3D GAN inversion for a facial image by optimizing camera parameters and face latent code, which gives a generated image. Besides, we estimate depth from the obtained latent code, create a depth-induced 3D mesh, and render it with updated camera parameters to obtain a warped portrait. Finally, we apply the visibility-based blending so that visible regions are reprojected, and occluded parts are restored with a generative model. Experiments on face undistortion benchmarks and on our self-collected Head Rotation dataset (HeRo), show that \ours{} outperforms previous approaches both qualitatively and quantitatively, opening new possibilities for photorealistic selfie editing.

\end{abstract}

\section{Introduction}
\label{sec:intro}

Selfies are arguably the most common type of photos taken with a smartphone providing high-quality images, yet they experience unwanted issues with facial geometry. Photos taken from close distances often suffer from perspective distortion, which reveals itself in malformed and asymmetrical traits such as an enormous nose and tiny or even hidden ears, providing unshaped images. Another 3D geometry-related issue is a misplaced head pose. Posing for a selfie requires practice, and selecting a proper viewpoint is non-trivial, requiring multiple attempts to adjust or even impossible in some scenarios. 
So the ability to modify the head pose during the post-processing is a demanded feature in selfie editing.

Various approaches to face geometry manipulation have been proposed so far, either based on warping, neural radiance fields, or generative models to synthesize a portrait with updated viewing conditions. Existing efforts automatically correct portrait perspective distortions often involving reconstruction-based warping \cite{fried2016perspective} and learning-based warping \cite{Zhao2019LearningPU}. Such methods perfectly preserve the details of the original image, yet, struggle with severe distortions due to using a 2D flow map to warp an image, and they are also unable to fill in occluded regions that arise inevitably. 

Another branch of face-image manipulation techniques relies on the use of generative models. In a generative pipeline, the original face image is encoded into a latent representation and further restored under novel viewing conditions and updated camera parameters with a pretrained 3D-aware GAN. The facial latent code, camera pose, and focal distance are estimated throughout a joint optimization procedure. However, fitting these parameters from a single distorted image is a challenging and ill-posed task, so existing methods struggle to restore the 3D geometry accurately. While this can be improved to a certain scale by imposing geometry constraints, there is an immanent major drawback: generative approaches cannot guarantee that the identity of a person is preserved. Besides, GANs tend to miss fine details, which severely affects the image quality. 

\ours{} combines the strengths of generative and warping paradigms, and leverages the power of a generative model while taking advantage of a 3D-based warping approach. Specifically, occluded regions are taken from a generated image, which allows restoring invisible parts of a face and head. Meanwhile, the visible part of the face is reprojected rather than generated, so the identity is preserved.

Overall, our contribution is a novel approach for portrait distortion and head-pose correction using perspective-aware 3D GAN inversion. It integrates generative and geometry-based techniques, offering a robust solution for generating high-quality, identity-preserving images from novel views in the selfie context. We prove the superior performance of \ours{} on face undistortion benchmarks and our Head Rotation dataset, HeRo, containing a number of identities with diverse head poses and camera-to-face distances.

\section{RELATED WORK}
\label{sec:related-work}

\subsection{Novel-View Synthesis of Faces}

Various approaches to face geometry manipulation have been proposed so far, either based on warping, neural radiance fields, or generative models to synthesize a portrait with updated viewing conditions. When changing a head pose, disocclusions are inevitable and more prominent, since yet invisible parts of the human head and face get revealed. Thus, warping-based methods are insufficient for head pose correction if applied solely, and trainable 3D-aware methods dominate the field. 


To restore a face under novel viewing conditions, the original face should first be mapped into the latent space of a pre-trained GAN. This technique is referred to as a GAN inversion. 2D GAN inversion methods~\cite{roich2022pivotal} do not maintain the facial structure during the transformation, so the multi-view consistency is not guaranteed. Recently introduced 3D GANs~\cite{Chan2021EfficientG3} demonstrated their capability of generating consistent outputs based on implicit 3D representations. 3D GAN inversion methods ~\cite{ide3d} leverage 2D GAN inversion methods with pre-defined camera parameters, which can be estimated with another algorithm and then fixed~\cite{yin20223d} or further tuned along with the face latent code optimization~\cite{ko20223d}. Rather than by optimization, GAN inversion can be performed with encoder-based techniques~\cite{yuan2023make}, which transform an input image into the latent space with a single forward pass and are orders of magnitude faster. In \cite{Xie2022Highfidelity3G}, training a GAN is supervised with images obtained by warping, making synthesized images more realistic. Hence, the output is entirely GAN-generated, while warped images are used only as a guidance. In our pipeline, we directly blend warped and generated images to produce a final result. 

\subsection{Portrait Perspective Undistortion}

\begin{figure*}[h!]
  \centering
  \centerline{\includegraphics[width=1.\linewidth]{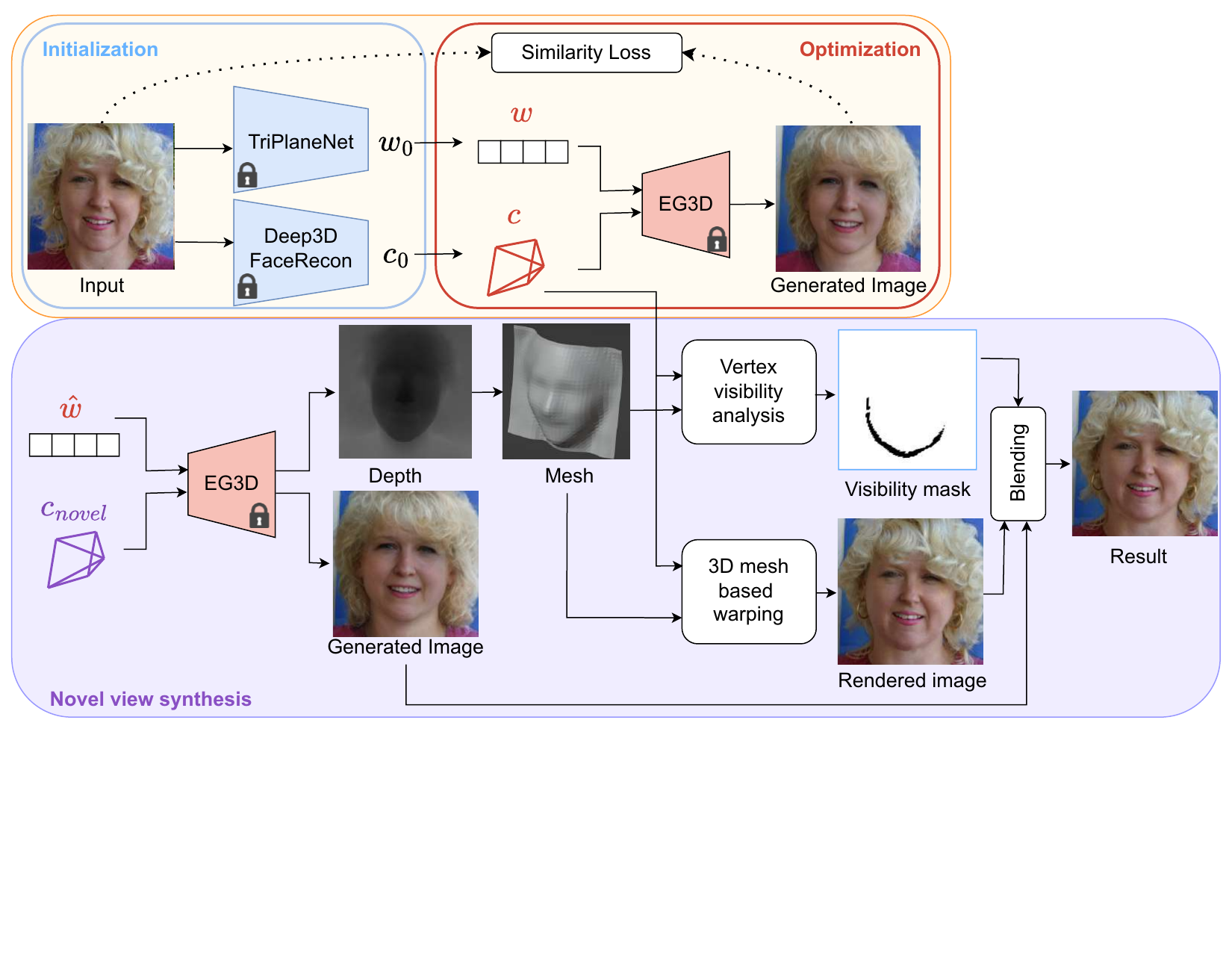}}
  \vskip -1.2in
    \boldmath
    \caption{An overview of our pipeline. We leverage a TriPlaneNet encoder to obtain an initial face latent code $w_0$, and Deep3DFaceRecon to estimate initial camera parameters $c_0$. Then, we perform an 3D GAN inversion for an input facial image by optimizing the camera parameters $c$ and a face latent code $w$. The optimized face latent code $\hat{w}$ and novel camera parameters $c_{novel}$ are passed into the EG3D~\cite{Chan2021EfficientG3} model, that generates an image and estimates a depth map. Afterward, we create a depth-induced 3D mesh, and render this mesh to obtain a warped portrait. The final novel view is synthesized by visibility-based blending, so that visible regions are reprojected, and occluded parts are restored with a generative model.
    }
    \label{fig:method}
\end{figure*}

Undistortion task arising in close-up photography is a specific, narrow case of novel view synthesis, which implies that only camera parameters are modified, while the head pose remains unchanged. Accordingly, apart from general-purpose novel view synthesis approaches, portrait undistortion is addressed with task-specific techniques.

Existing efforts to automatically correct portrait perspective distortions rely on reconstruction-based warping \cite{fried2016perspective} and learning-based warping \cite{Zhao2019LearningPU}. Such methods perfectly preserve the details of the original image, yet, struggle with severe distortions due to using a 2D flow map to warp an image, and they are also unable to fill in occluded regions that arise inevitably. 3DP~\cite{shih20203d} can render novel views from a single RGB-D image, although it can manipulate the body and somewhat mitigate face distortion by using the depth from 3D GAN, the face is still distorted.
 
Recent DisCO~\cite{wang2023disco} introduced a sophisticated 3D inversion scheme for optimizing the face latent code and camera parameters, which includes initializing with a short camera-to-face distance and reparametrizing the camera. The results, achieved with a multi-stage optimization schedule with landmarks and geometric regularization, look promising. Yet, an optimization-based GAN inversion is slow, and as a purely generative approach, DisCO does not preserve identity and outputs images lacking fine details.

\section{PROPOSED METHOD}
\label{sec:method}

The overview of the proposed method is shown in Fig.~\ref{fig:method}. We directly transfer the textures from visible parts of the face by 3D warping. To restore hidden areas, we apply the pretrained 3D GAN $G$ for generating realistic, 3D-consistent images, which can then be composed with warped images rendered from a depth-induced mesh, thereby producing a complete novel view.

Here, we focus on the cropped face regions; yet, \ours{} could easily be extended to handle full-frame images, e.g., by following the steps described in~\cite{wang2023disco}. Faces are detected with \cite{bazarevsky2019blazeface} and masked using a MODNet \cite{MODNet}.

\subsection{Generative Pipeline}
\label{ssec:generative}

\subsubsection{Initialization}
\label{sssec:initialization}

To render novel views with a 3D GAN, we need to obtain the face latent code and the original camera parameters given a target image $x$. We approximate the initial camera parameters $c_0$, namely rotation, translation and focal length, with Deep3DFaceRecon \cite{Deep3DFaceRecon}. Following DisCo \cite{wang2023disco}, we set the initial camera translation $t_{z0}$ to be reasonably small (we divide initial value, obtained for each image individually, by 2 in our experiments), and re-parameterize the focal length $f$ to $t_z$ using the estimated depth of eyes $d_0$ and ensuring that it remains unchanged:
$f = \alpha f_0$, 
where $\alpha = (d_0 - (t_{z0} - t_z )) / d_0 $.

Instead of an exhaustive optimization-based GAN inversion exploited by DisCo~\cite{wang2023disco}, we tackle inversion problem~\ref{gan-obj} with an encoder network $E$, that maps a real image into a latent code $w_0$. 
We opt for a camera-conditioned TriPlaneNet encoder \cite{bhattarai2024triplanenet}, as it is able to separate geometry from camera effects, which is crucial in our case. 

\subsubsection{Optimization}
\label{sssec:optimization}

After the initialization, the face latent code and camera parameters are jointly optimized by minimizing loss between the generated and target images:
\begin{equation}
    \hat{\textbf{w}}, \hat{\textbf{c}} = \operatorname*{argmin}_{\textbf{w}, \textbf{c}}\mathcal{L}(x, G(\textbf{w}, \textbf{c}; \theta))
    \label{gan-obj}
\end{equation}
where $G(\textbf{w}, \textbf{c}; \theta)$ is an image produced by a pre-trained generator $G$, parameterized with weights $\theta$ over the latent $\textbf{w}$ and camera extrinsic and intrinsic parameters $\textbf{c}$. 

\begin{figure*}[t!]
  \centering
  \begin{tabular}{*{9}{@{\hspace{1pt}}c}}
    \multirow{2}{*}[1.48cm]{\includegraphics[width=0.151\linewidth]{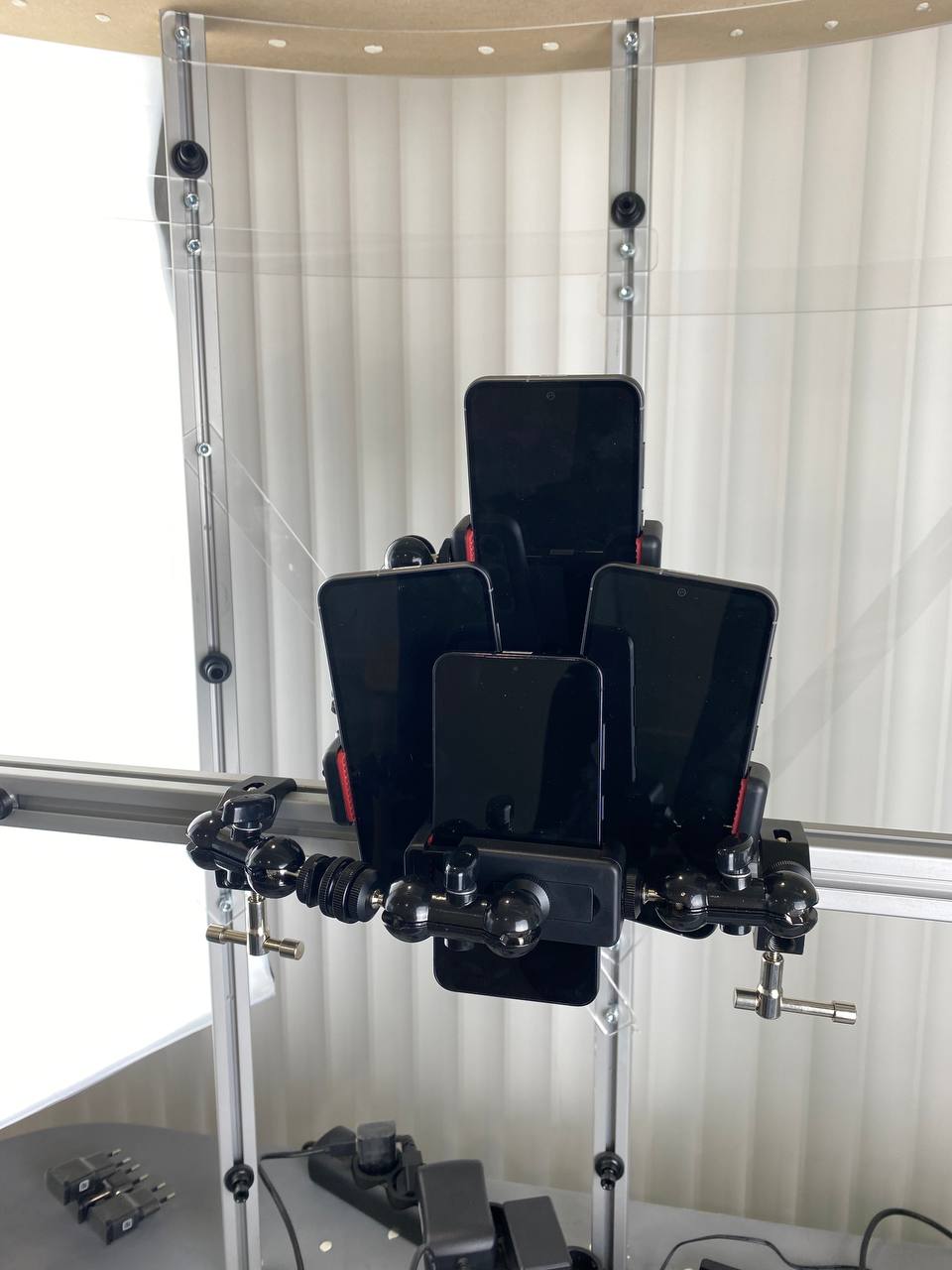}}
    \vspace{-0.1cm}
    & 
    \includegraphics[width=0.1\linewidth]{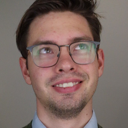}&
    \includegraphics[width=0.1\linewidth]{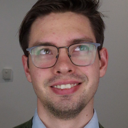}&
    \includegraphics[width=0.1\linewidth]{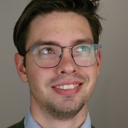}&
    \includegraphics[width=0.1\linewidth]{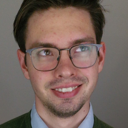}&
    \includegraphics[width=0.1\linewidth]{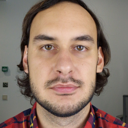}&
    \includegraphics[width=0.1\linewidth]{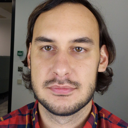}&
    \includegraphics[width=0.1\linewidth]{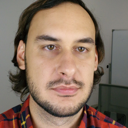}&
    \includegraphics[width=0.1\linewidth]{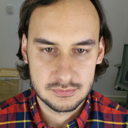}\\
    & 
    \includegraphics[width=0.1\linewidth]{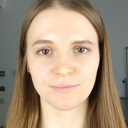}&
    \includegraphics[width=0.1\linewidth]{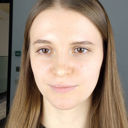}&
    \includegraphics[width=0.1\linewidth]{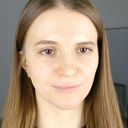}&
    \includegraphics[width=0.1\linewidth]{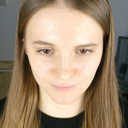}&
    \includegraphics[width=0.1\linewidth]{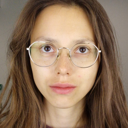}&
    \includegraphics[width=0.1\linewidth]{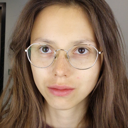}&
    \includegraphics[width=0.1\linewidth]{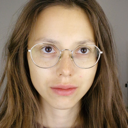}&
    \includegraphics[width=0.1\linewidth]{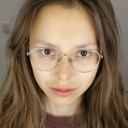}\\
    \vspace{-0.1cm}
    Capturing setup & Front & Left & Right & Top & Front & Left & Right & Top \\
  \end{tabular}
  \caption{First to the left: our HeRo capturing setup with smartphones assembled on a rig. Other: series of photos of the same individuals, simultaneously captured by Front, Left, Right, and Top cameras.}
  \label{fig:capturing-setup}
\end{figure*}

The objective is a combination of the LPIPS~\cite{zhang2018perceptual} and face landmark losses:
\begin{equation}
\begin{split}
    L(x, G(\textbf{w},\textbf{c}; \theta)) = \alpha_1 L_{LPIPS}(x, G(\textbf{w}, \textbf{c}; \theta)) + \\
    \alpha_2 L_{landmark}(f(x), f(G(\textbf{w}, \textbf{c}; \theta)))
    \label{losses-eq}
\end{split}
\end{equation}
where $f$ denotes MediaPipe FaceMesh-V2 landmark estimation model \cite{mediapipe}. The losses are equally weighted with $\alpha_1=1, \alpha_2=1$. The landmark loss is a conventional $L_2$ loss:
\begin{equation}
    L_{landmark}(m, \hat{m}) = \sum_{i=1}^{\|M\|} \|m_i - \hat{m}_i\|^2_2
    \label{keypoints}
\end{equation}
where $m$ and $\hat{m}$ are normalized 3D keypoints, and $\|M\|$ is a number of landmark points (in our case, $\|M\|=478$). Unlike DisCo~\cite{wang2023disco}, we do not impose additional constraints based on keypoint uncertainty. The latent code and camera parameters are optimized with a learning rate of 0.001 over 200 iterations. 

During the optimization, we alternate computing gradients w.r.t. a face latent code and camera. 
Besides, since the optimization starts with a reasonable initial approximation, fewer steps are required to obtain the same quality. As shown empirically, the difference in the number of iterations is up to 6 times in favor of our method, leading to a respective speed-up of the inference.

\subsection{Novel View Synthesis}
\label{ssec:warping}

To obtain a novel view, we pass the optimized face latent code $\hat{w}$ and novel camera parameters $c_{novel}$ to EG3D~\cite{Chan2021EfficientG3}, that outputs both a generated image and a dense depth map.

\subsubsection{3D warping pipeline}
\label{sssec:warping}

The dense depth map predicted by EG3D~\cite{Chan2021EfficientG3} is back-projected onto a 3D range grid. By connecting neighboring vertices in the grid, we obtain a coarse mesh. It is further refined through bilateral blur smoothing in order to avoid any unnatural sharp angles appearing in a rendered image. We use a kernel size of 5, $\sigma_{color}=0.1, \sigma_{space}=1$. Finally, we project the mesh using an estimated original camera pose to get texture coordinates, and resample the texture to get a novel view. 

\subsubsection{Blending}
\label{sssec:composing}

A depth map in the 3D warping pipeline is predicted with the same EG3D~\cite{Chan2021EfficientG3} model used to generate an image in the generative pipeline. Accordingly, a warped image obtained by rendering the depth-induced mesh, and a generated image are well-aligned by design, which allows composing them with minimal effort. Particularly, we first compute the visibility for each vertex in the mesh using the z-buffer of a rasterization algorithm. Secondly, we filter out mesh faces that are almost orthogonal (comprise an angle exceeding $80 ^{\circ}$) with the viewing direction. 

To mitigate possible inconsistencies along the face mask boundary, we dilate the mask with blur kernels, which gives a composition mask. Finally, we blend the image obtained through warping with an image generated by a neural network, using a blurred mask obtained at the previous step. Blending is done with a three-level Laplacian Pyramid.

\begin{figure*}[h!]
  \centering
  \begin{tabular}{*{8}{@{\hspace{2pt}}c}}
    Input & Fried's~\cite{fried2016perspective} & 3DP~\cite{shih20203d} & HFGi3D\cite{Xie2022Highfidelity3G} & \small{TriPlaneNet\cite{bhattarai2024triplanenet}} & DisCO~\cite{wang2023disco} & \textbf{\ours} & Reference \\
    \vspace{-0.07cm}
    \includegraphics[width=0.105\linewidth]{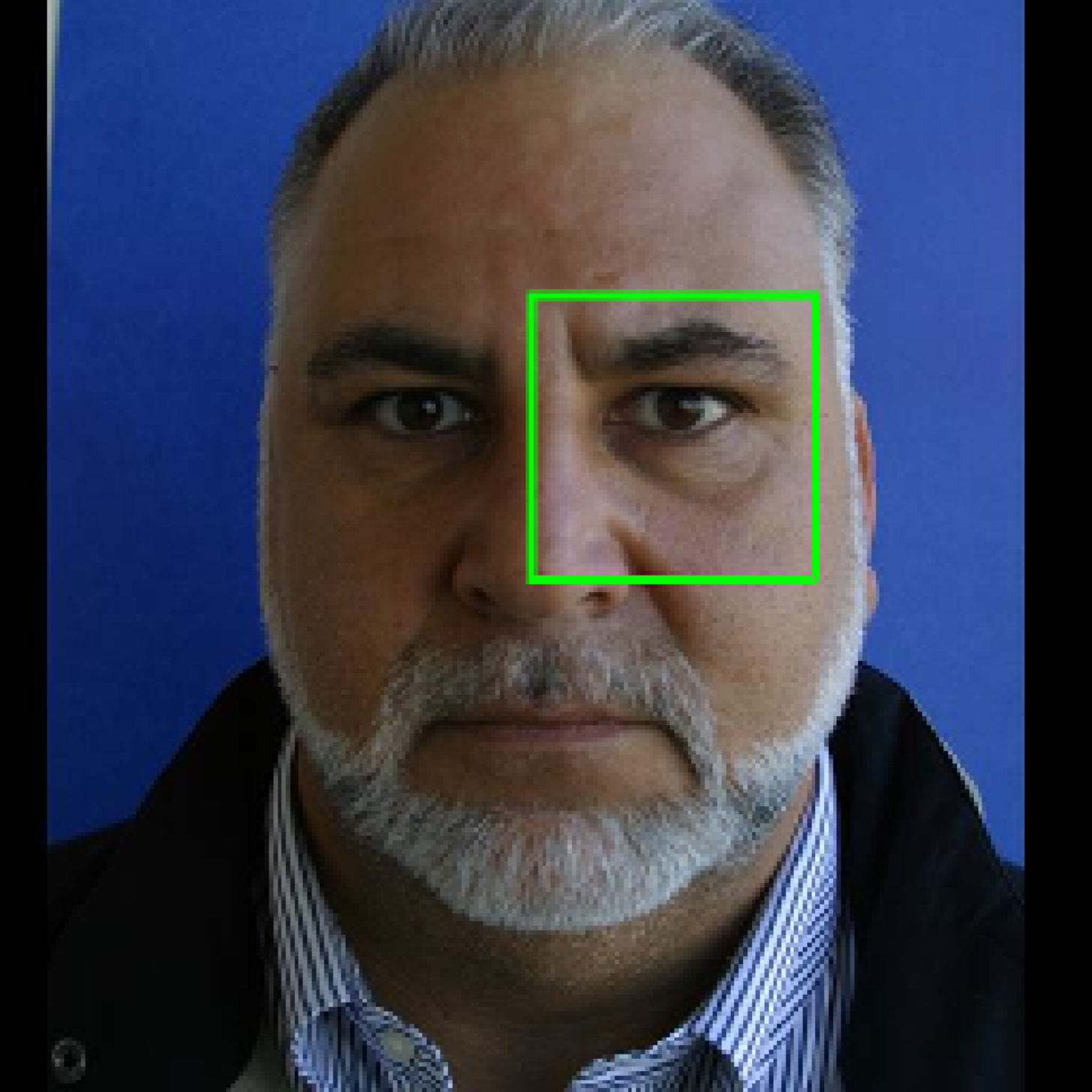}&
    \includegraphics[width=0.105\linewidth]{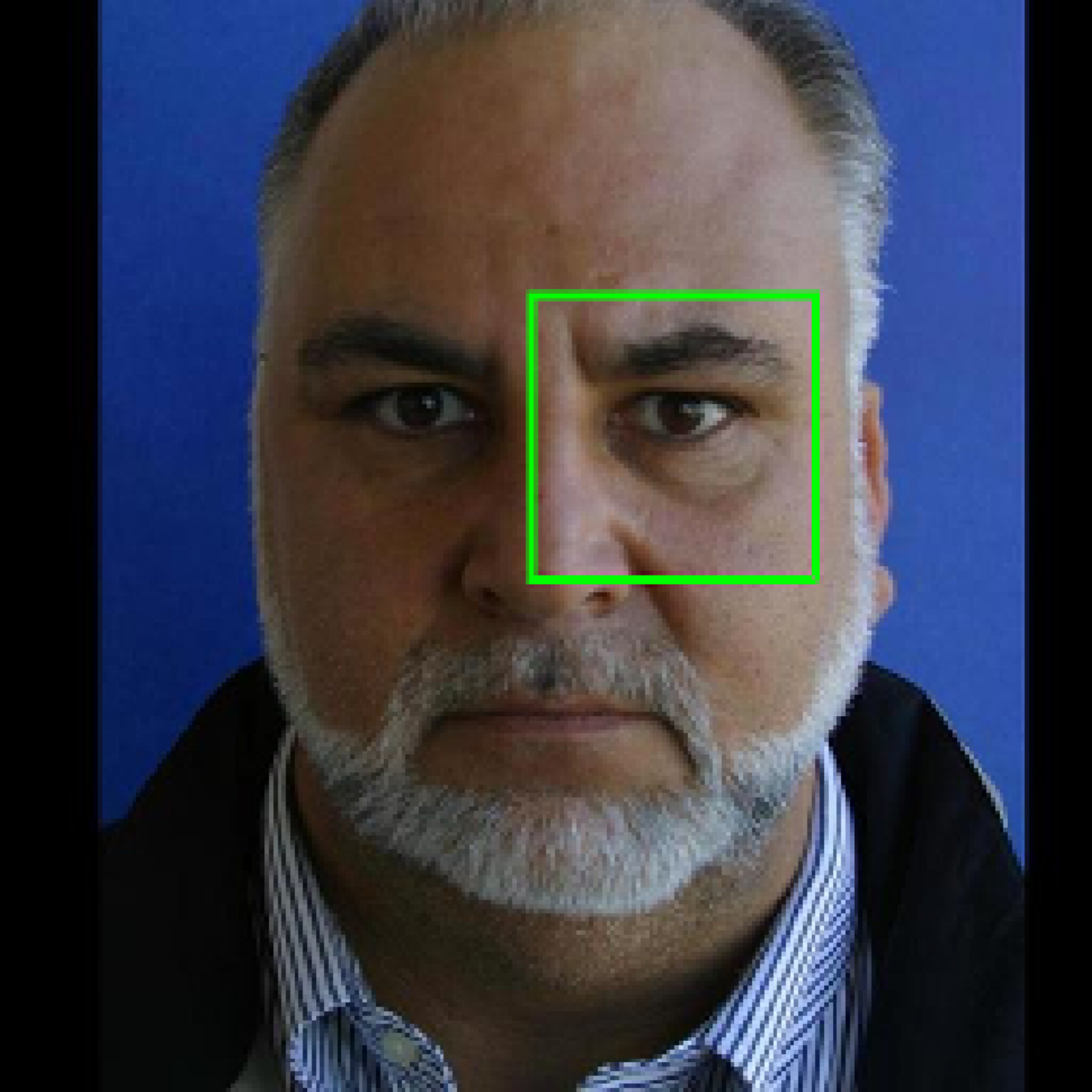}&
    \includegraphics[width=0.105\linewidth]{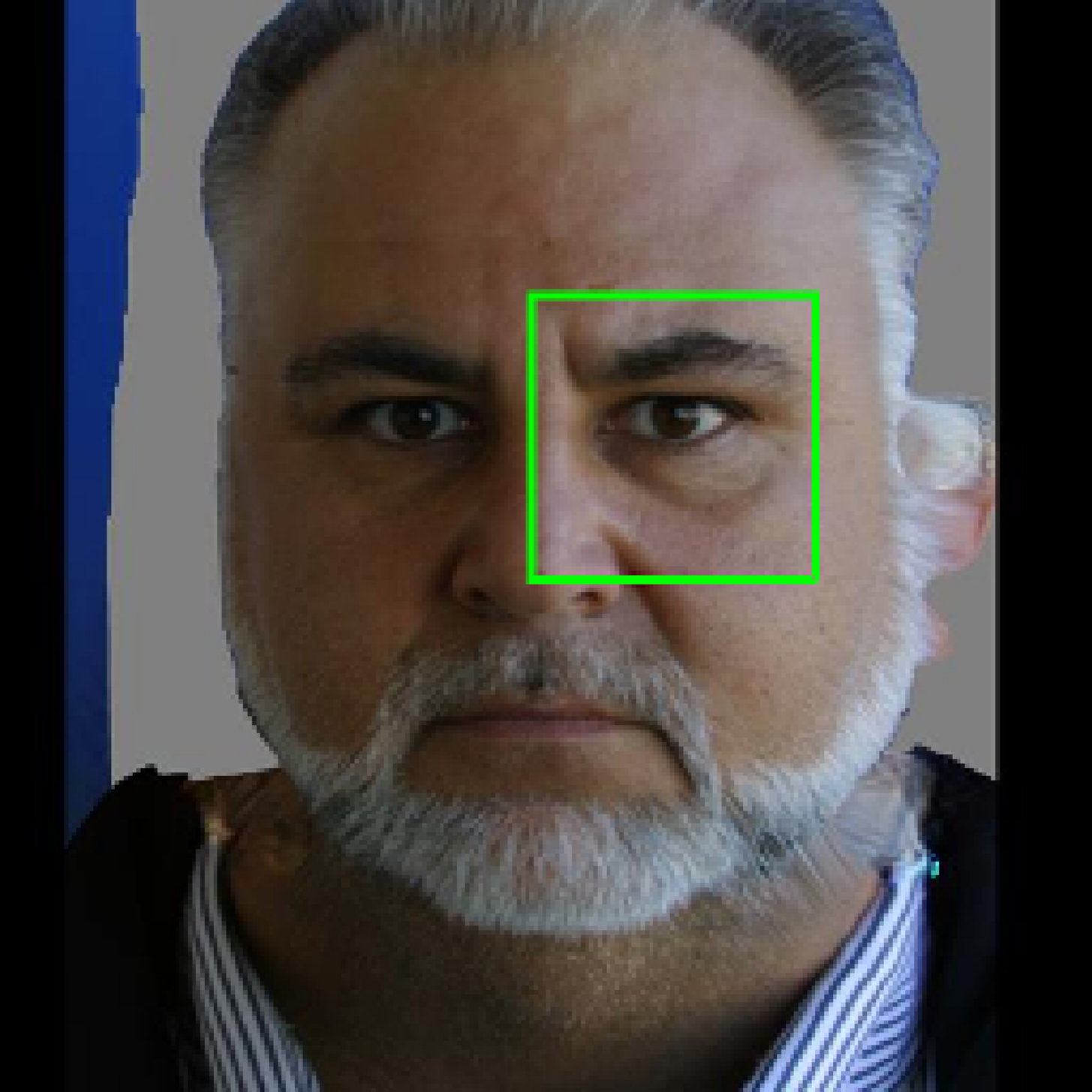}&
    \includegraphics[width=0.105\linewidth]{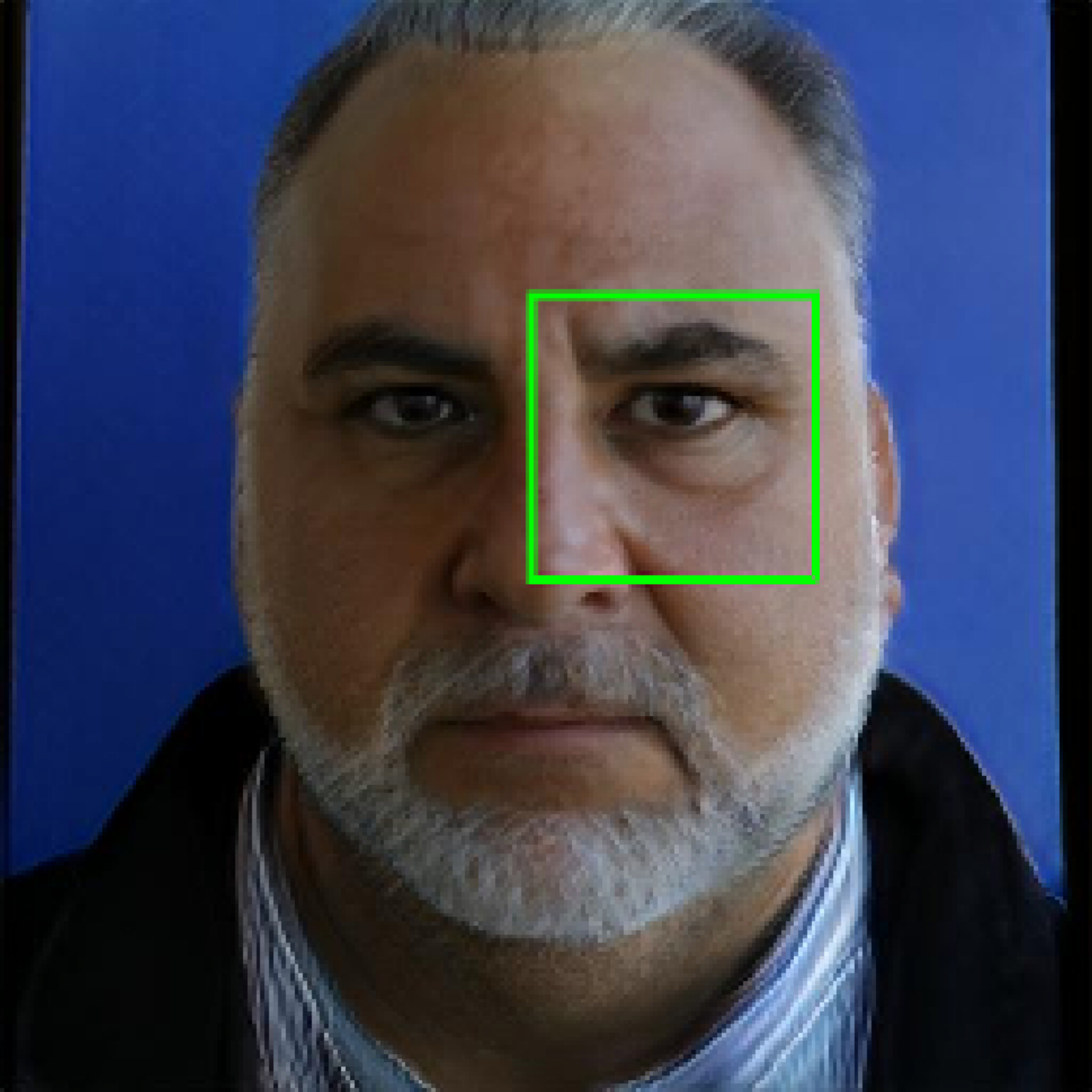}&
    \includegraphics[width=0.105\linewidth]{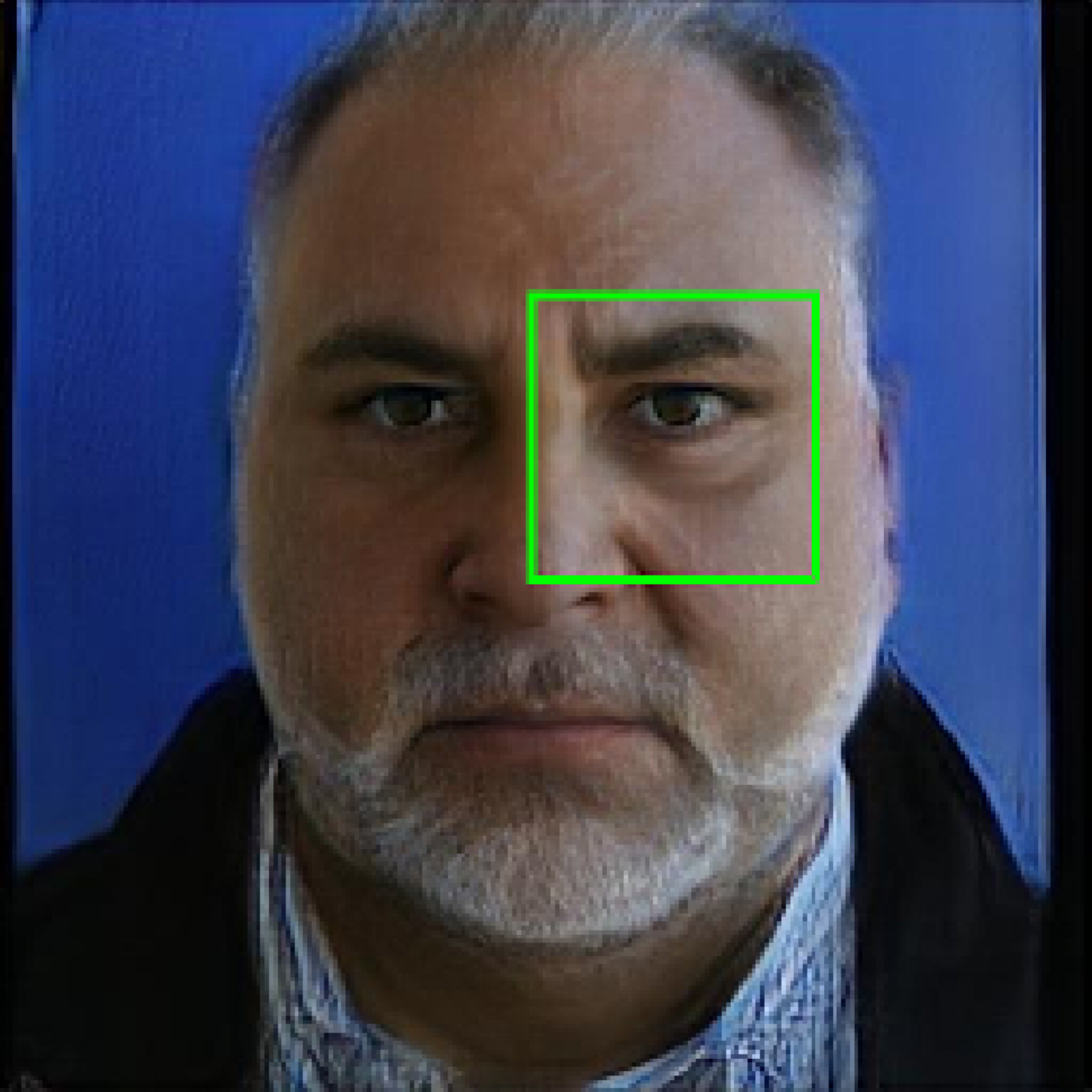}&
    \includegraphics[width=0.105\linewidth]{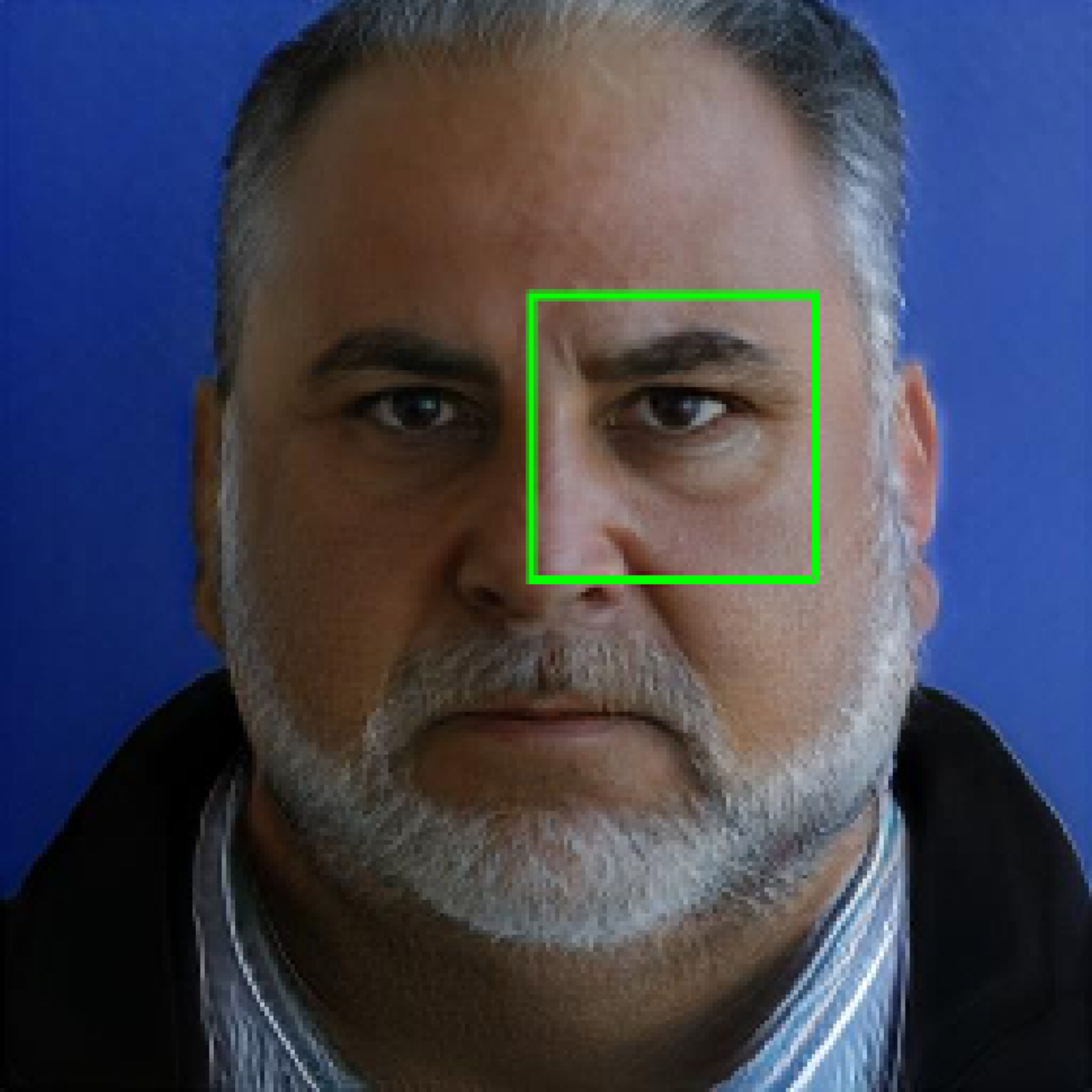}&
    \includegraphics[width=0.105\linewidth]{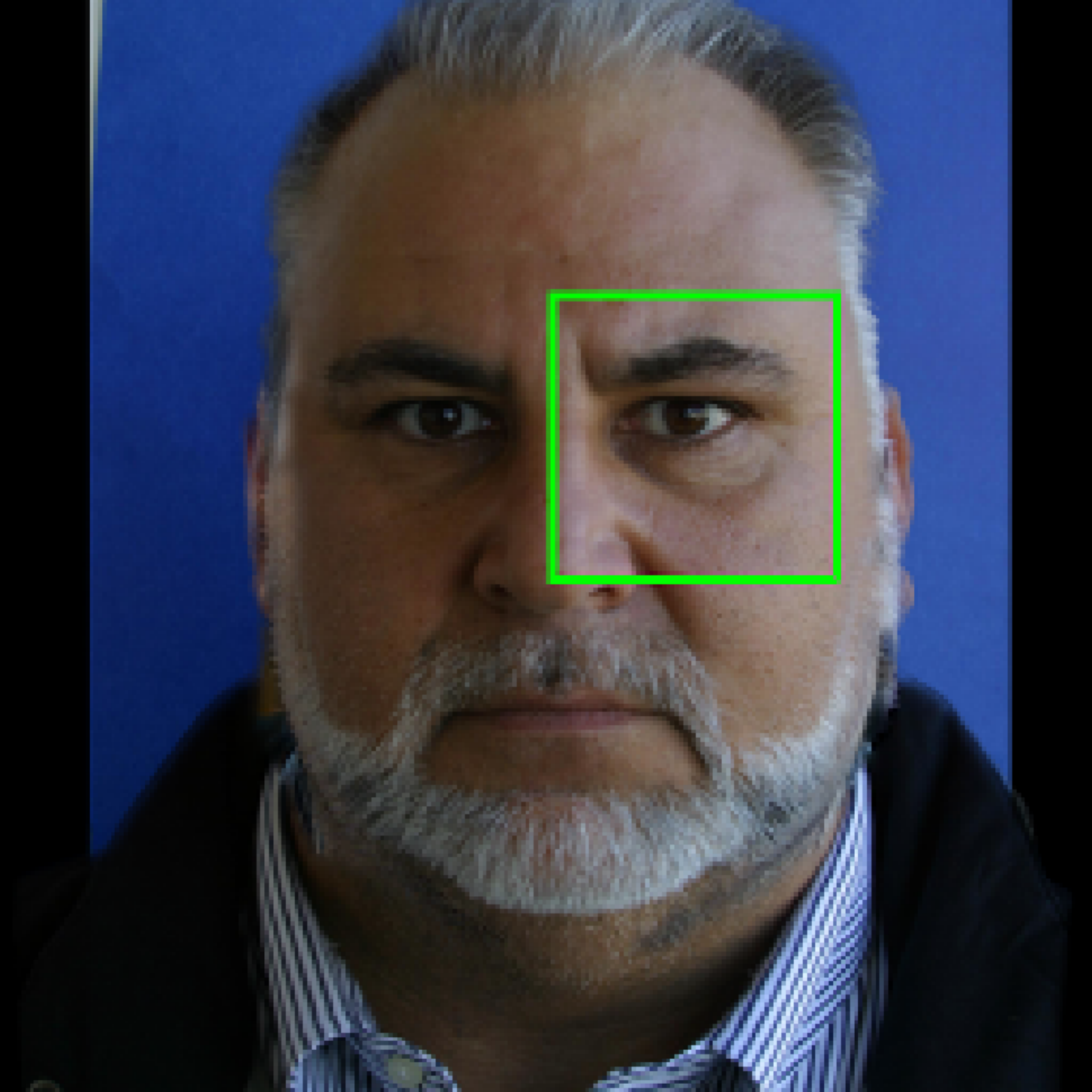}&
    \includegraphics[width=0.105\linewidth]{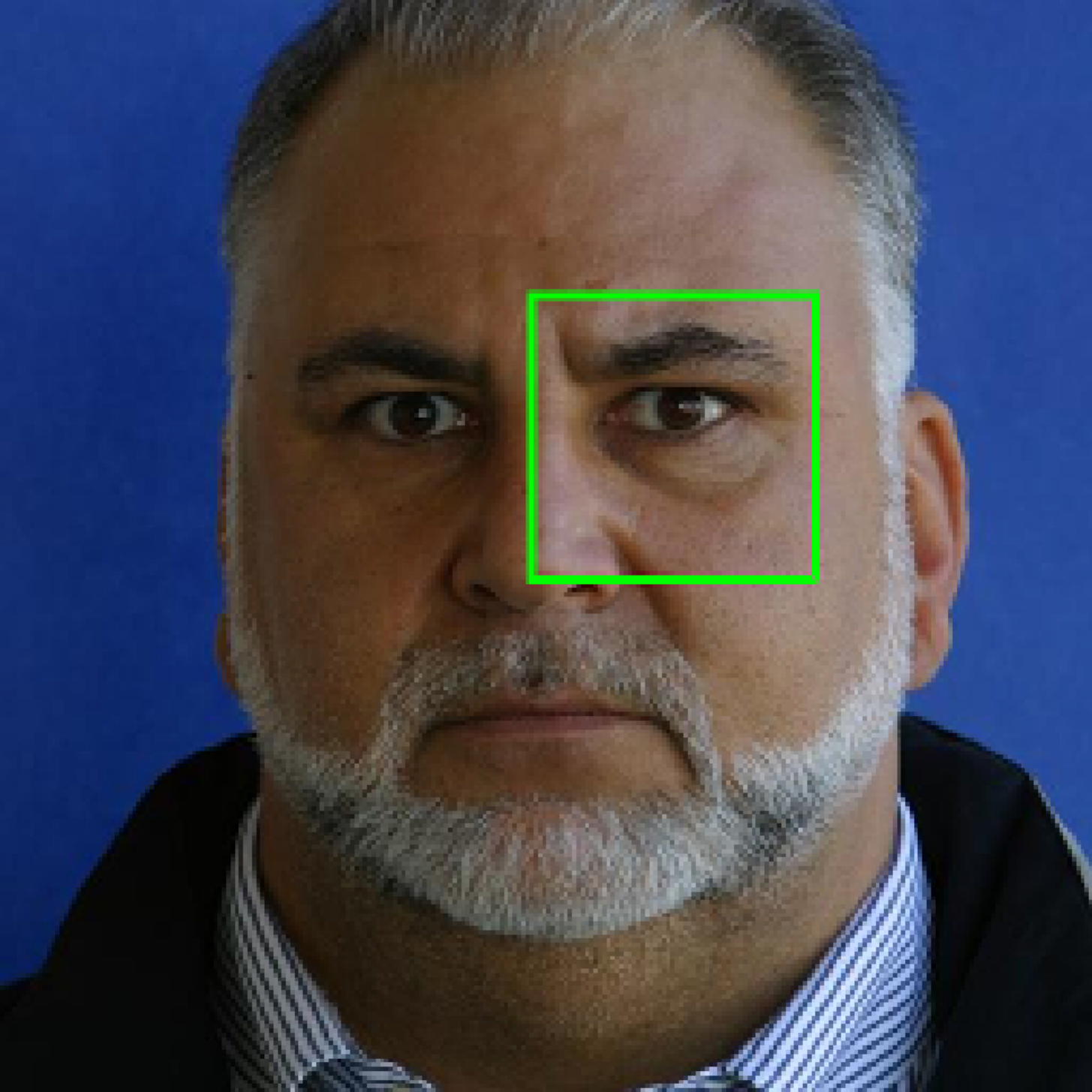}\\
    \vspace{-0.07cm}
    \includegraphics[width=0.105\linewidth]{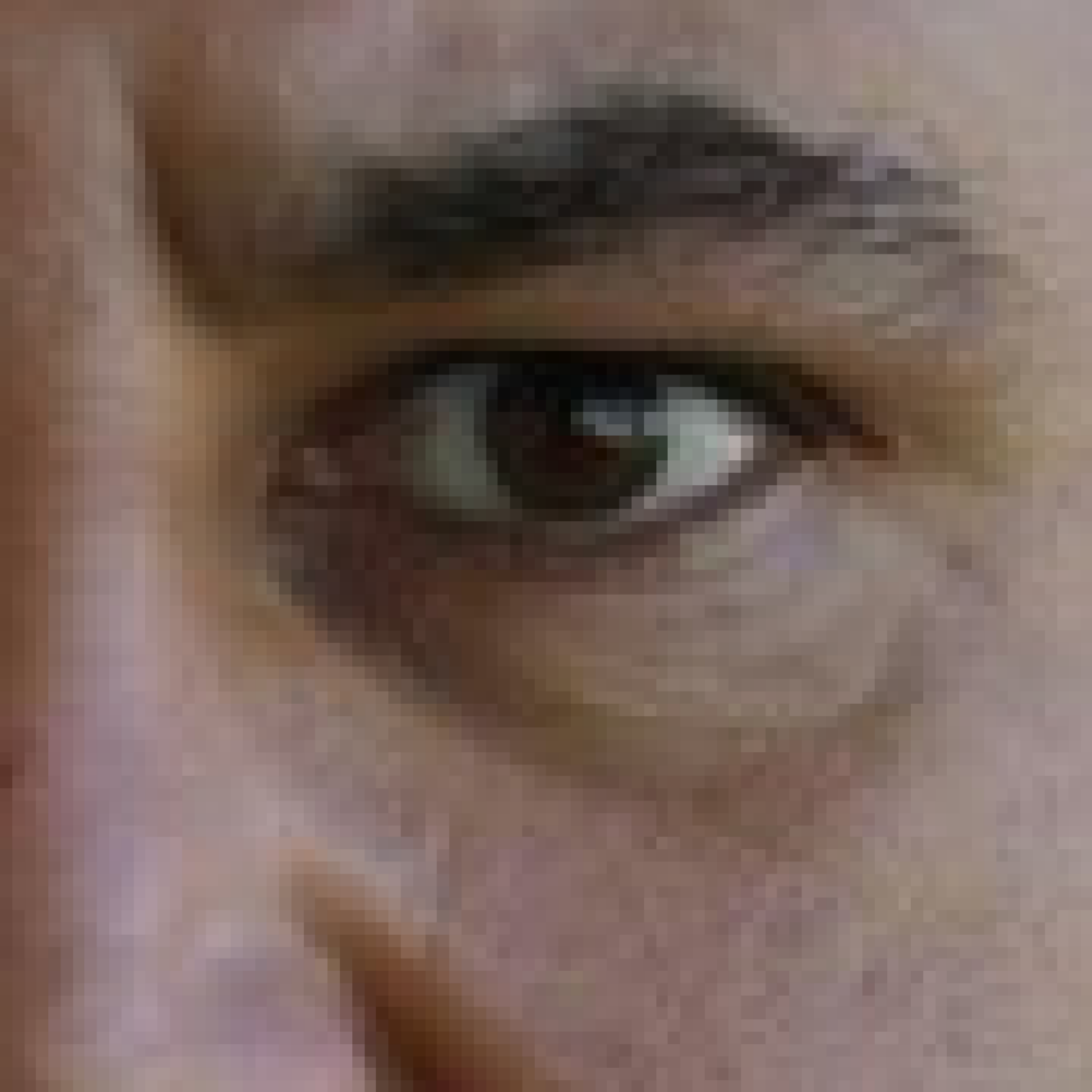}&
    \includegraphics[width=0.105\linewidth]{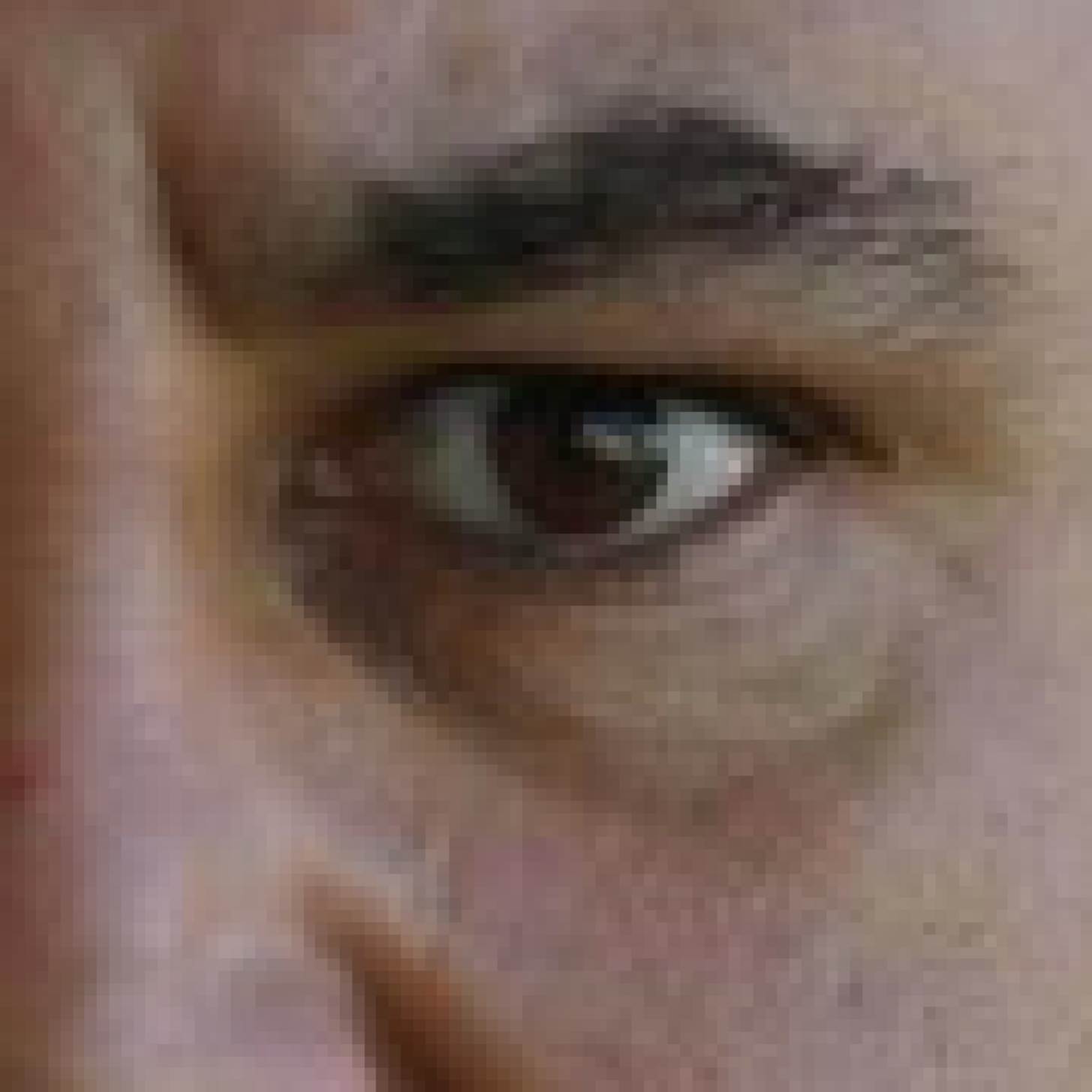}&
    \includegraphics[width=0.105\linewidth]{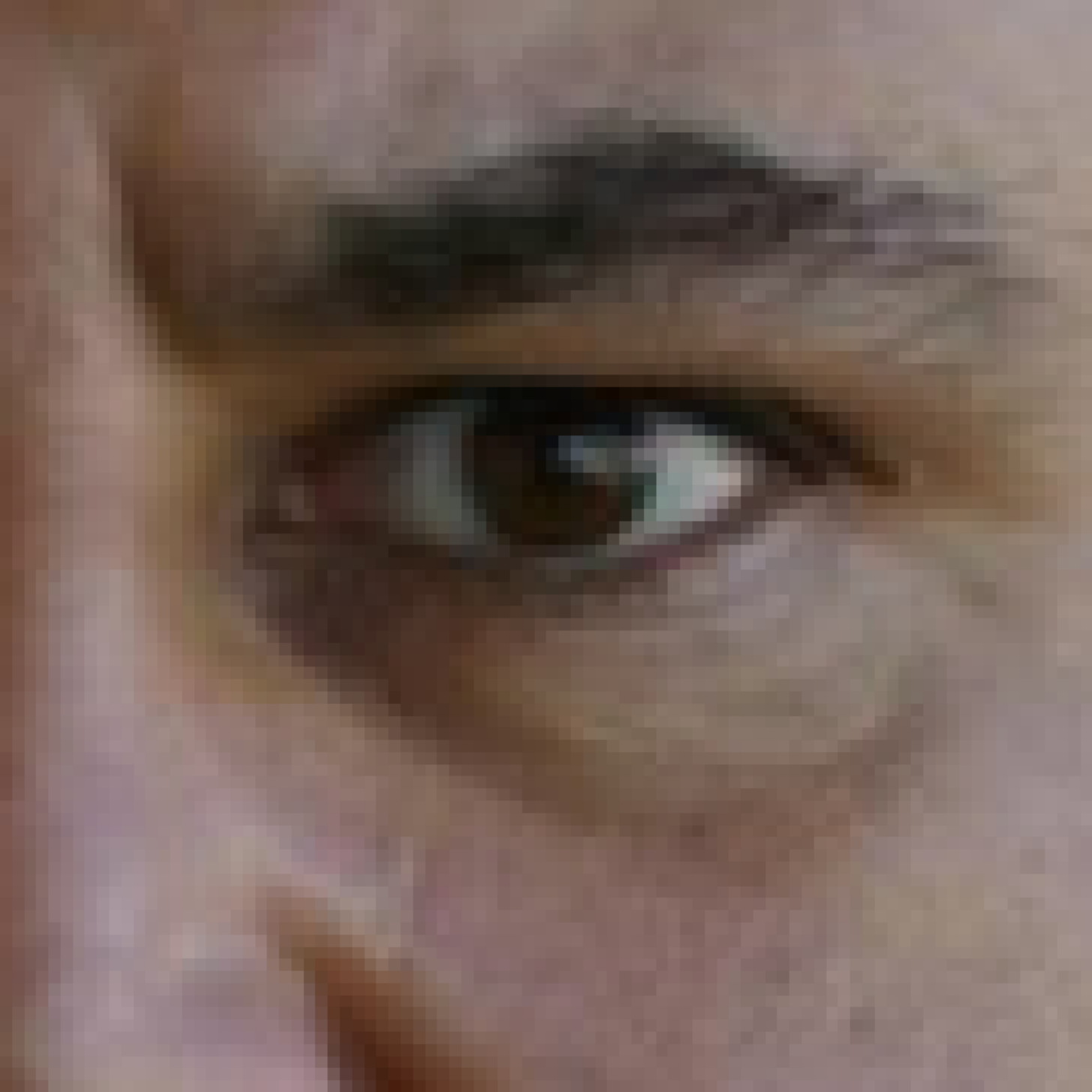}&
    \includegraphics[width=0.105\linewidth]{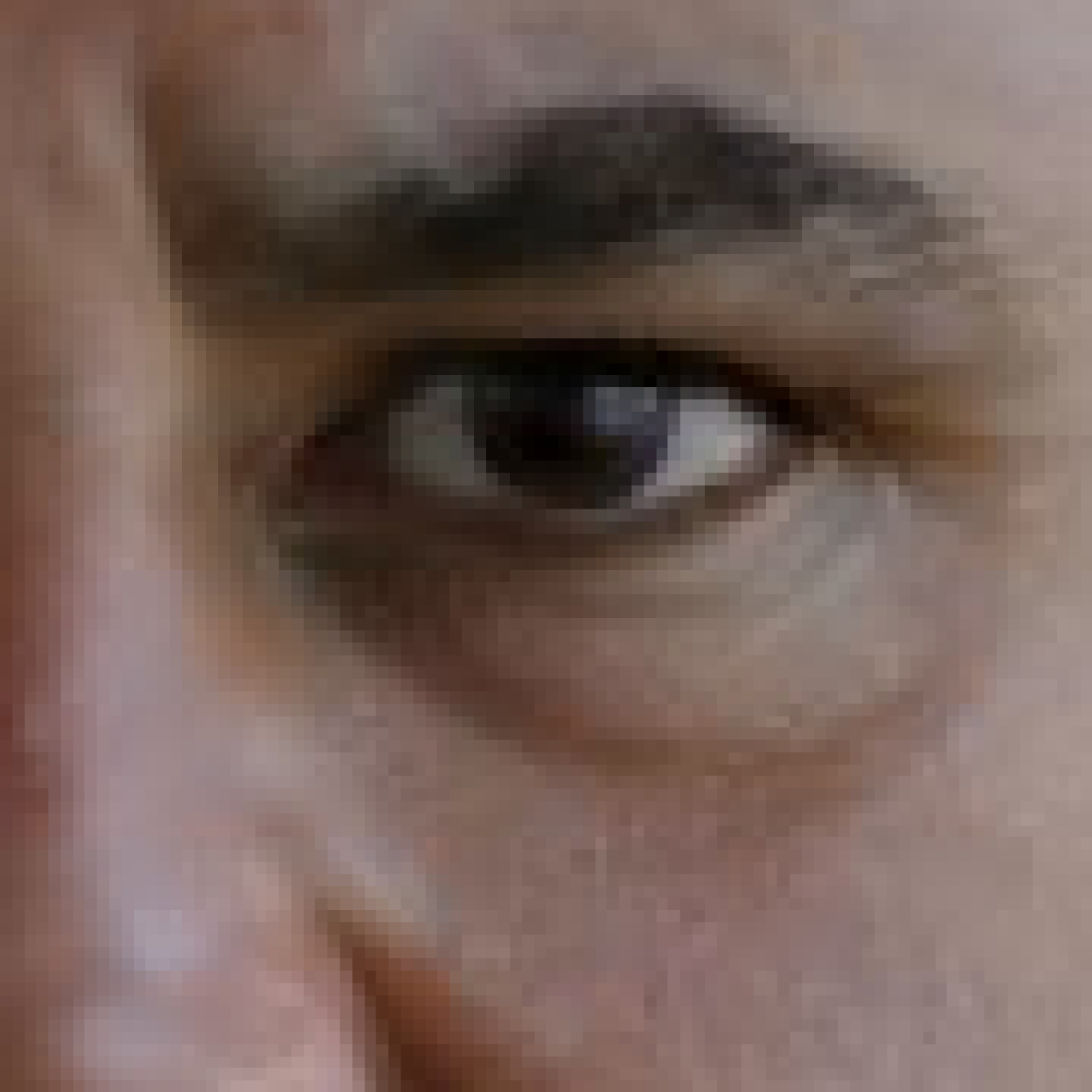}&
    \includegraphics[width=0.105\linewidth]{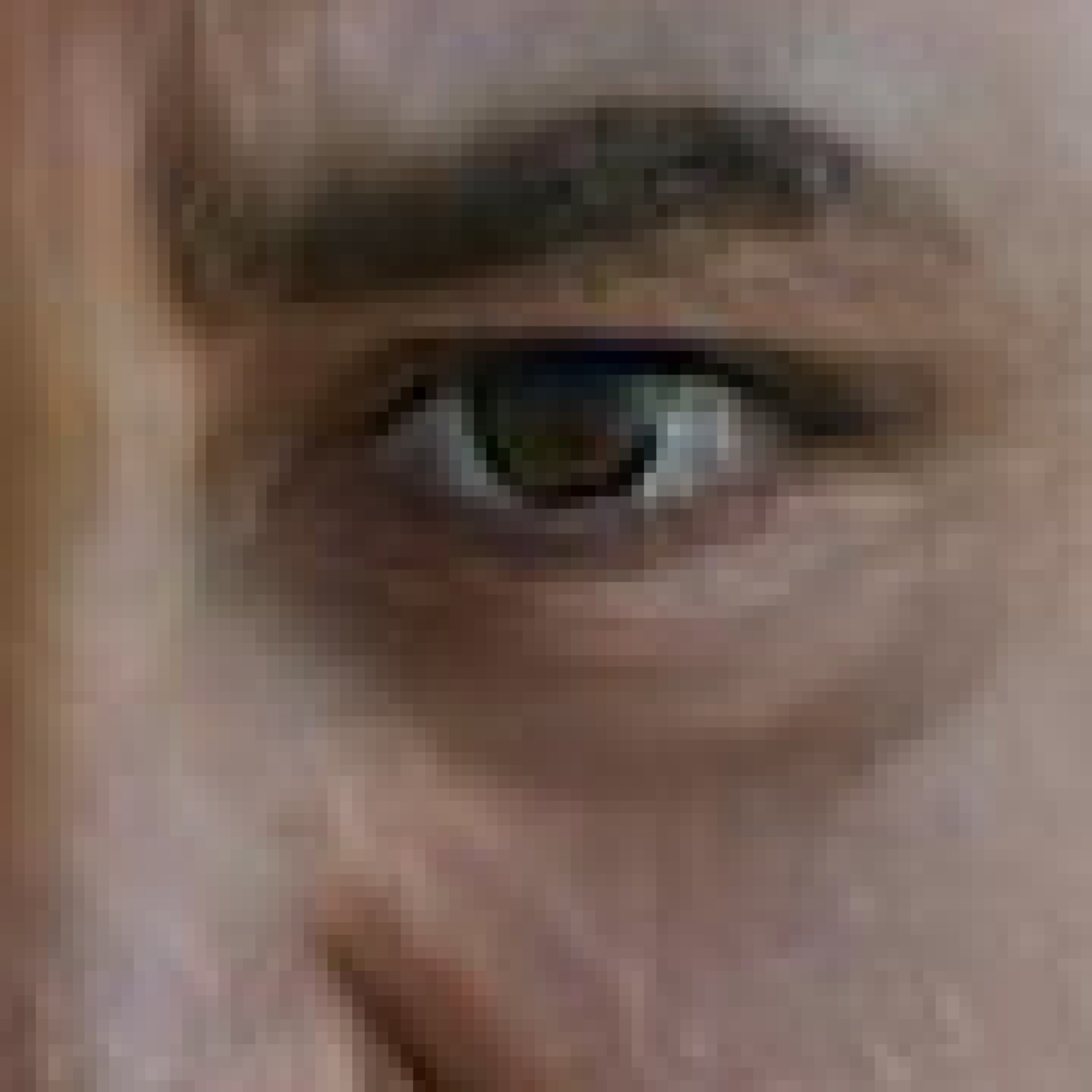}&
    \includegraphics[width=0.105\linewidth]{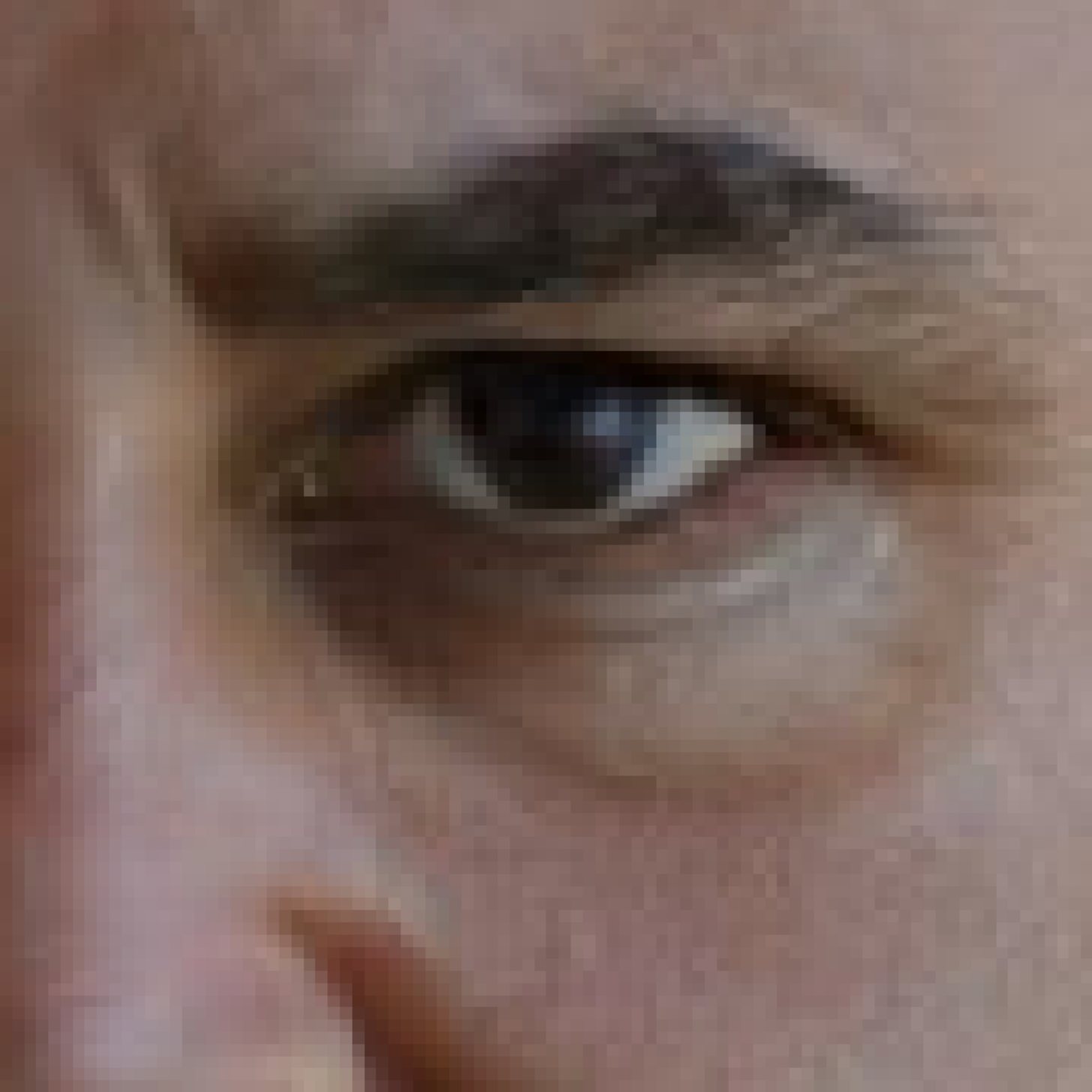}&
    \includegraphics[width=0.105\linewidth]{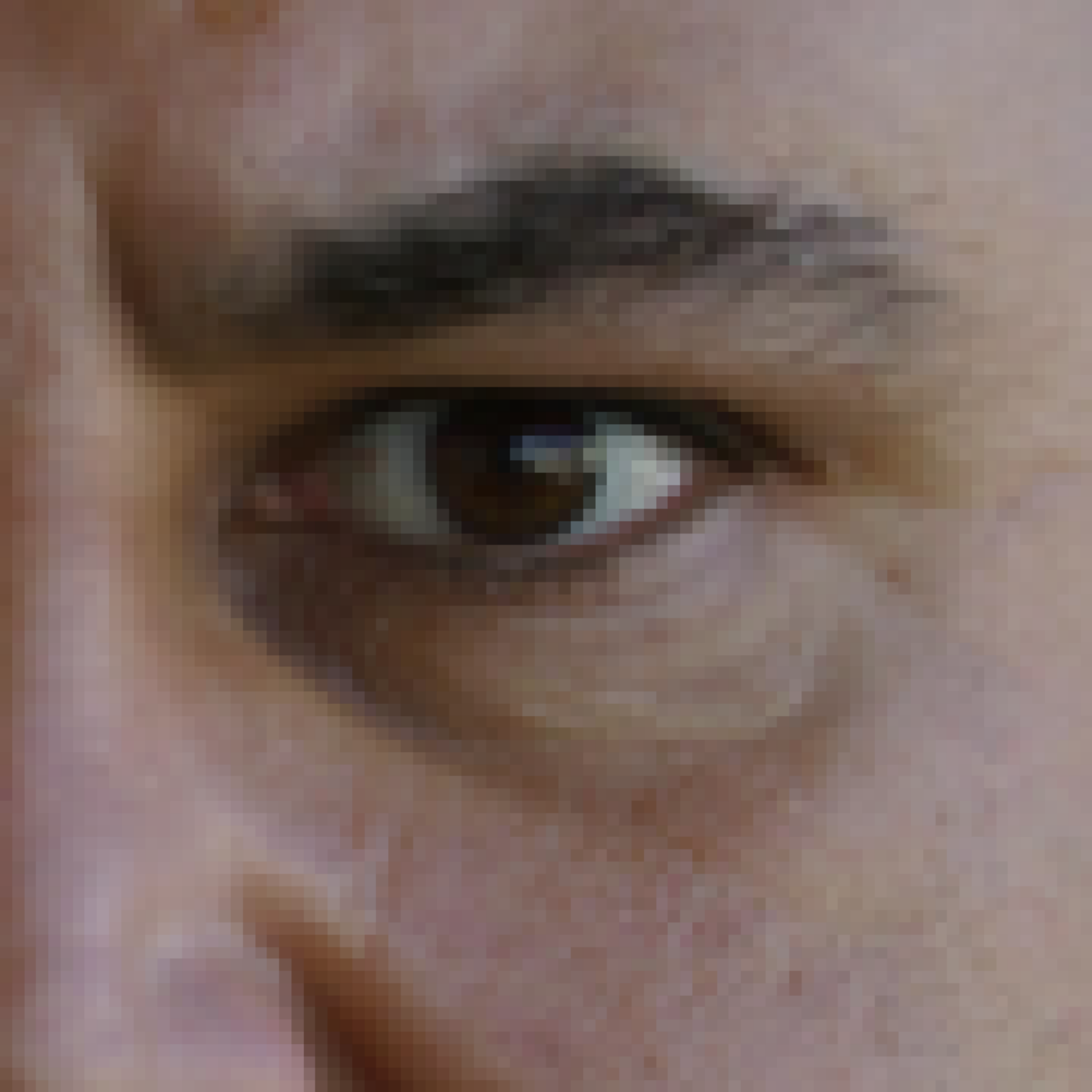}&
    \includegraphics[width=0.105\linewidth]{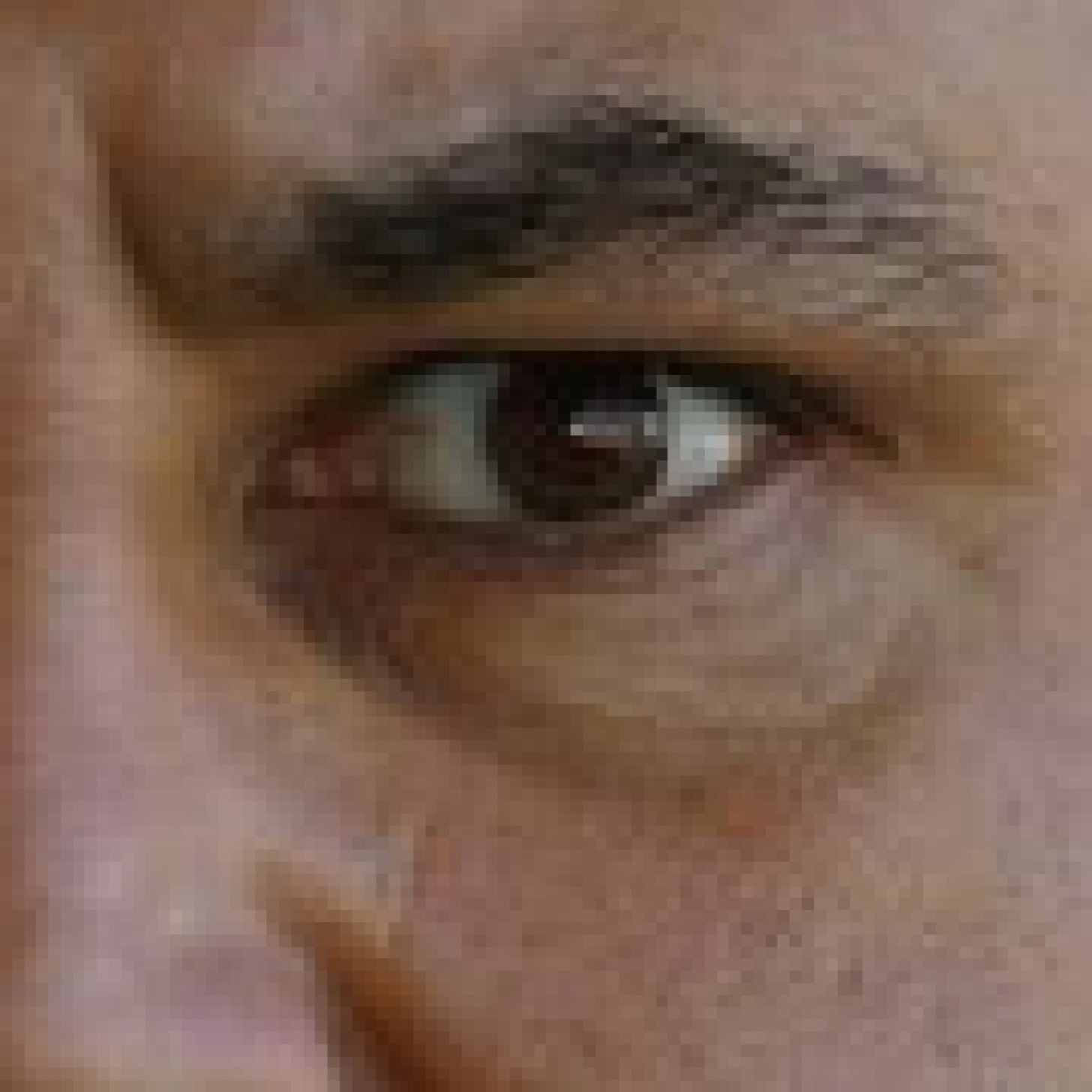}\\
    \vspace{-0.07cm}
    \includegraphics[width=0.105\linewidth]{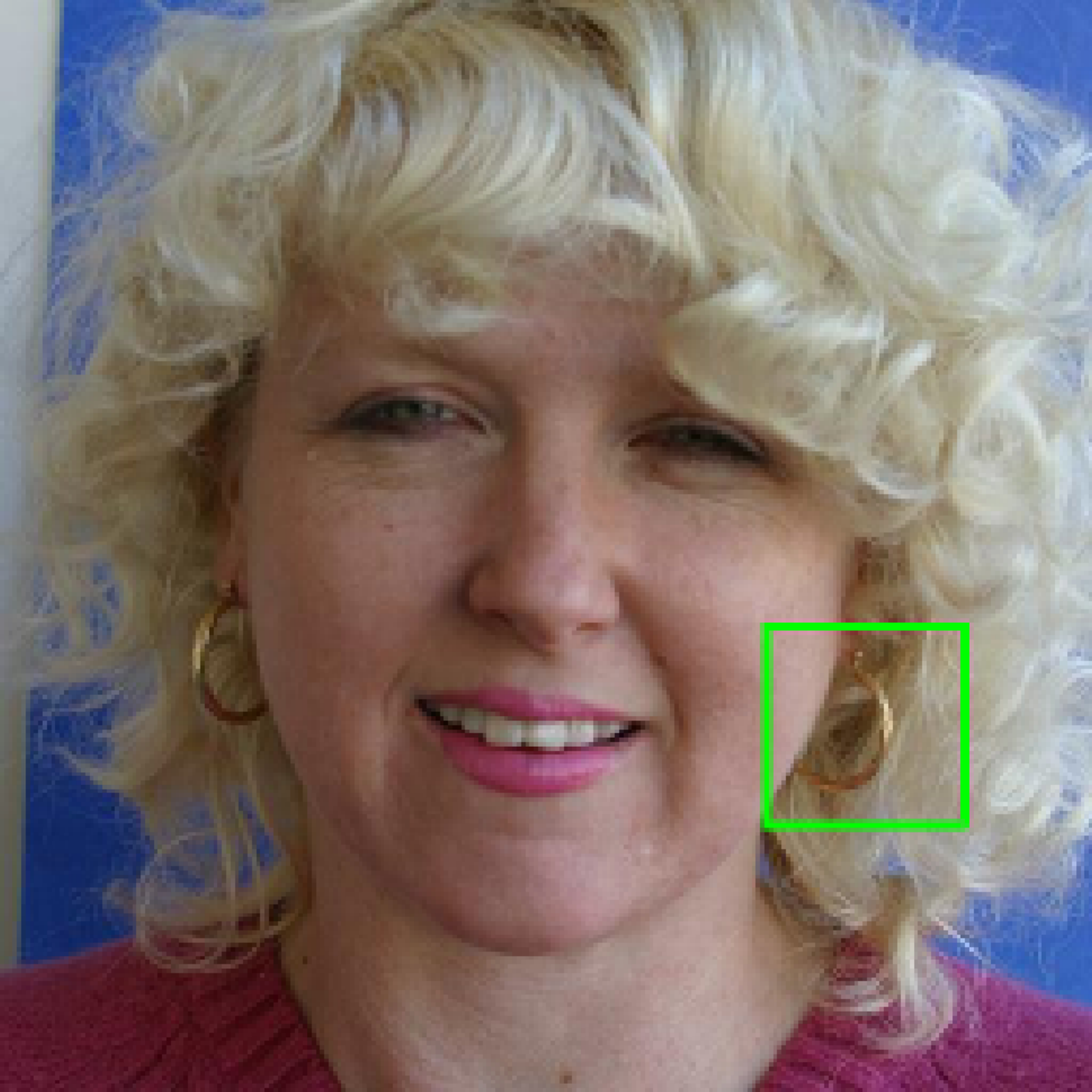}&
    \includegraphics[width=0.105\linewidth]{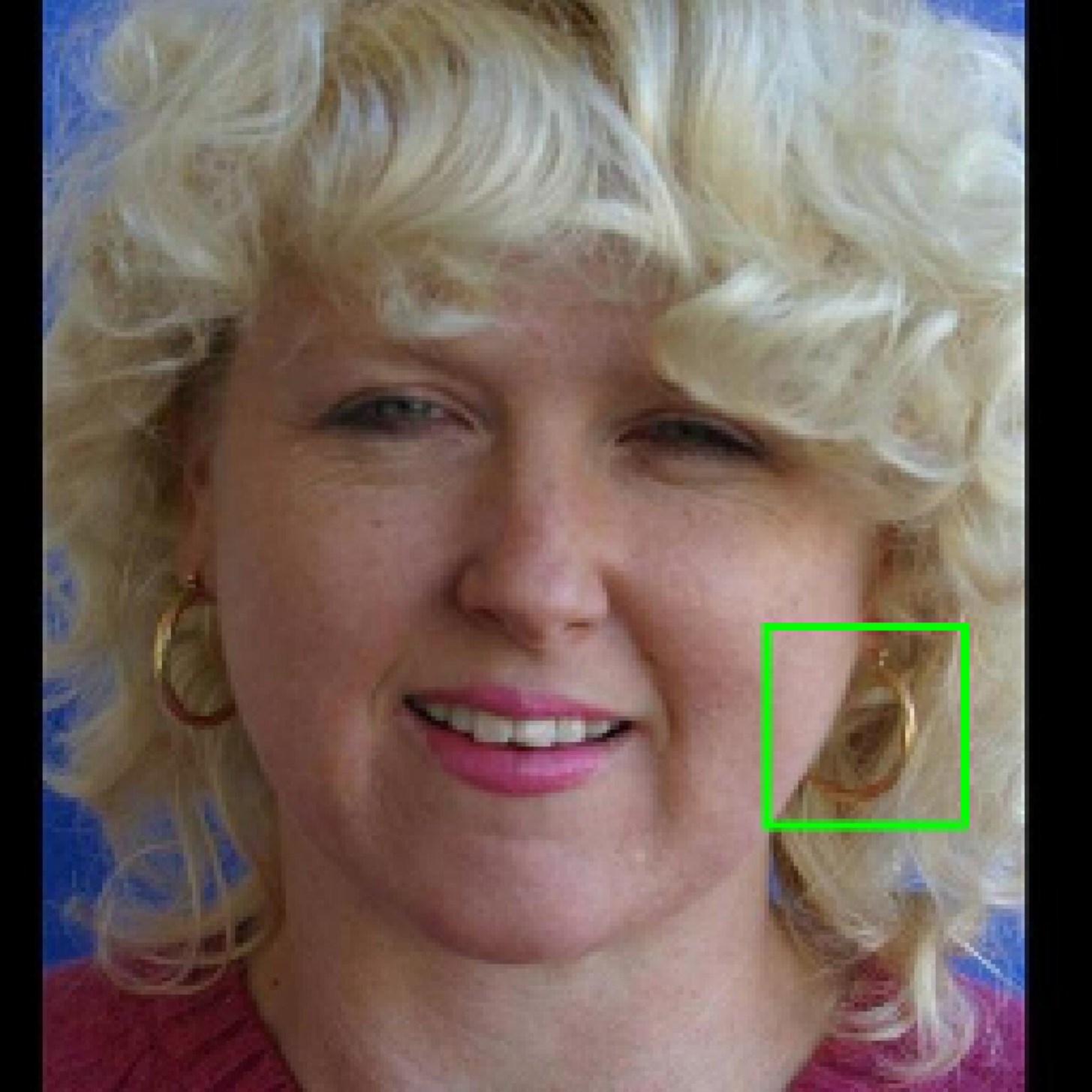}&
    \includegraphics[width=0.105\linewidth]{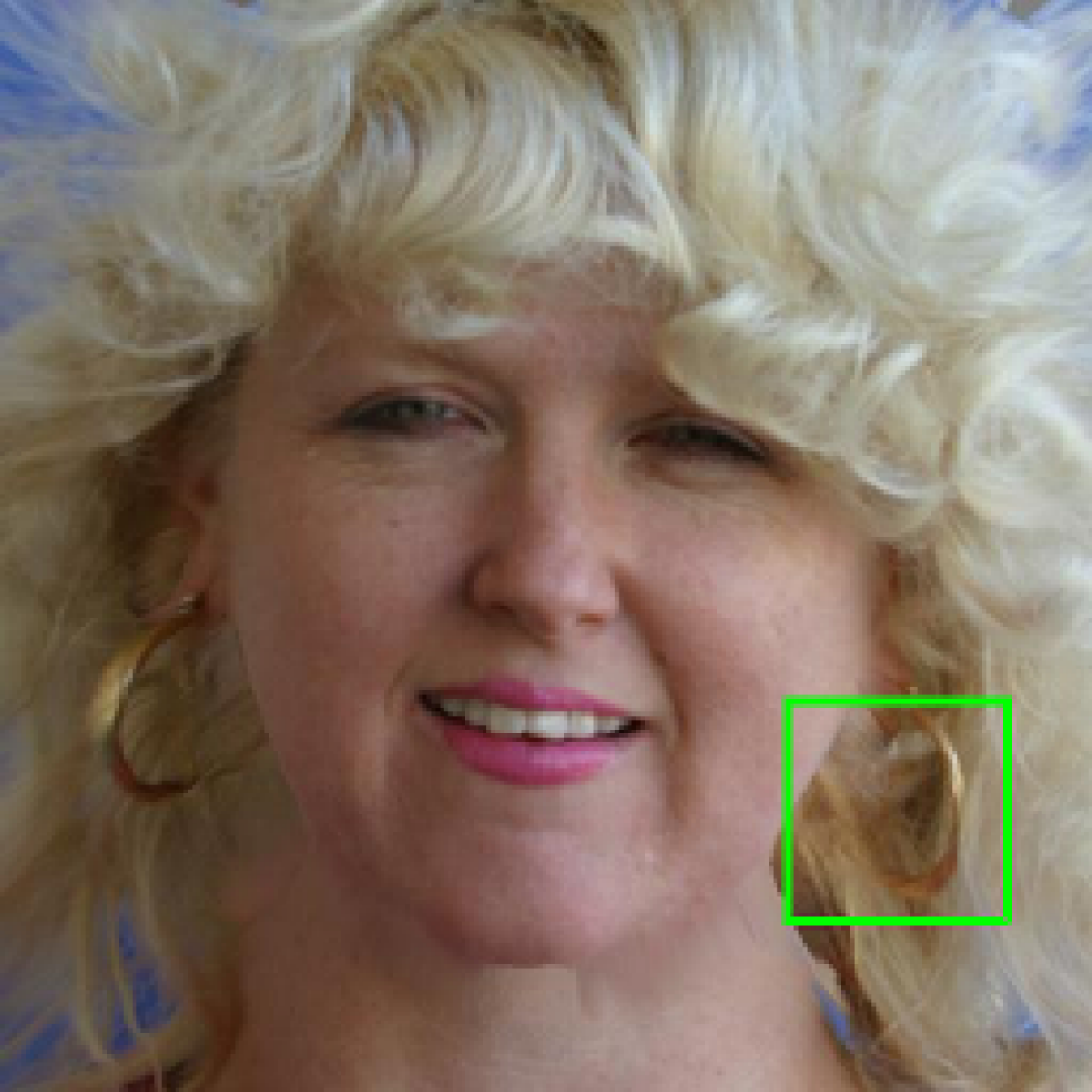}&
    \includegraphics[width=0.105\linewidth]{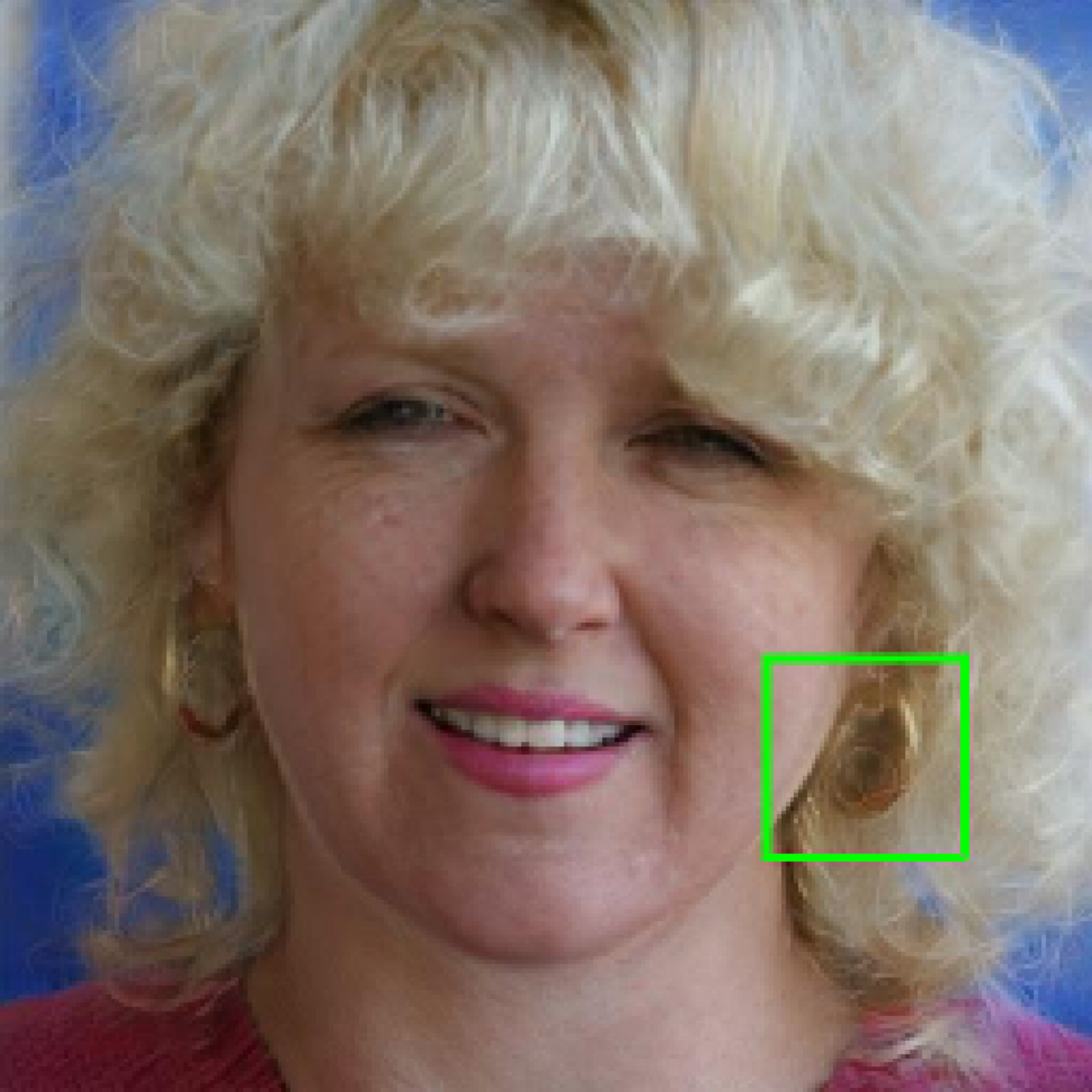}&
    \includegraphics[width=0.105\linewidth]{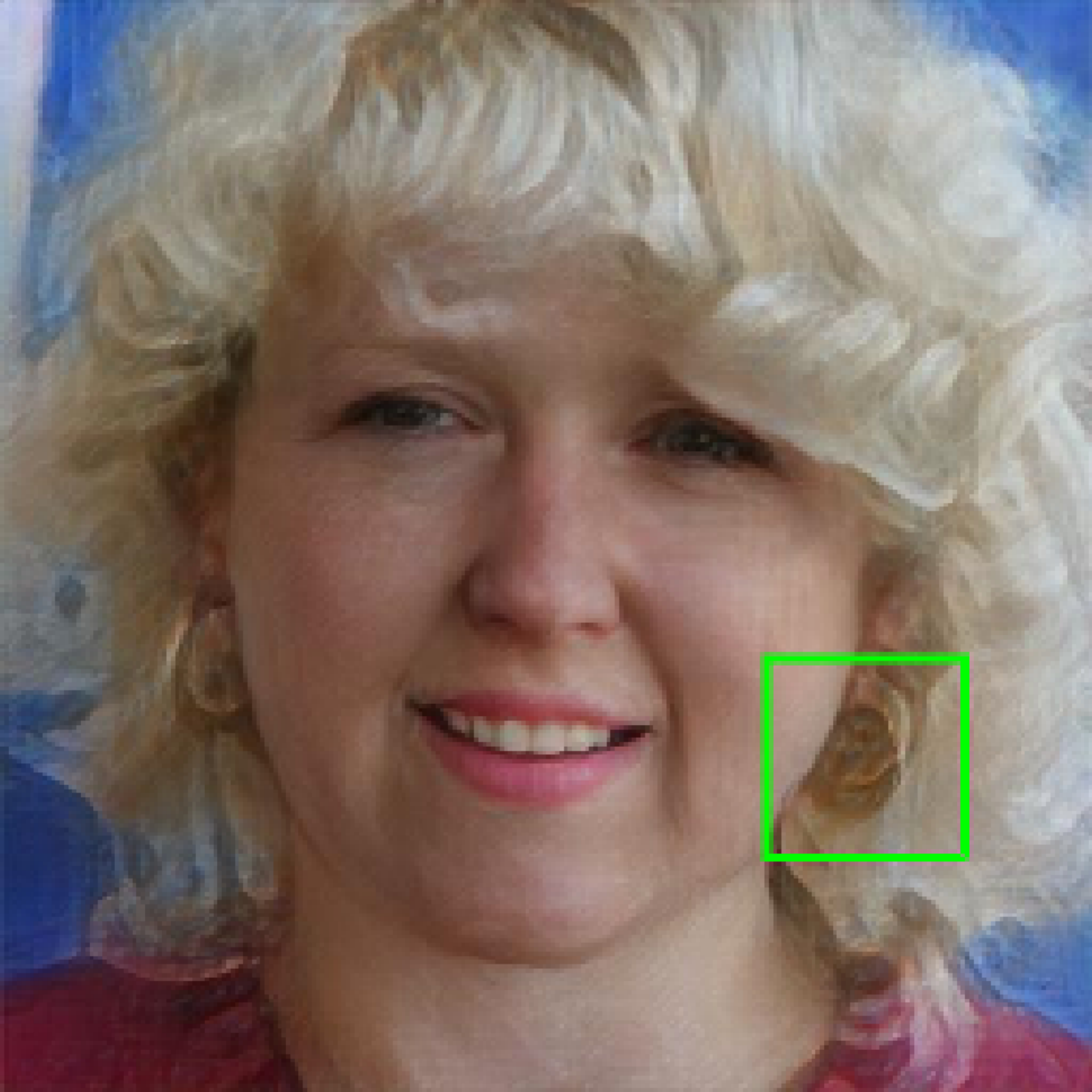}&
    \includegraphics[width=0.105\linewidth]{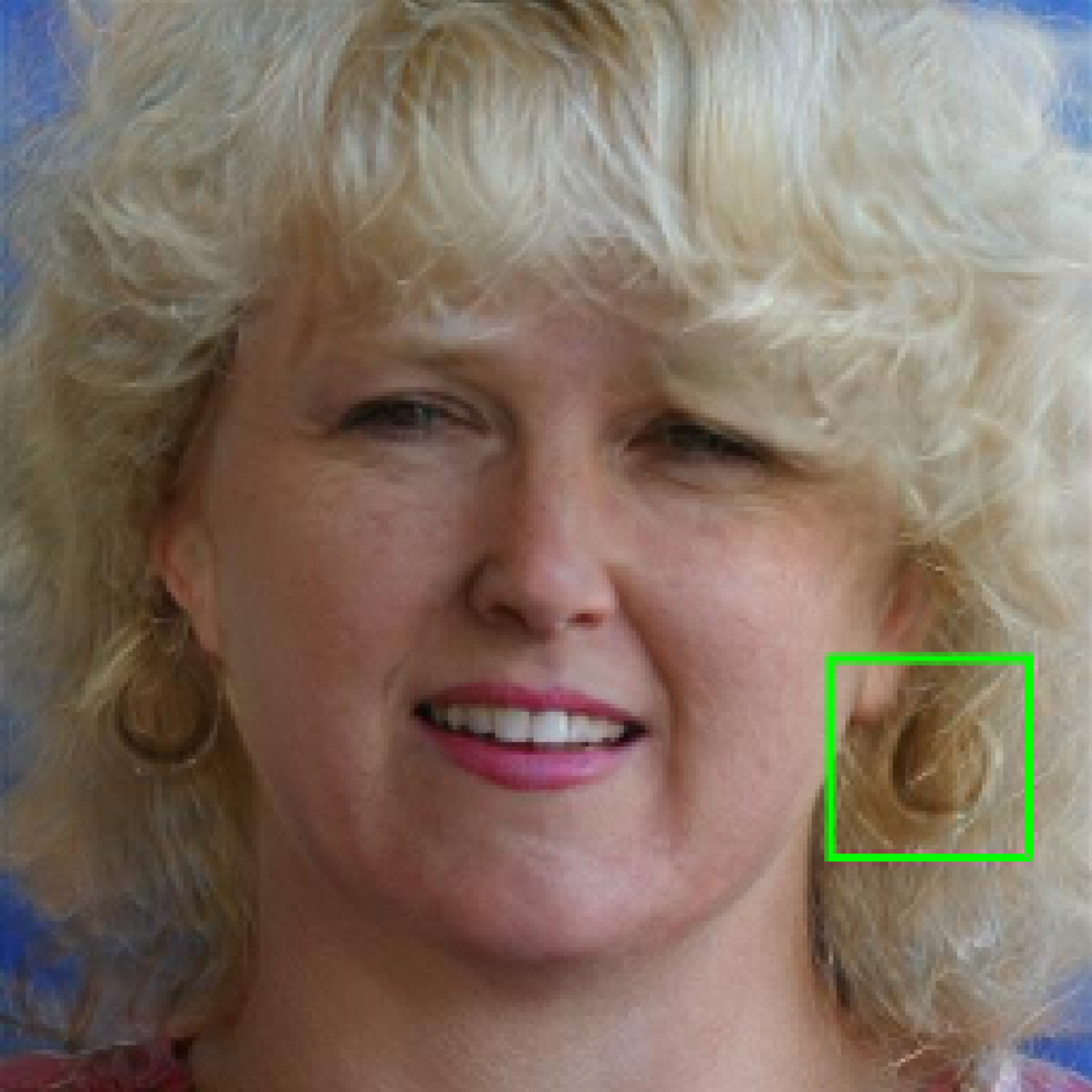}&
    \includegraphics[width=0.105\linewidth]{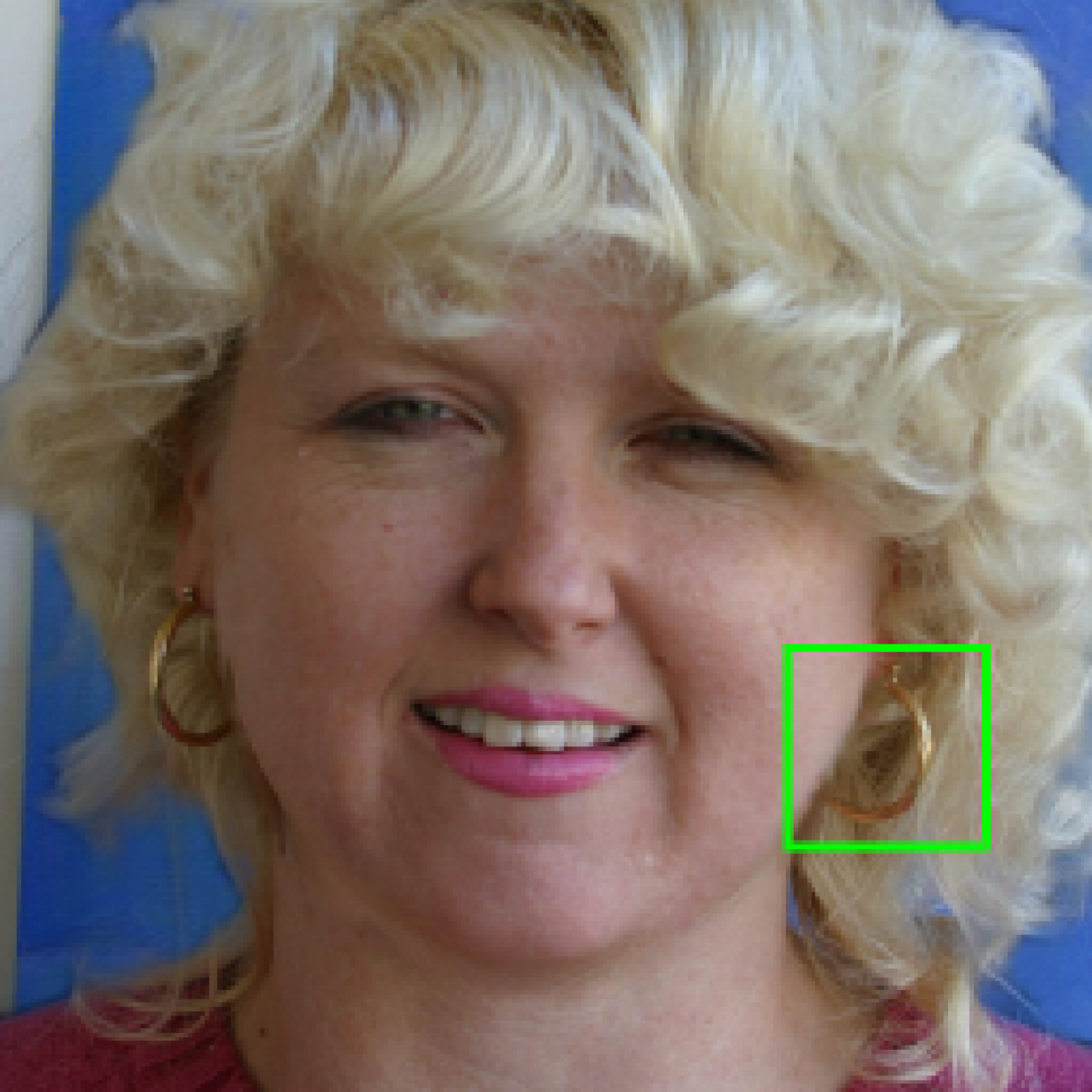}&
    \includegraphics[width=0.105\linewidth]{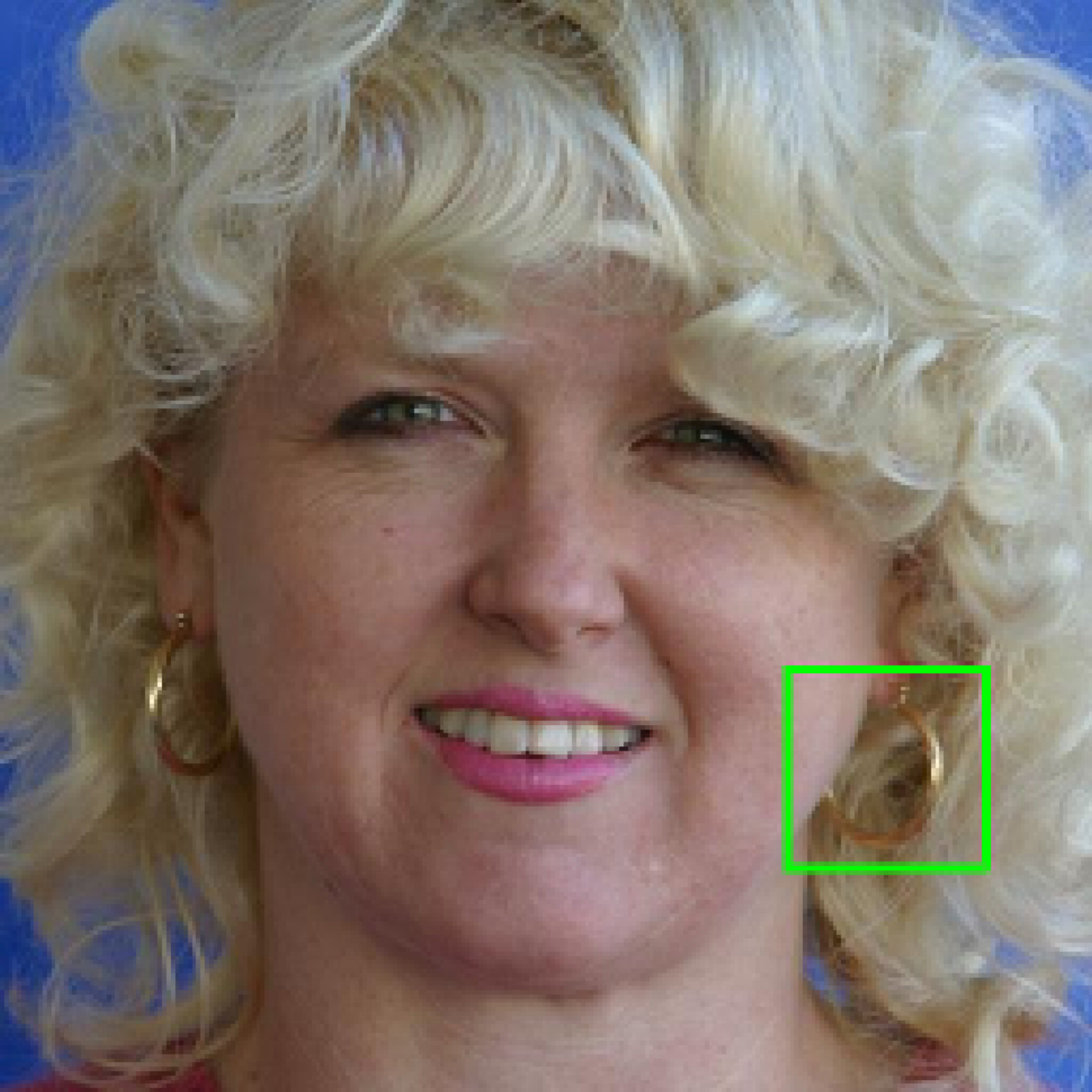}\\
    \vspace{-0.07cm}
    \includegraphics[width=0.105\linewidth]{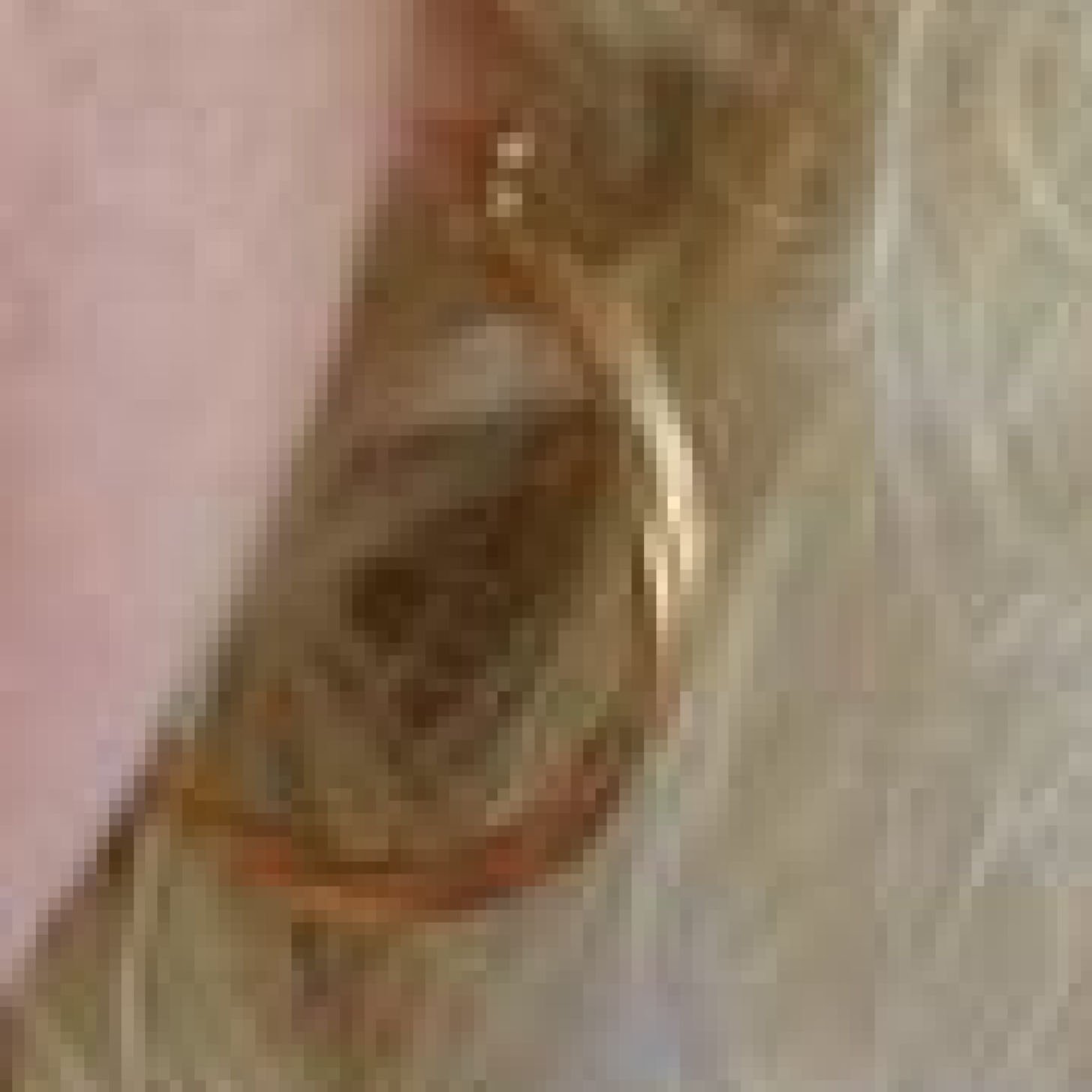}&
    \includegraphics[width=0.105\linewidth]{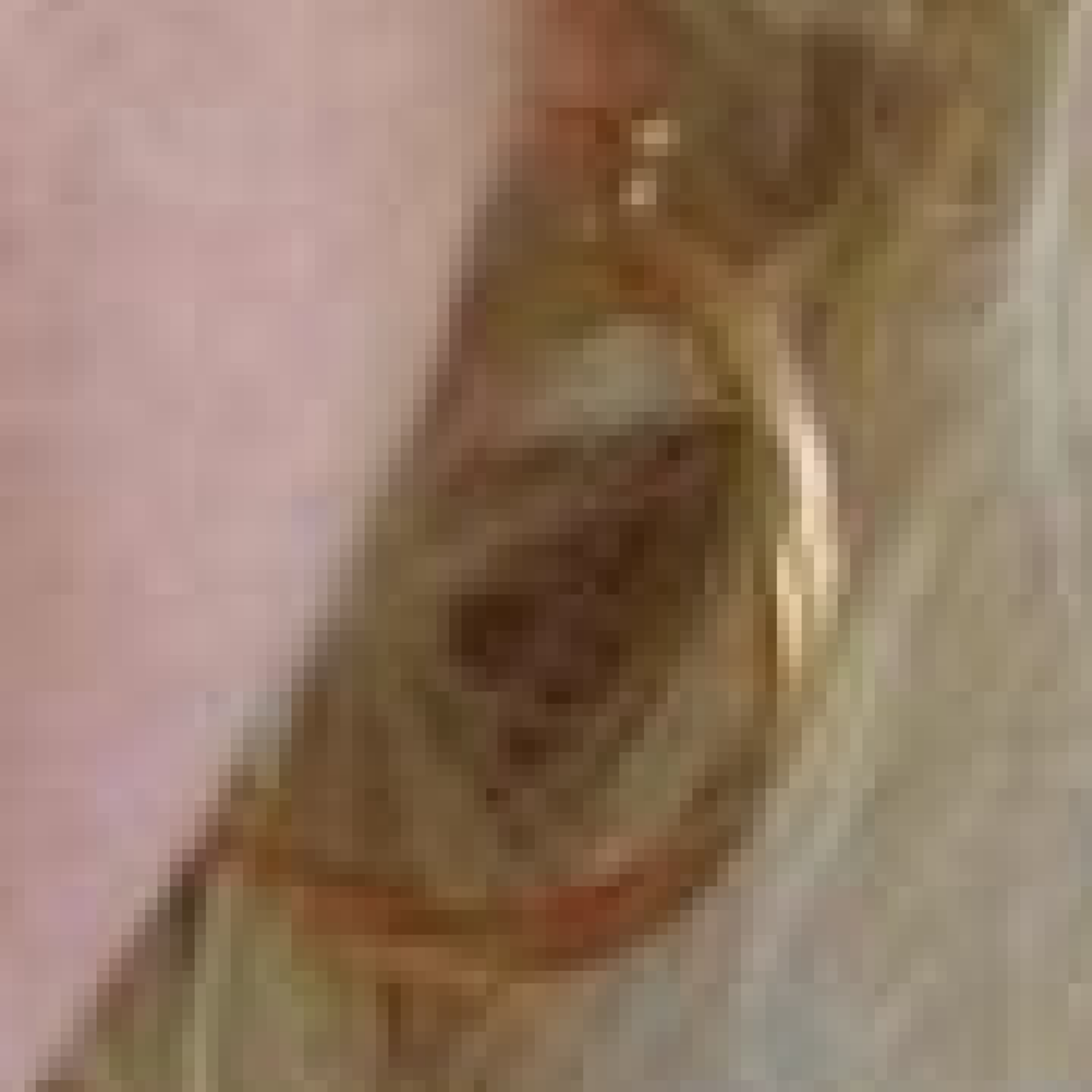}&
    \includegraphics[width=0.105\linewidth]{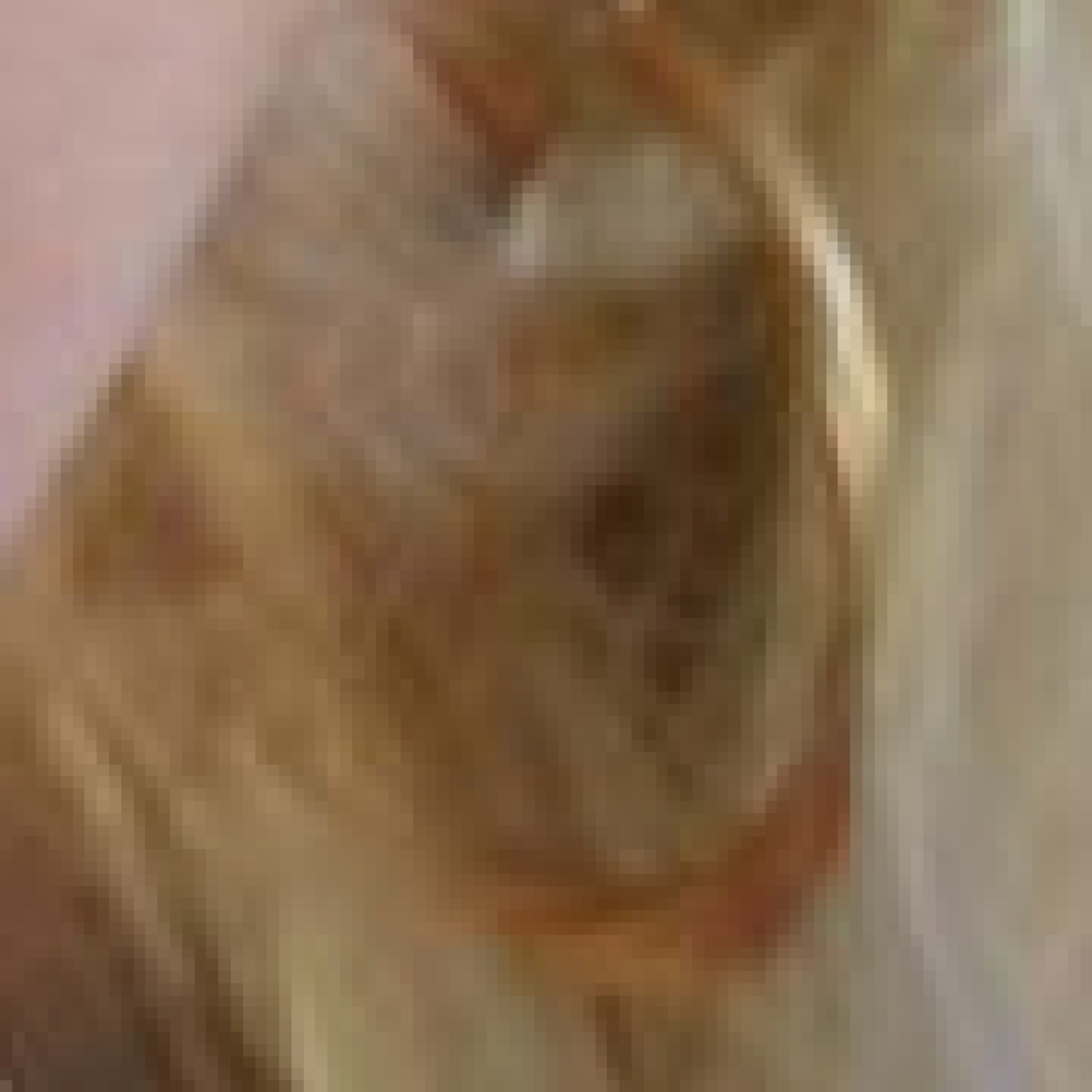}&
    \includegraphics[width=0.105\linewidth]{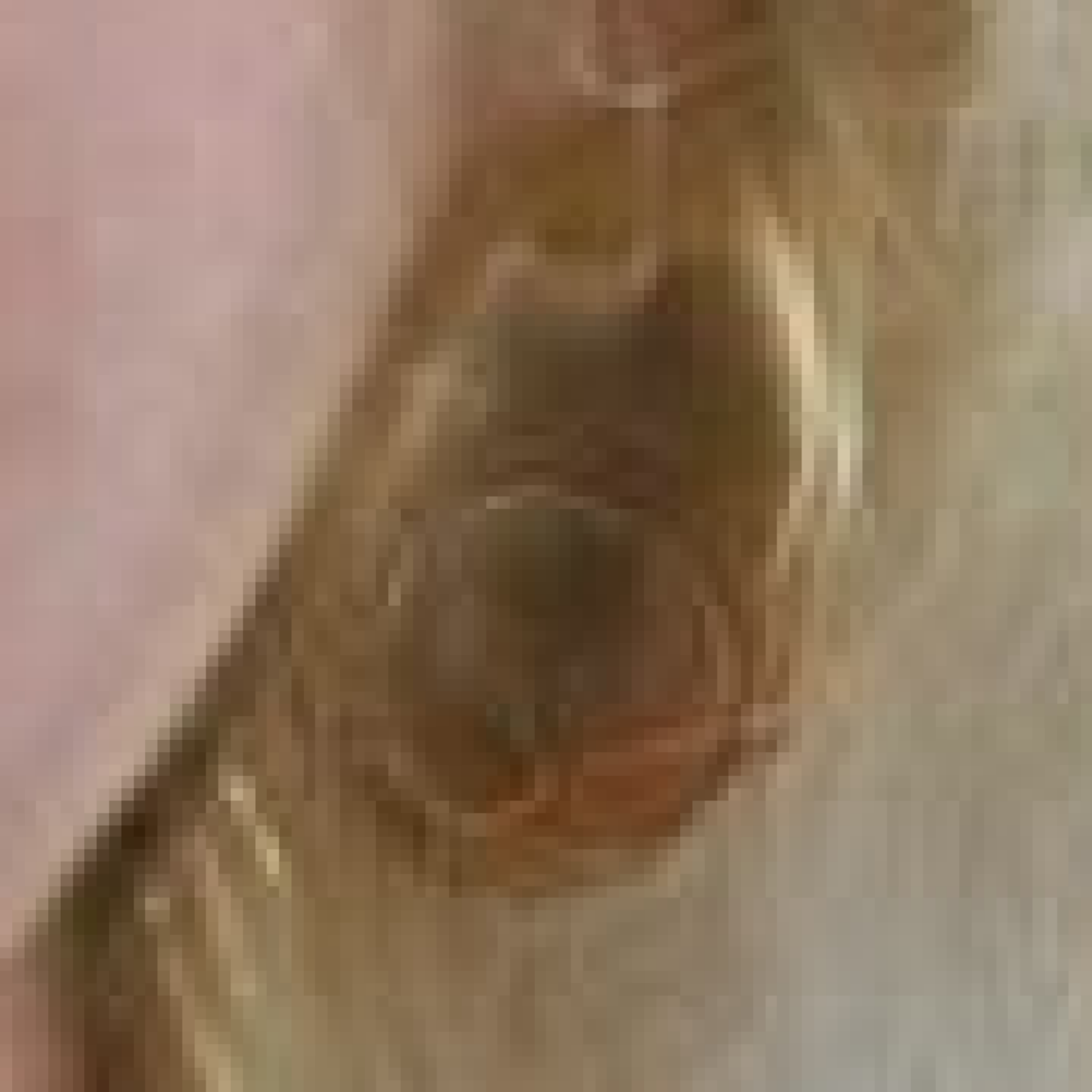}&
    \includegraphics[width=0.105\linewidth]{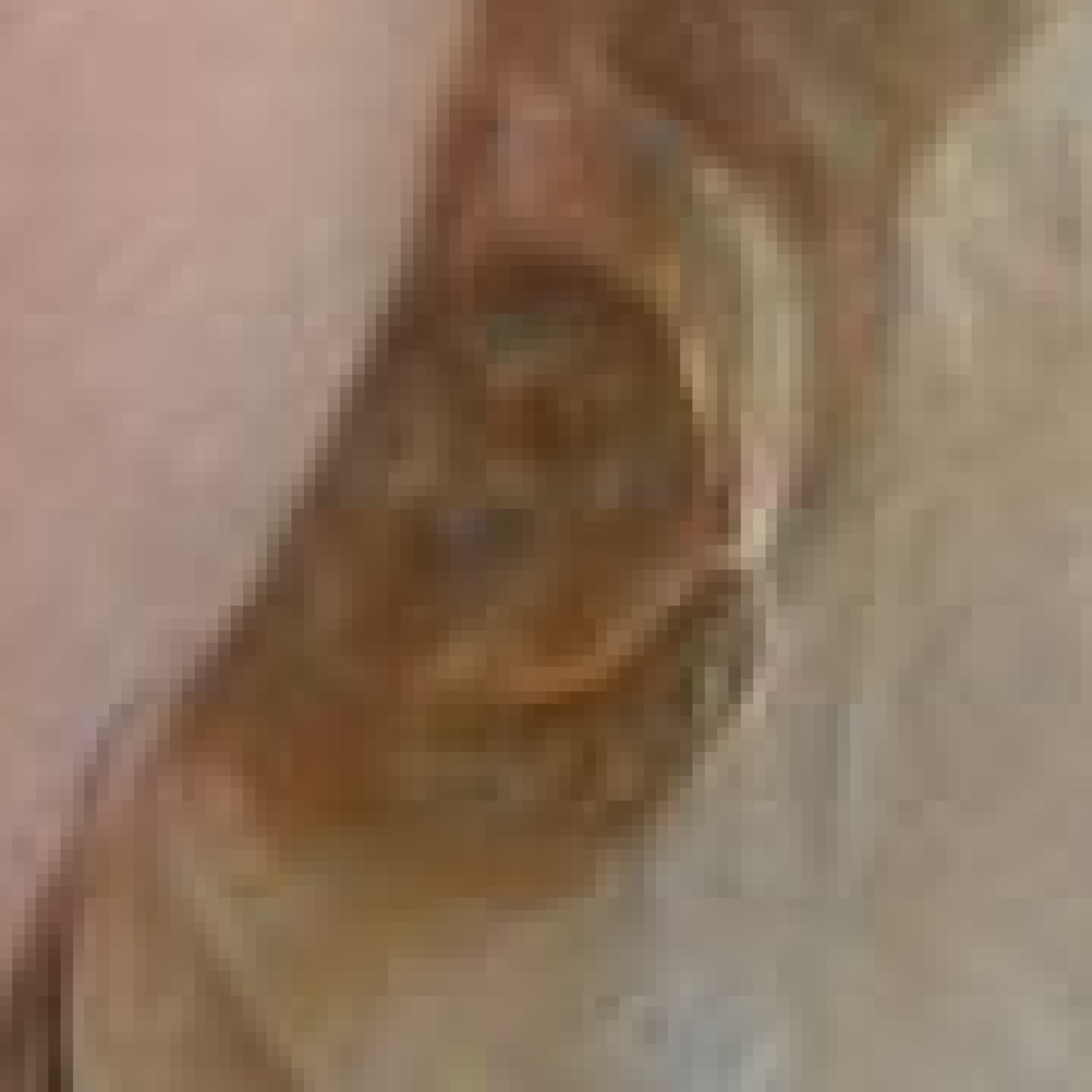}&
    \includegraphics[width=0.105\linewidth]{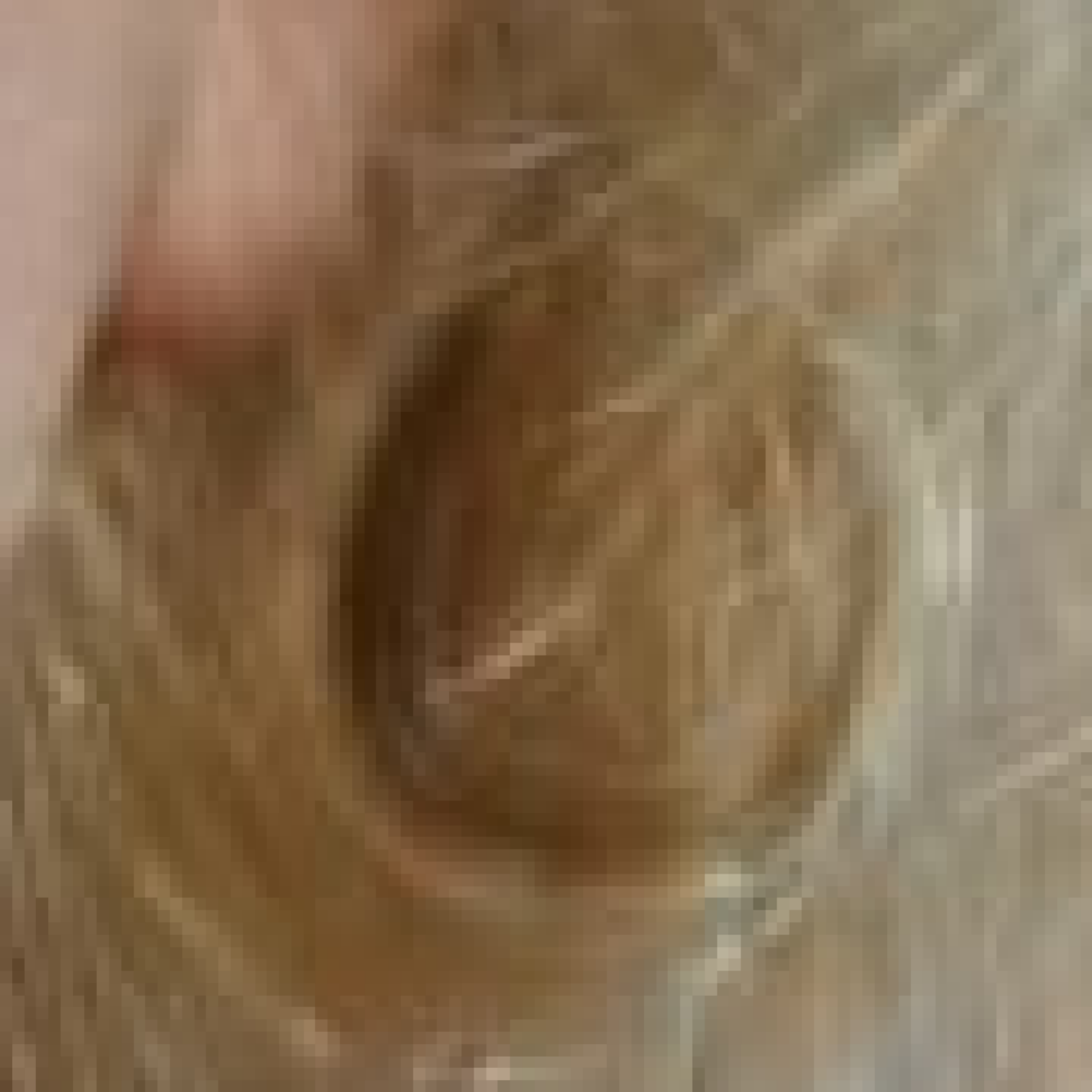}&
    \includegraphics[width=0.105\linewidth]{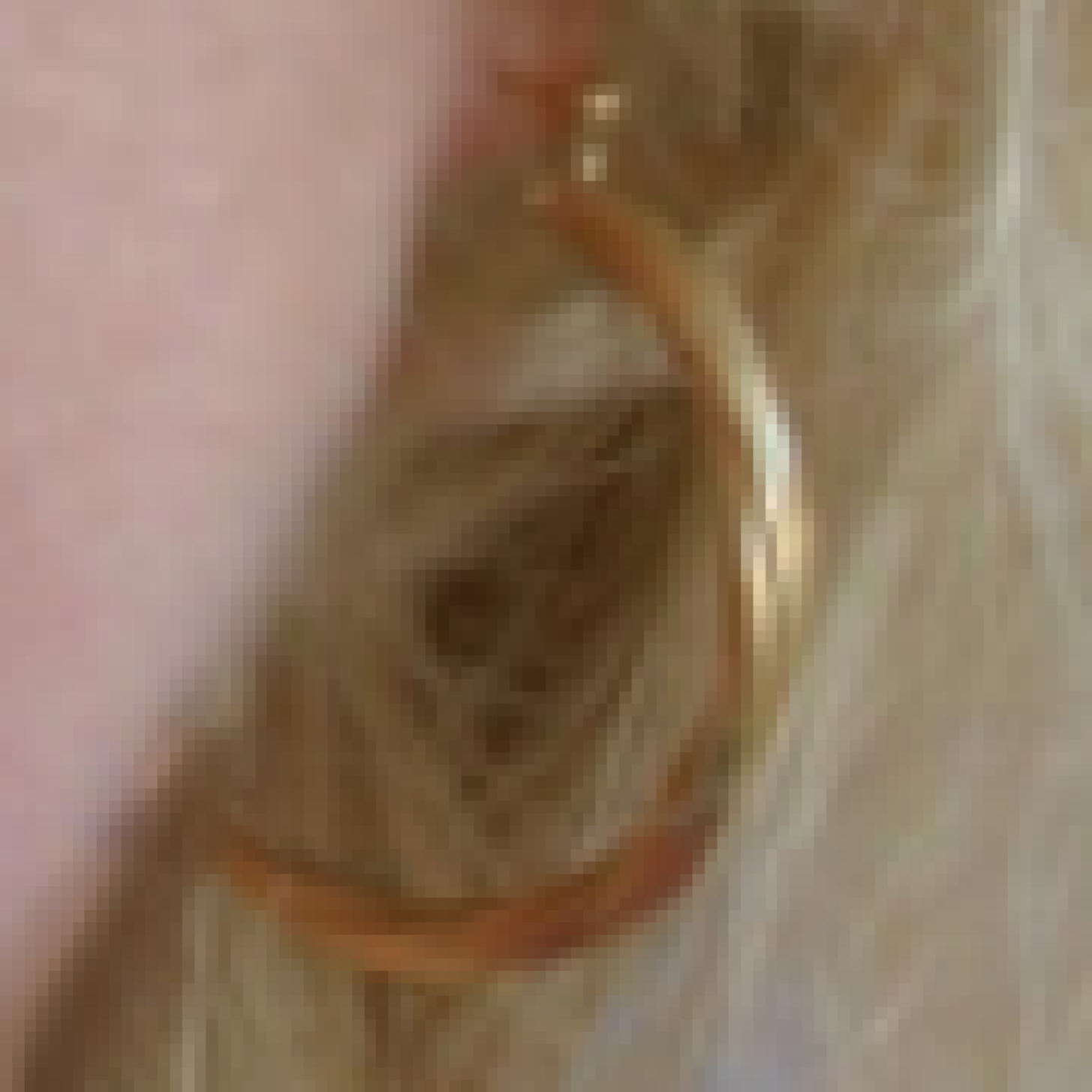}&
    \includegraphics[width=0.105\linewidth]{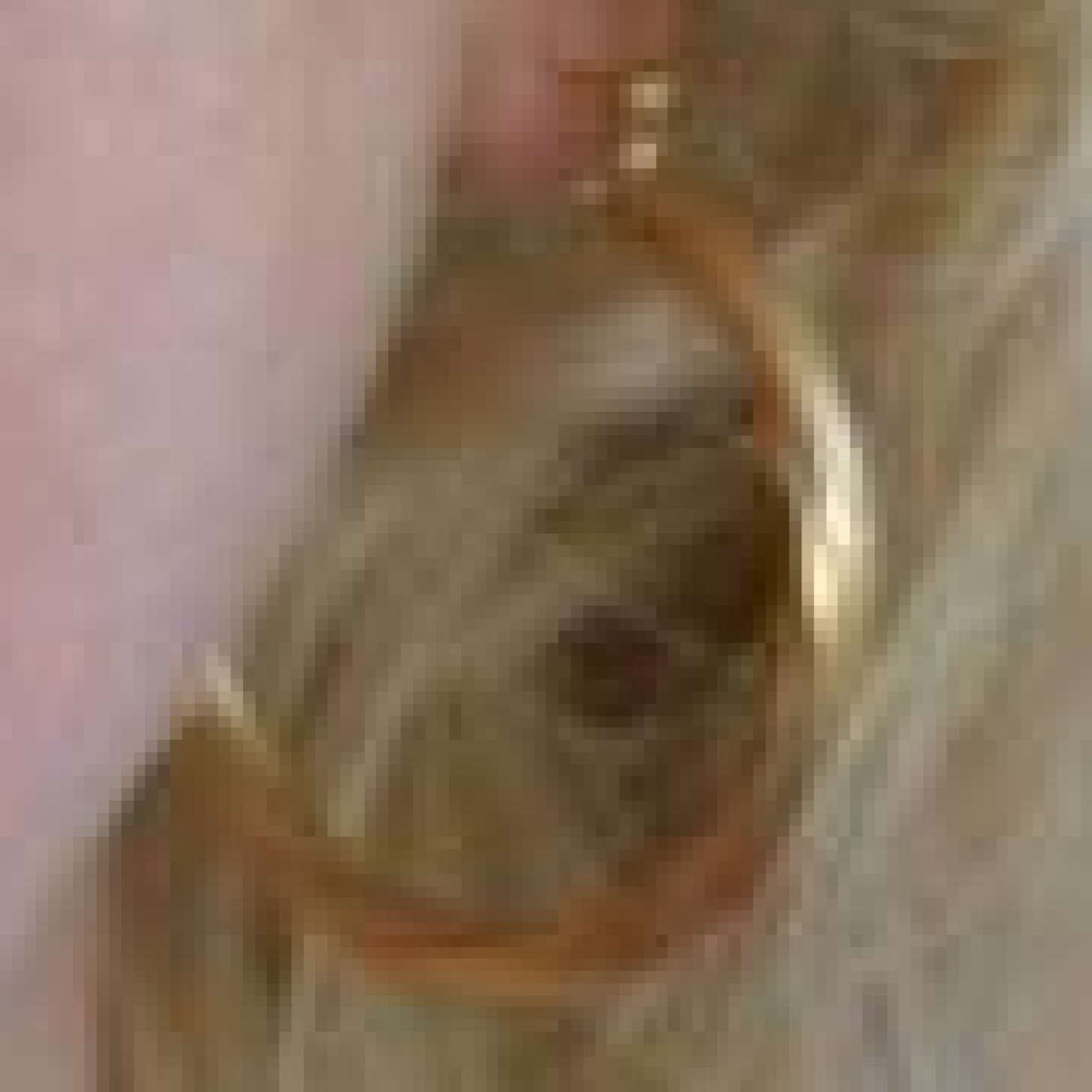} \\
  \end{tabular}
  \vspace{-0.2cm}
  \caption{Qualitative comparisons on the CMDP dataset. \ours{} excels in processing of severely distorted faces. The model not only successfully restores occluded regions, but also preserves crucial identity details, as highlighted in crops.}
  \label{fig:cmdp_results}
\end{figure*}

\section{EXPERIMENTS}
\label{sec:experiments}

\subsection{Datasets}
\label{ssec:datasets}

\subsubsection{Face undistortion datasets}

We evaluated face undistortion on two publicly available datasets. \textbf{Caltech Multi-Distance Portraits (CMDP) } \cite{fried2016perspective} contains frontal portraits of 53 individuals with various face attributes, each photographed from seven distances. \textbf{In-the-wild images} \cite{wang2023disco} features  severely distorted portraits, scraped from the web; we only consider images from the Unsplash \footnote{https://unsplash.com/} due to licensing issues. As there are no references or ground truth images, we use them only for qualitative comparison.

\subsubsection{Head rotation dataset}



We collected our own Head Rotation (HeRo) dataset, containing portraits of 19 people with varying attributes (glasses, facial hair, facial expressions). Overall, there are 67 series of four photos in each series.

Portraits were taken using frontal cameras of four Samsung Galaxy S23FE smartphones fixed on a rig (Fig. ~\ref{fig:capturing-setup}). All the devices were synchronized with \cite{Ansari_2019}, so that series of 4 images were captured simultaneously. 

The participants stood in front of Camera 1 (Front) at a variety of distances, from close-up to far. Other cameras were configured to observe the photographed person at an average angle of $\approx$ 5 degrees in the Left, Right and Top directions. 

\begin{table}[h!]
\renewcommand\tabcolsep{3pt}
\centering 
\begin{tabular}{lcccc} \hline
Methods & PSNR$\uparrow$ &
SSIM$\uparrow$ & LPIPS$\downarrow$ & ID$\uparrow$\\
\hline
Fried’s \cite{fried2016perspective} & 15.41 & 0.724  & 0.188 & 0.893 \\
3DP \cite{shih20203d} & 13.08 & 0.696 & 0.268 & 0.847 \\
PTI \cite{roich2022pivotal} & 15.92 & 0.717 & 0.197 & 0.758 \\
Ko’s  \cite{ko20223d} & 15.41 & 0.710 & 0.206 & 0.689 \\
HFGI3D \cite{Xie2022Highfidelity3G} & 15.75 & 0.724 & 0.198 & 0.829 \\
TriPlaneNet \cite{bhattarai2024triplanenet} & 14.80 & 0.705 & 0.243 & 0.812 \\
DisCO \cite{wang2023disco} & 17.52 & 0.747 & 0.167 & 0.859 \\
\textbf{\ours, ours} & \textbf{17.78} & \textbf{0.777} & \textbf{0.151} & \textbf{0.932} \\ 
\hline
\end{tabular}
\caption{
Quantitative comparison of face undistortion methods on CMDP \cite{fried2016perspective}. \ours{} significantly outperforms other methods based on image-quality metrics and ID score. 
}
\vskip -0.25cm
\label{tab:results-undistortion}
\end{table}

\subsection{Metrics}
\label{ssec:metrics}
We use four standard evaluation metrics to assess the portrait perspective correction. These are photometric errors between aligned output images and corresponding references: PSNR, SSIM, and LPIPS \cite{LPIPS}. Besides, we assess identity preservation with an ID score, which is a cosine distance between predicted and reference face features from ArcFace \cite{Deng2018ArcFaceAA}.

\section{RESULTS}
\label{sec:results}

\begin{figure}[h!]
  \centering
  \includegraphics[width=0.95\linewidth]{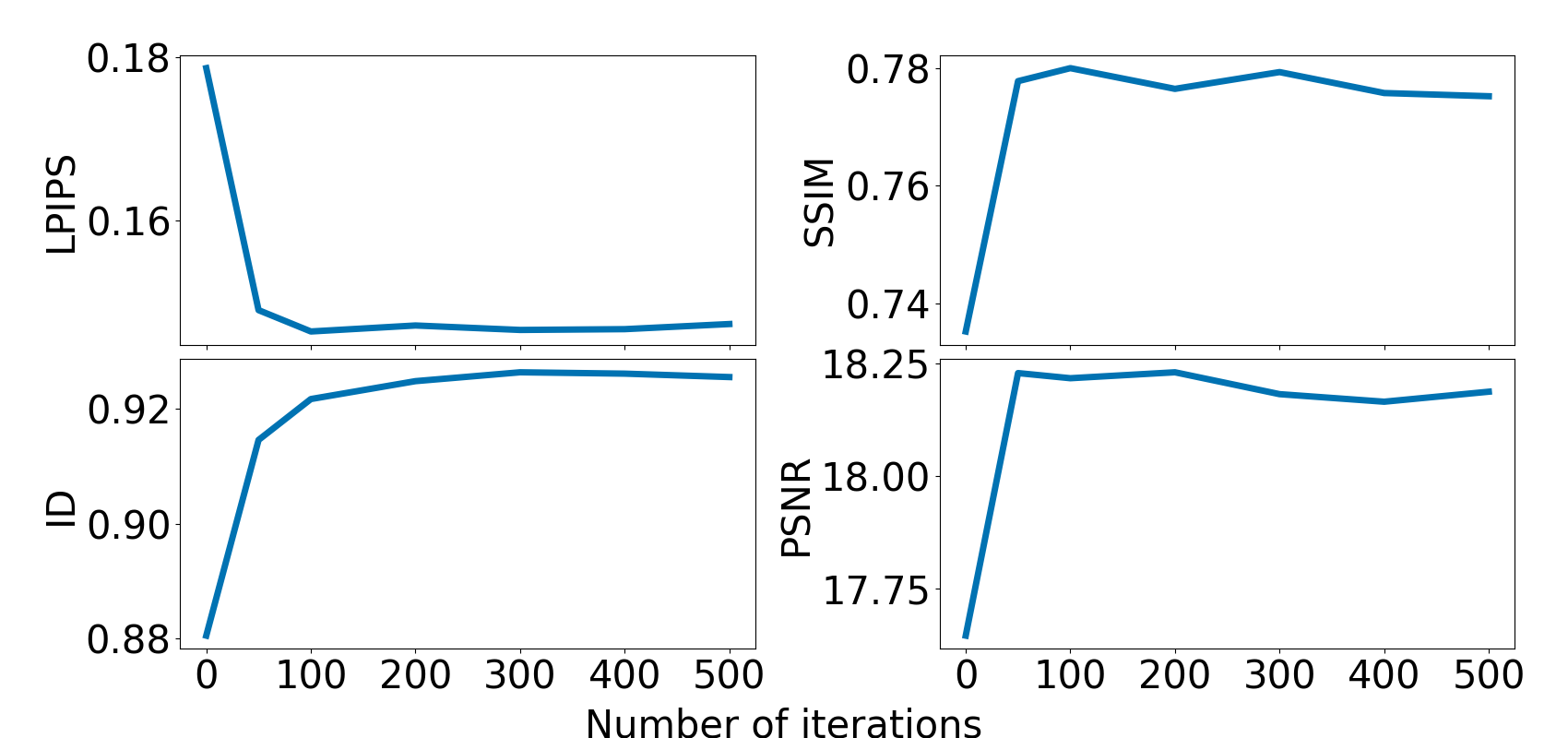}
    \caption{PSNR, SSIM, and LPIPS scores for different number of iterations. Optimal quality is achieved during the first 100 iterations, while further optimization brings a negligible growth of the ID score, yet does not improve other scores.}
    \label{fig:ablation-iterations}
    \vskip -0.25cm
\end{figure}

\begin{figure*}[ht!]
  \centering
  \begin{tabular}{*{7}{@{\hspace{3pt}}c}}
    Source  & PTI \cite{roich2022pivotal} & Ko's \cite{ko20223d} & HFGI3D \cite{Xie2022Highfidelity3G}& 
    \small{TriPlaneNet\cite{bhattarai2024triplanenet}} & \textbf{\ours} & Target  \\
    \includegraphics[width=0.12\linewidth]{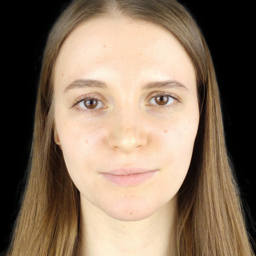}&
    \includegraphics[width=0.12\linewidth]{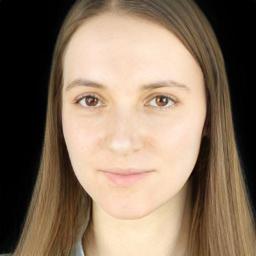}&
    \includegraphics[width=0.12\linewidth]{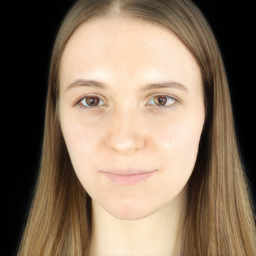}& 
    \includegraphics[width=0.12\linewidth]{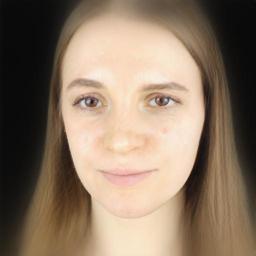}&
    \includegraphics[width=0.12\linewidth]{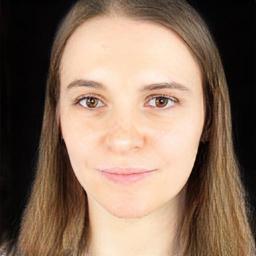} &
    \includegraphics[width=0.12\linewidth]{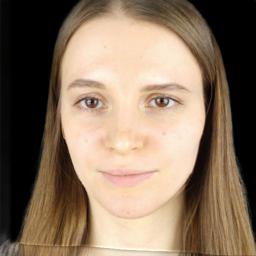}&
    \includegraphics[width=0.12\linewidth]{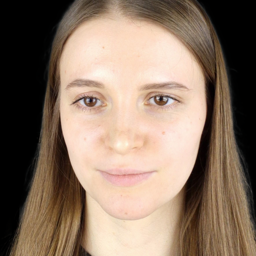} \\
    \vspace{-0.07cm}
    \includegraphics[width=0.12\linewidth]{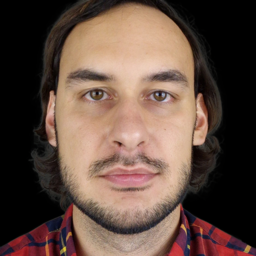}&
    \includegraphics[width=0.12\linewidth]{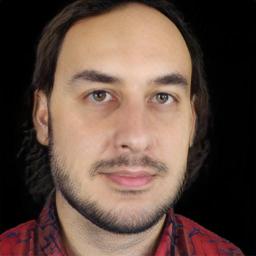}&
    \includegraphics[width=0.12\linewidth]{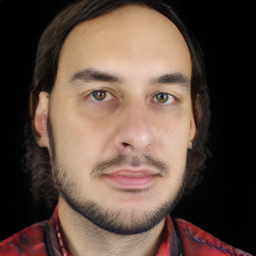}& 
    \includegraphics[width=0.12\linewidth]{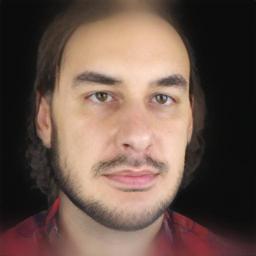}&
    \includegraphics[width=0.12\linewidth]{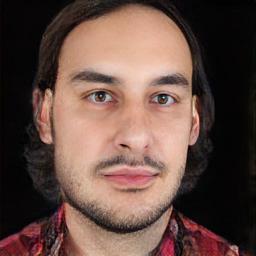} &
    \includegraphics[width=0.12\linewidth]{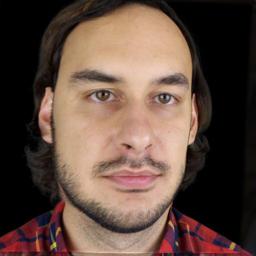}&
    \includegraphics[width=0.12\linewidth]{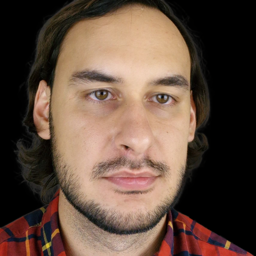}\\
    \vspace{-0.07cm}
    \includegraphics[width=0.12\linewidth]{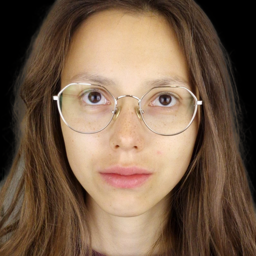}&
    \includegraphics[width=0.12\linewidth]{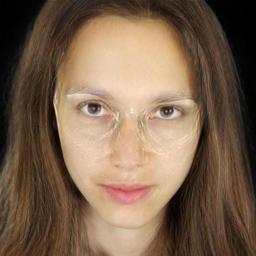}&
    \includegraphics[width=0.12\linewidth]{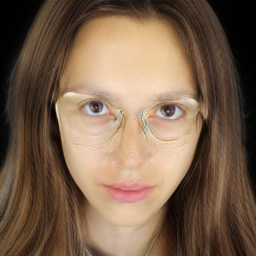}& 
    \includegraphics[width=0.12\linewidth]{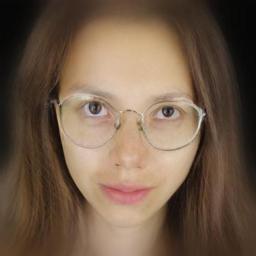}&
    \includegraphics[width=0.12\linewidth]{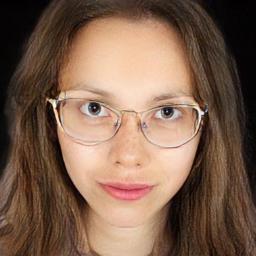} &
    \includegraphics[width=0.12\linewidth]{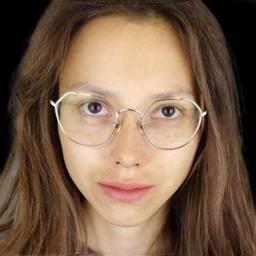}&
    \includegraphics[width=0.12\linewidth]{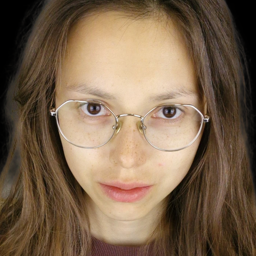}\\
  \end{tabular}
  \vspace{-0.2cm}
  \caption{Head pose correction of HeRo images. Photos from a Source (Front camera) are transformed into a Target (Left, Right, or Top view in each row respectively).}
  \label{fig:results-hero}
\end{figure*}

\subsection{Face Undistortion}
\label{ssec:face-unfistortion}

We present the evaluation of \ours{} against competing face undistortion methods in Table \ref{tab:results-undistortion}. Evidently, our method notably outperforms others in terms of all metrics, especially in identity preservation. Due to the sampling of pixels from the original image, \ours{} keeps crucial details of identity, such as eye color, wrinkles, earrings, etc.

\begin{figure}[ht!]
  \centering
  \begin{tabular}{*{5}{@{\hspace{2pt}}c}}
    Input & Fried's \cite{fried2016perspective} & Ko's \cite{ko20223d} & DisCO \cite{wang2023disco} & \textbf{\ours}\\
    \vspace{-0.07cm}
    \includegraphics[width=0.19\linewidth]{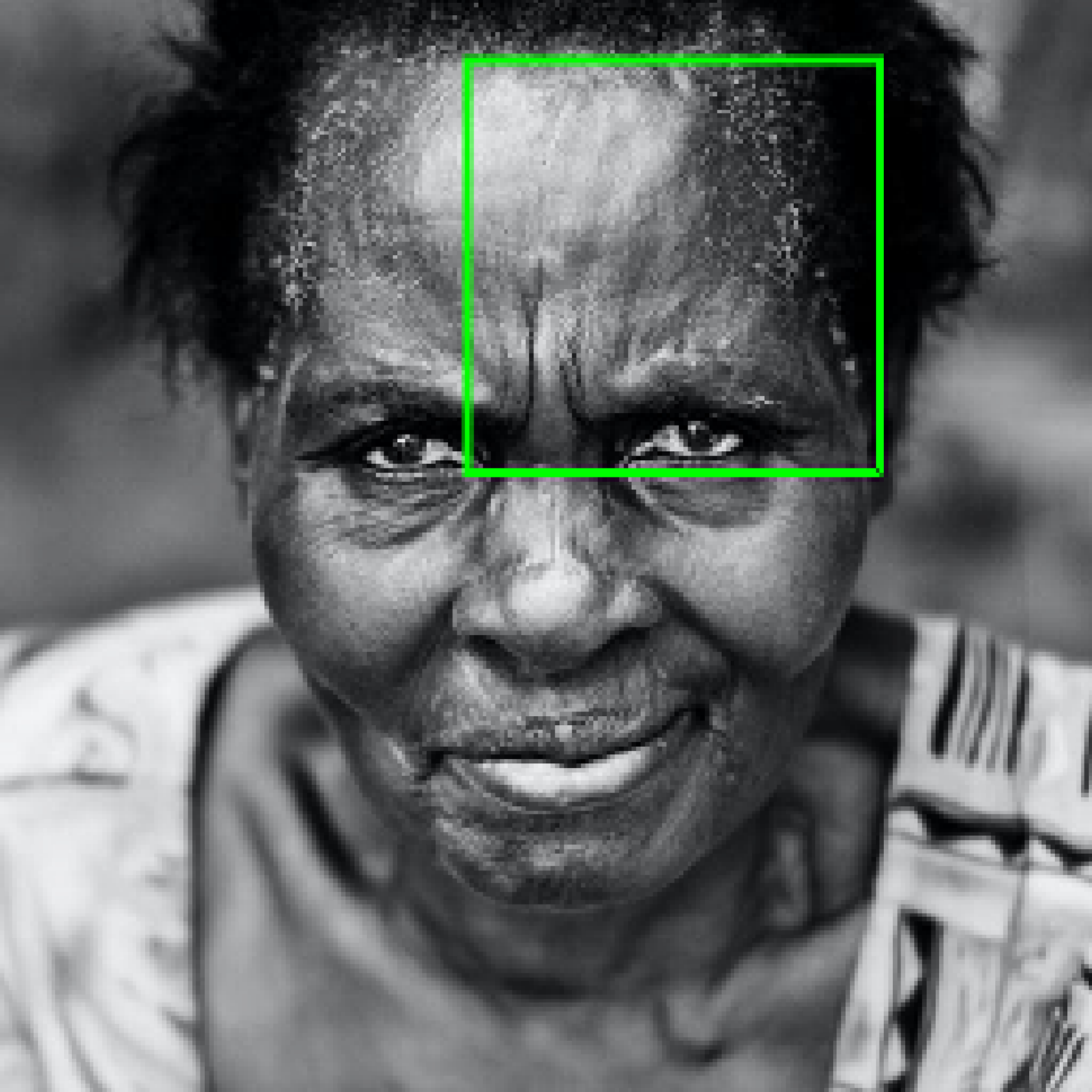}&
    \includegraphics[width=0.19\linewidth]{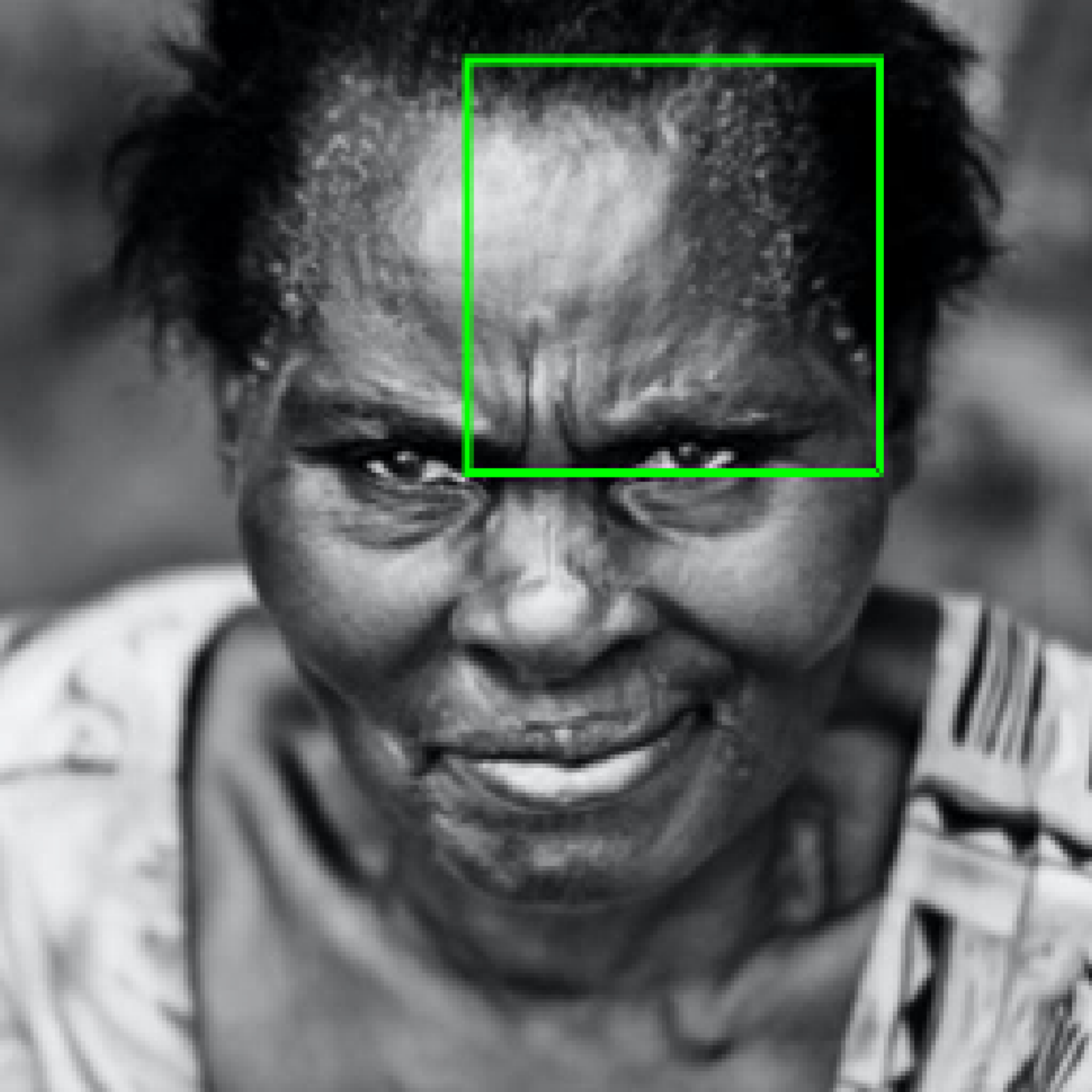}&
    \includegraphics[width=0.19\linewidth]{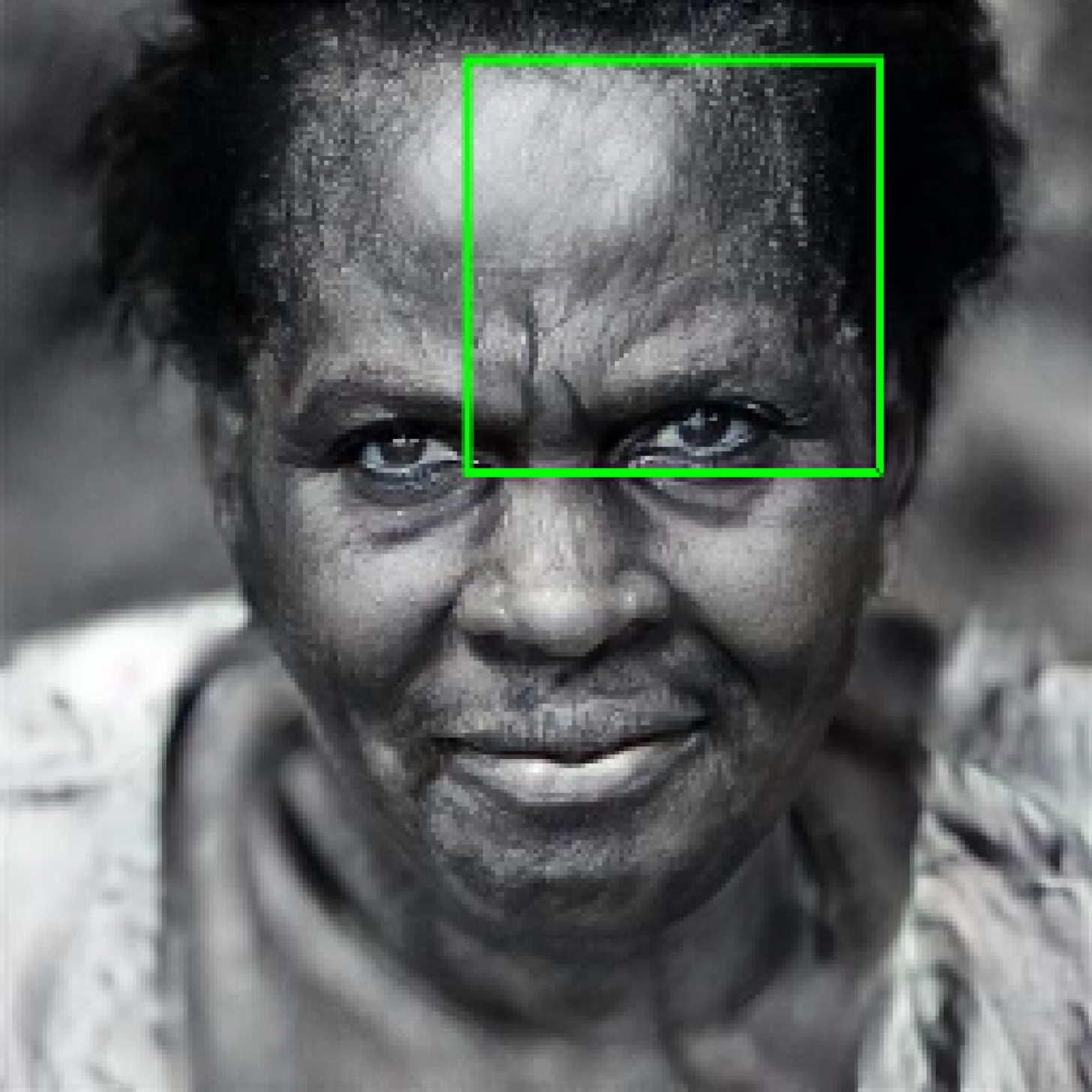}& 
    \includegraphics[width=0.19\linewidth]{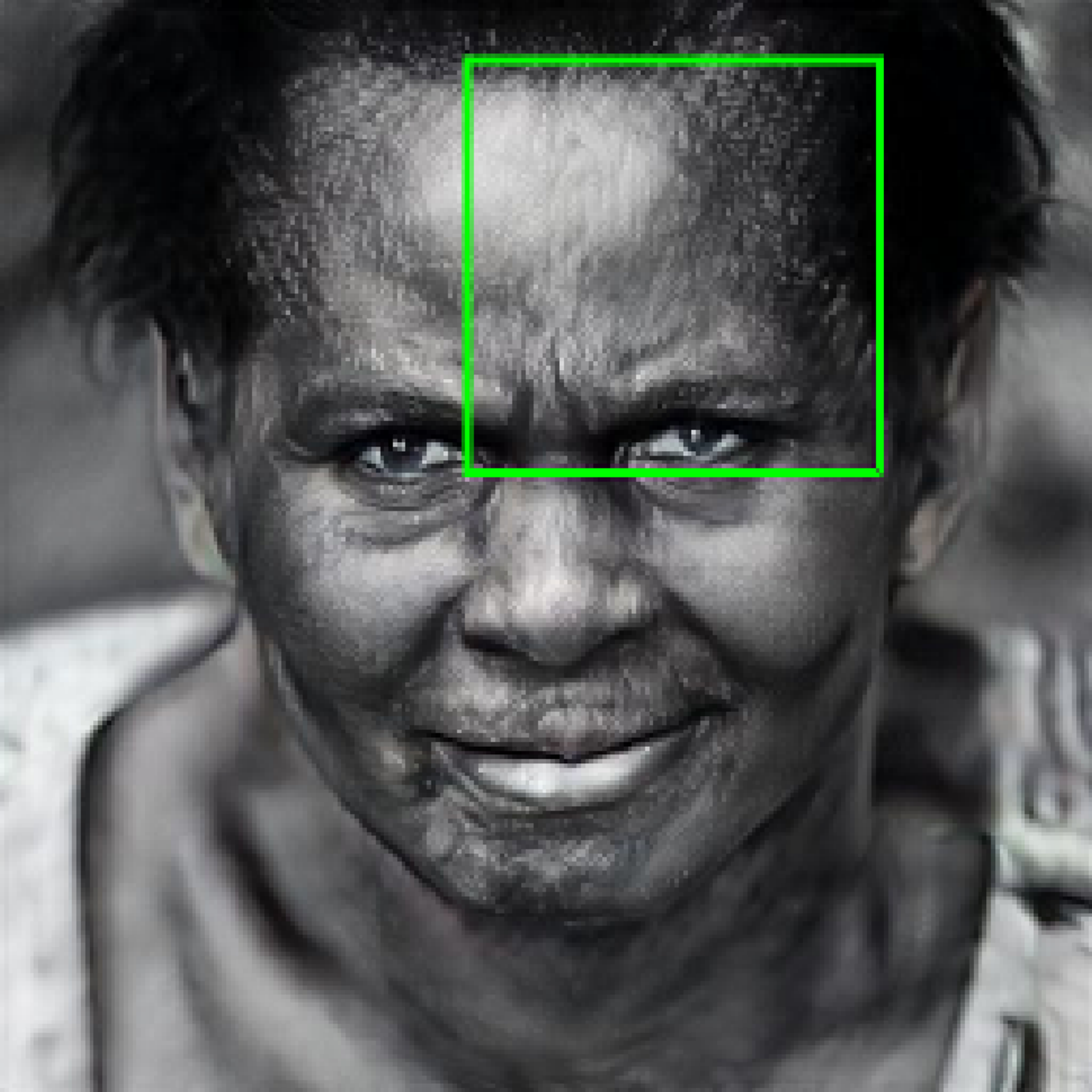}&
    \includegraphics[width=0.19\linewidth]{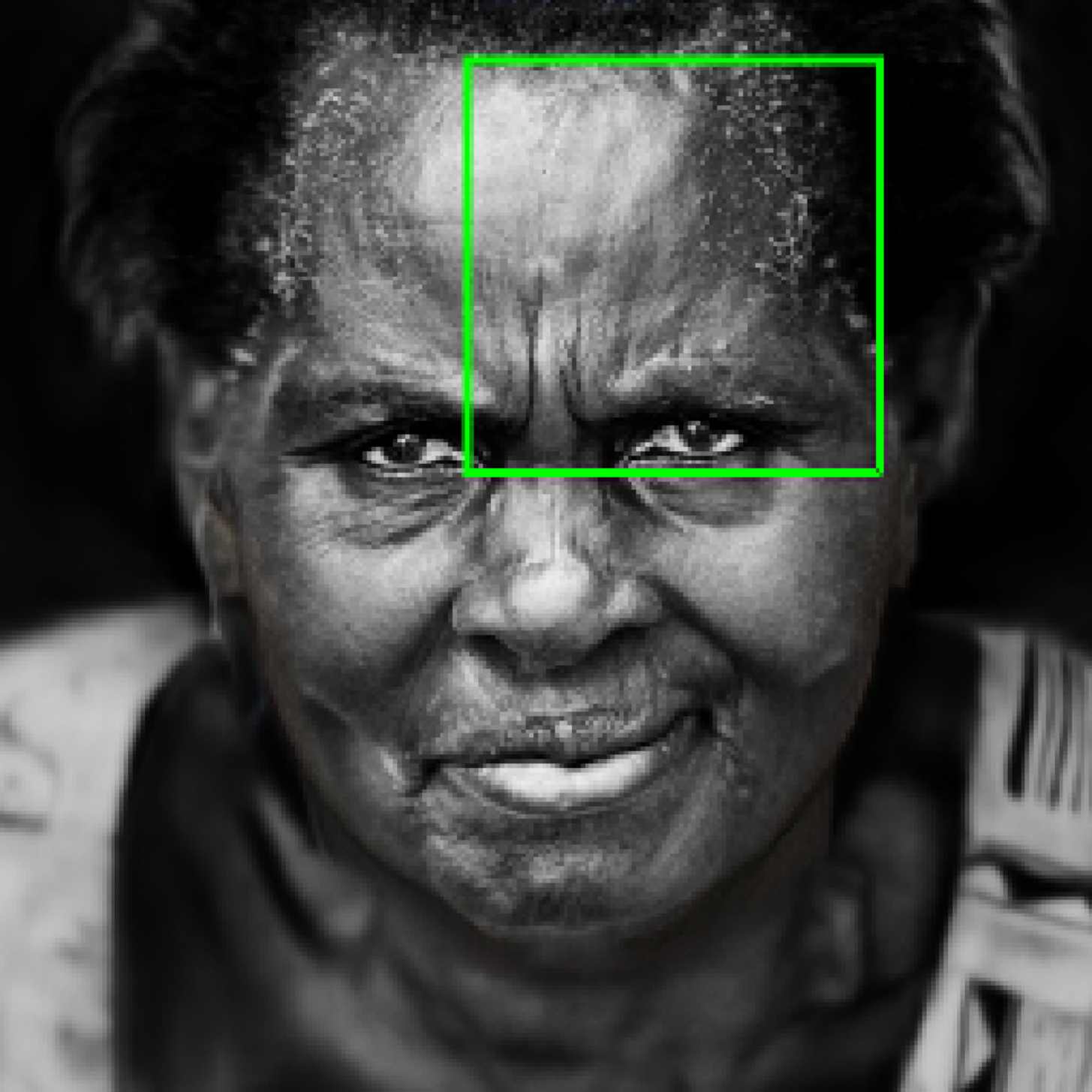}\\
    \vspace{-0.07cm}
    \includegraphics[width=0.19\linewidth]{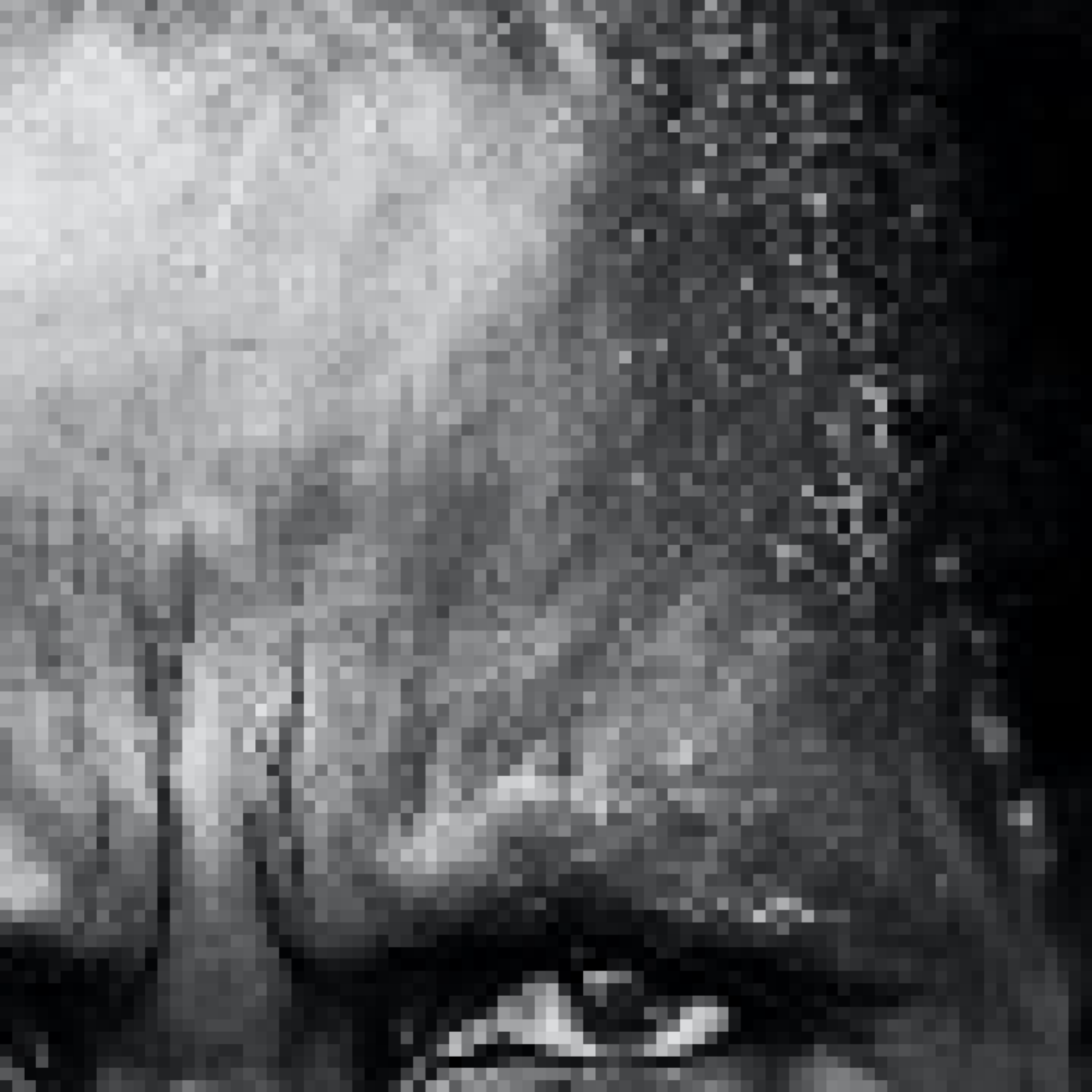}&
    \includegraphics[width=0.19\linewidth]{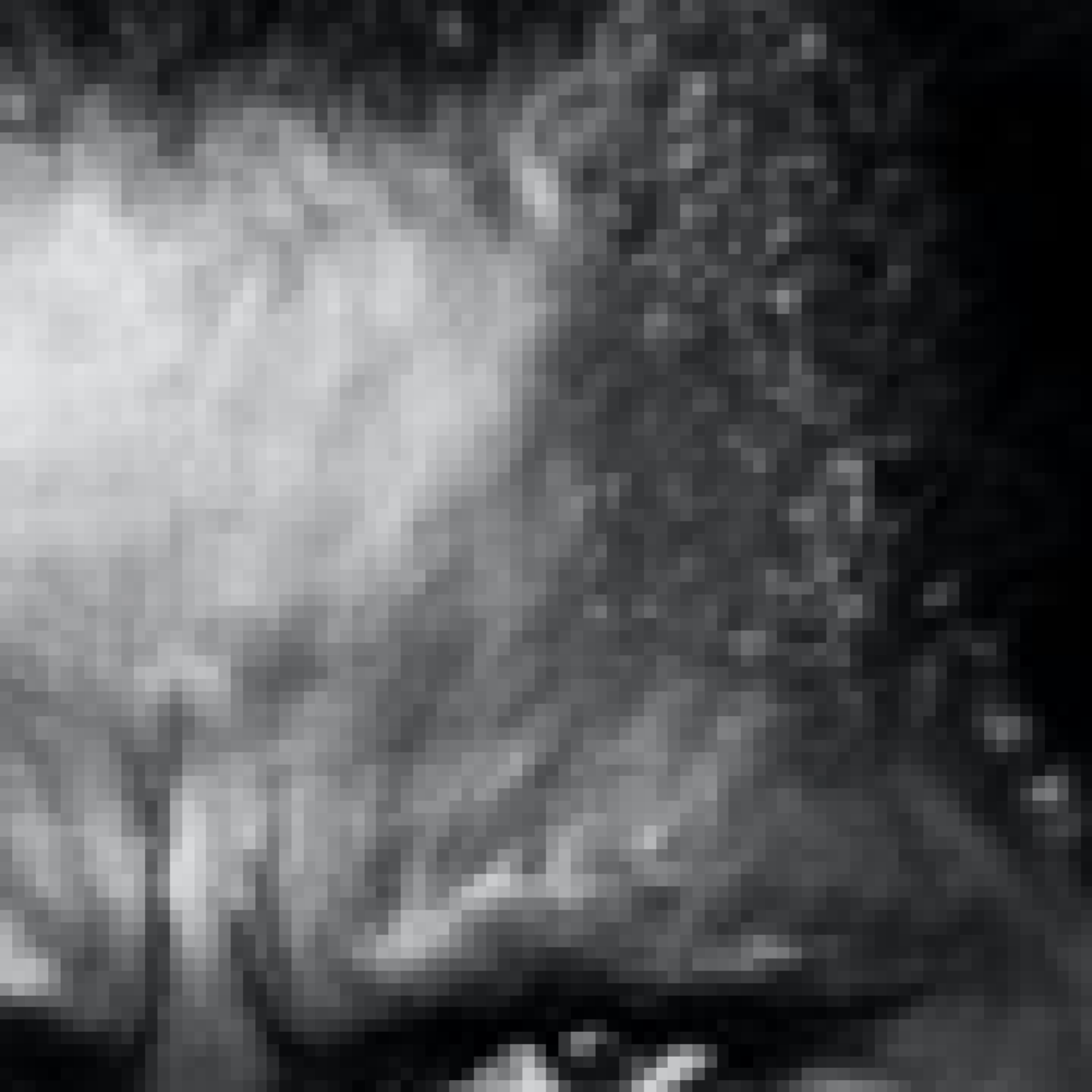}&
    \includegraphics[width=0.19\linewidth]{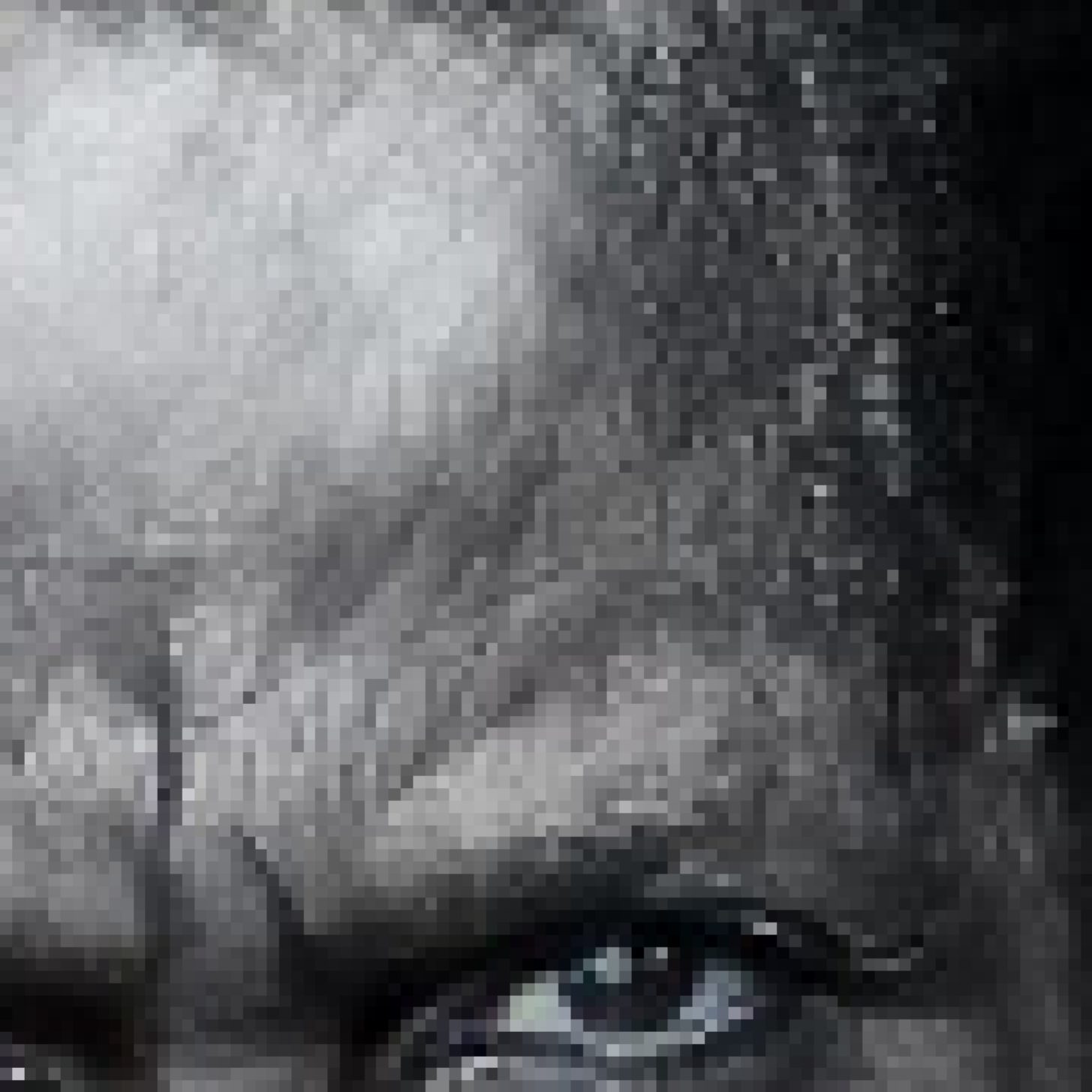}& 
    \includegraphics[width=0.19\linewidth]{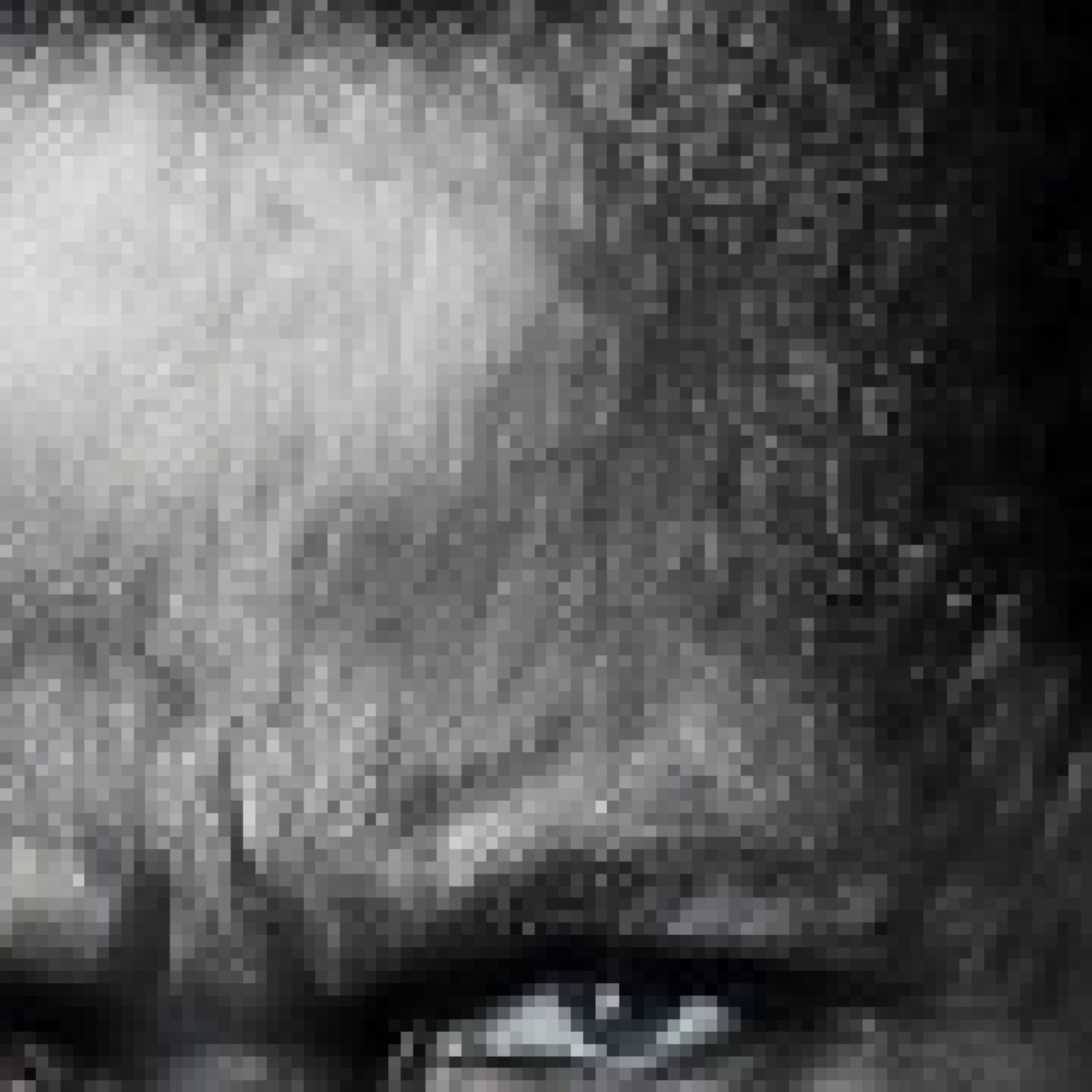}&
    \includegraphics[width=0.19\linewidth]{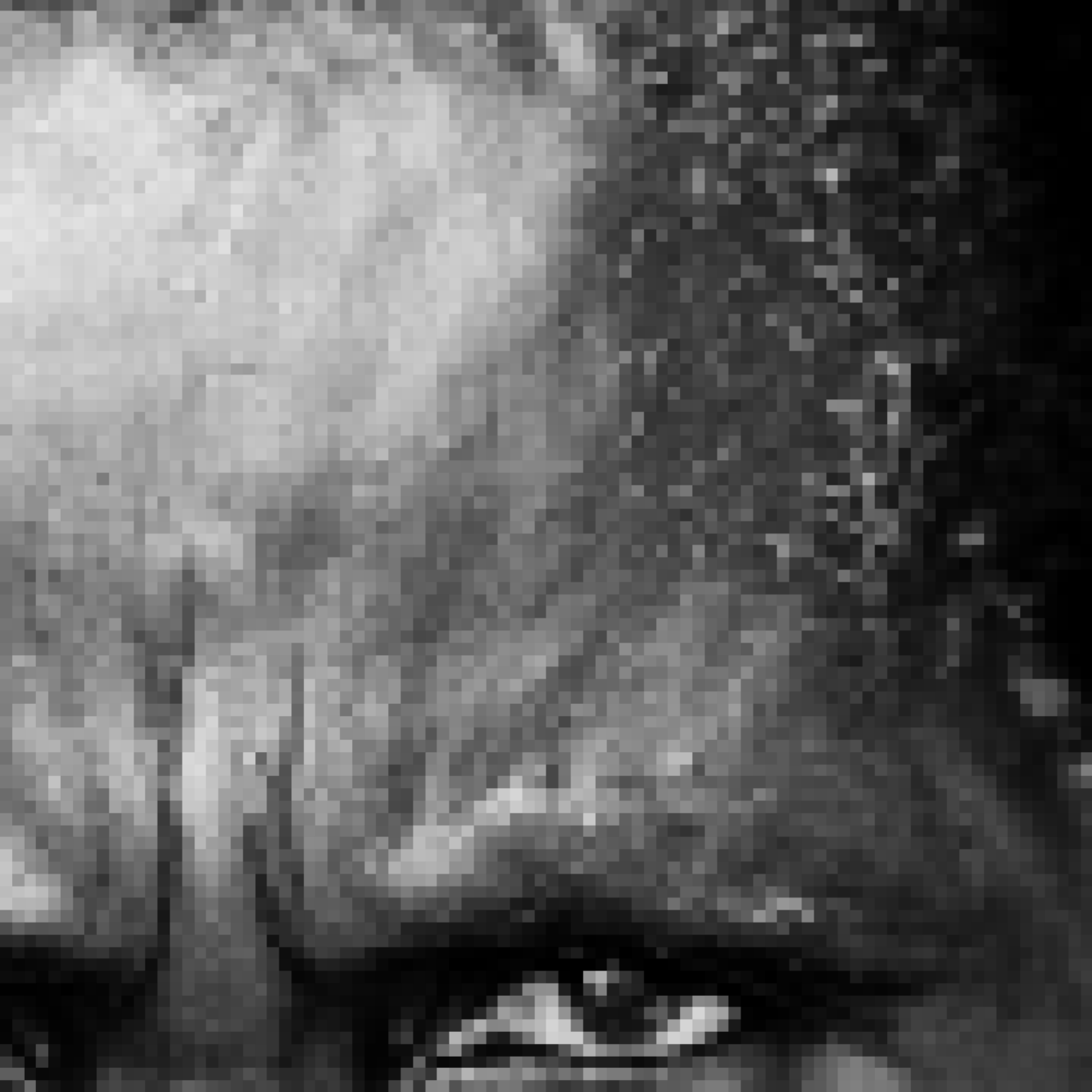}\\
    \vspace{-0.07cm}
    \includegraphics[width=0.19\linewidth]{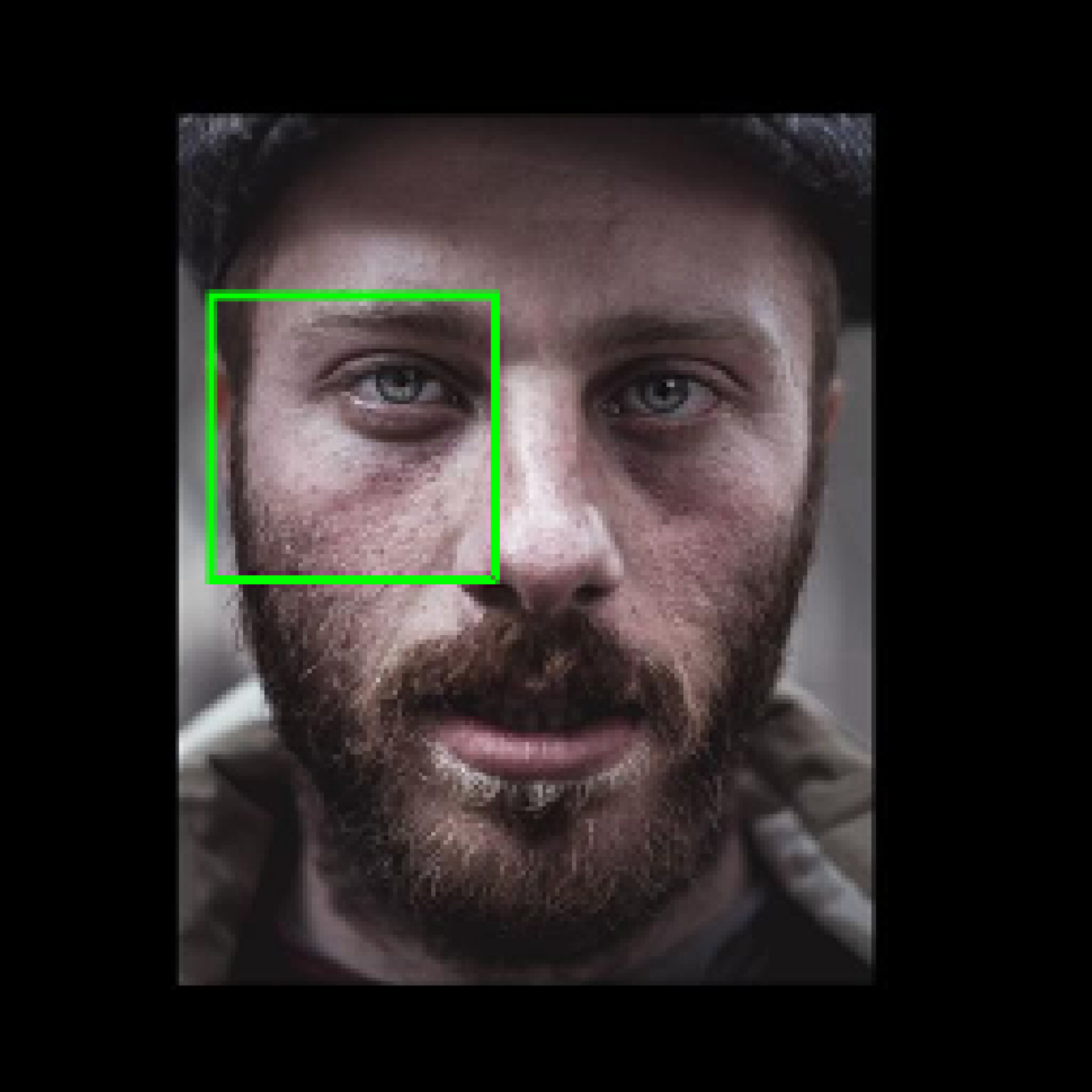}&
    \includegraphics[width=0.19\linewidth]{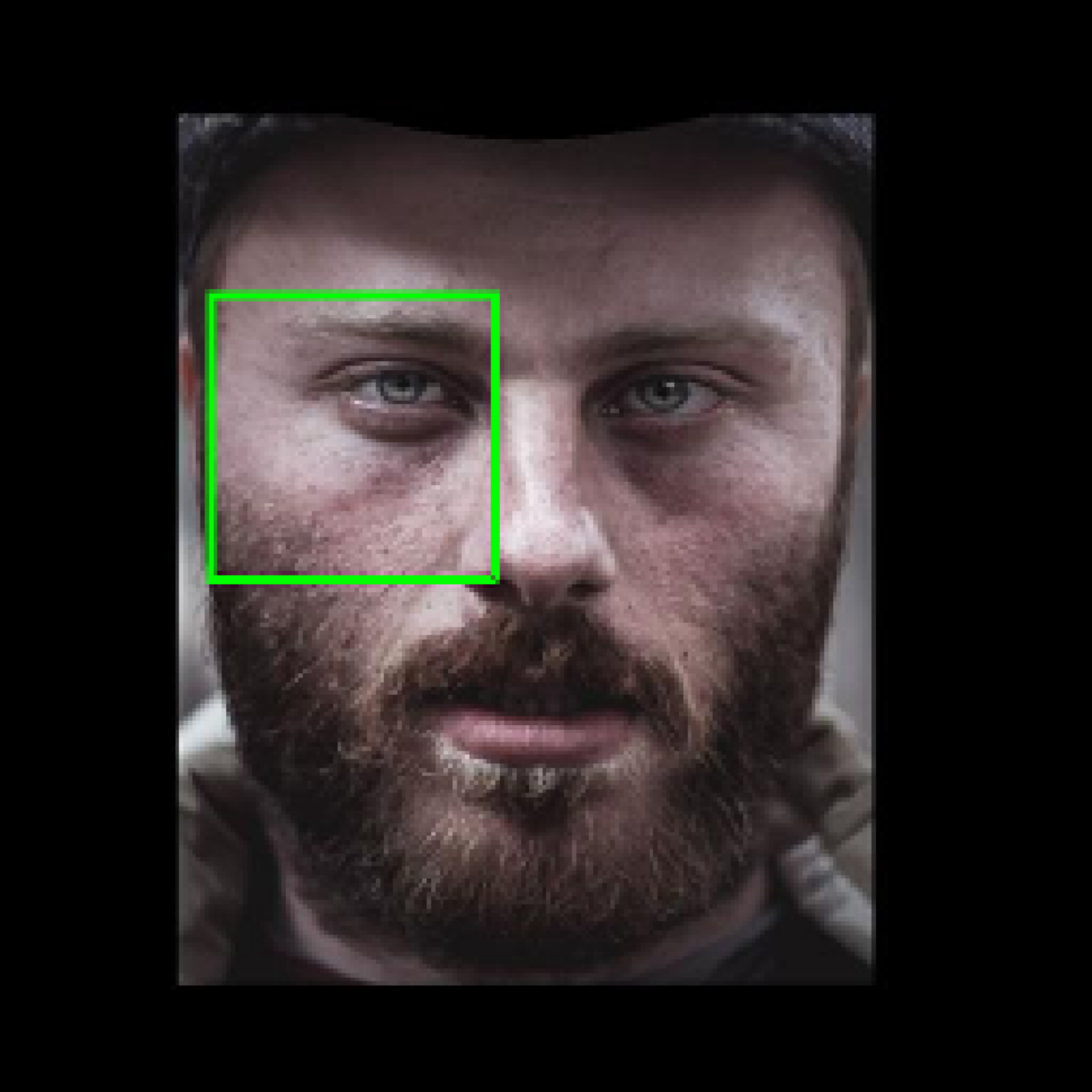}&
    \includegraphics[width=0.19\linewidth]{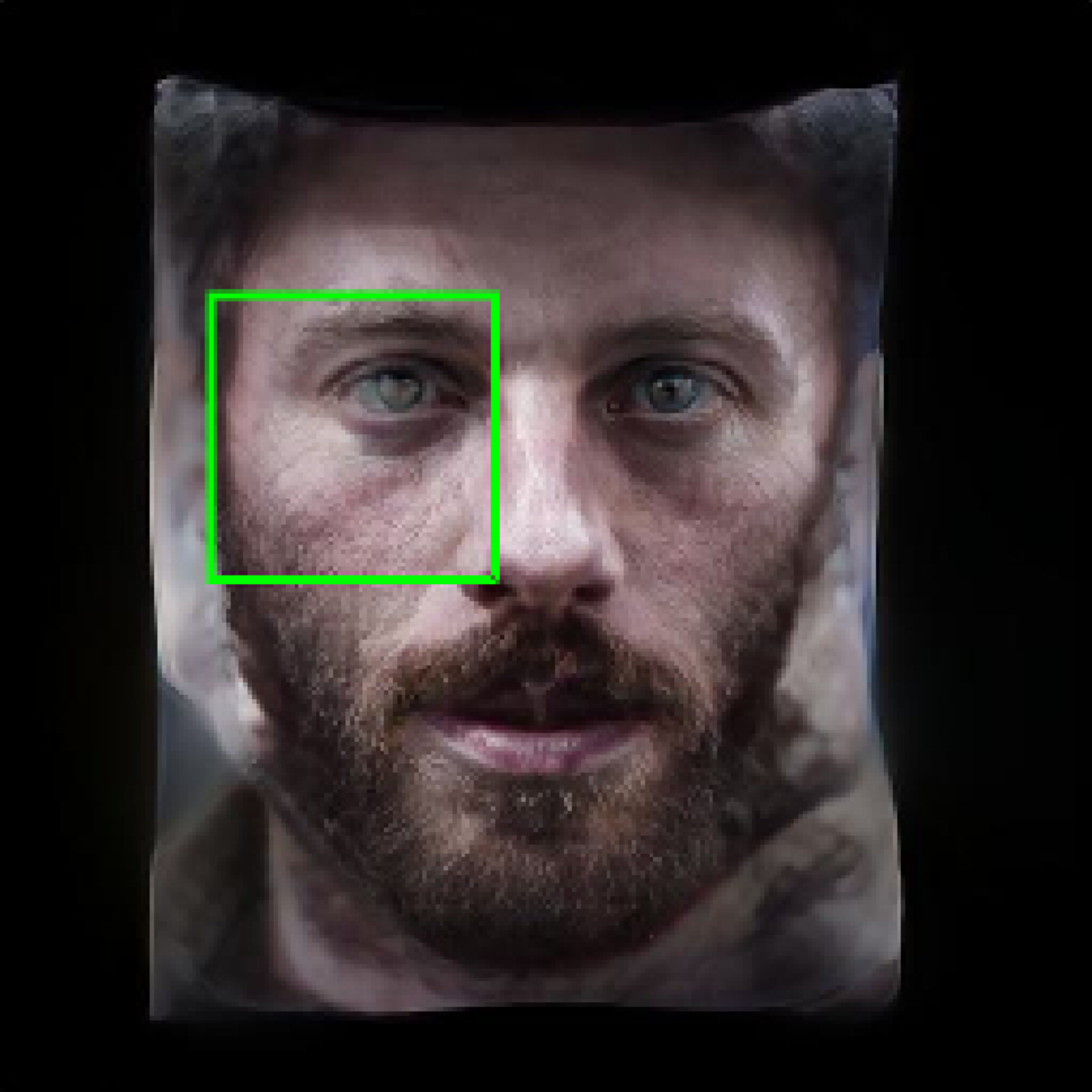}& 
    \includegraphics[width=0.19\linewidth]{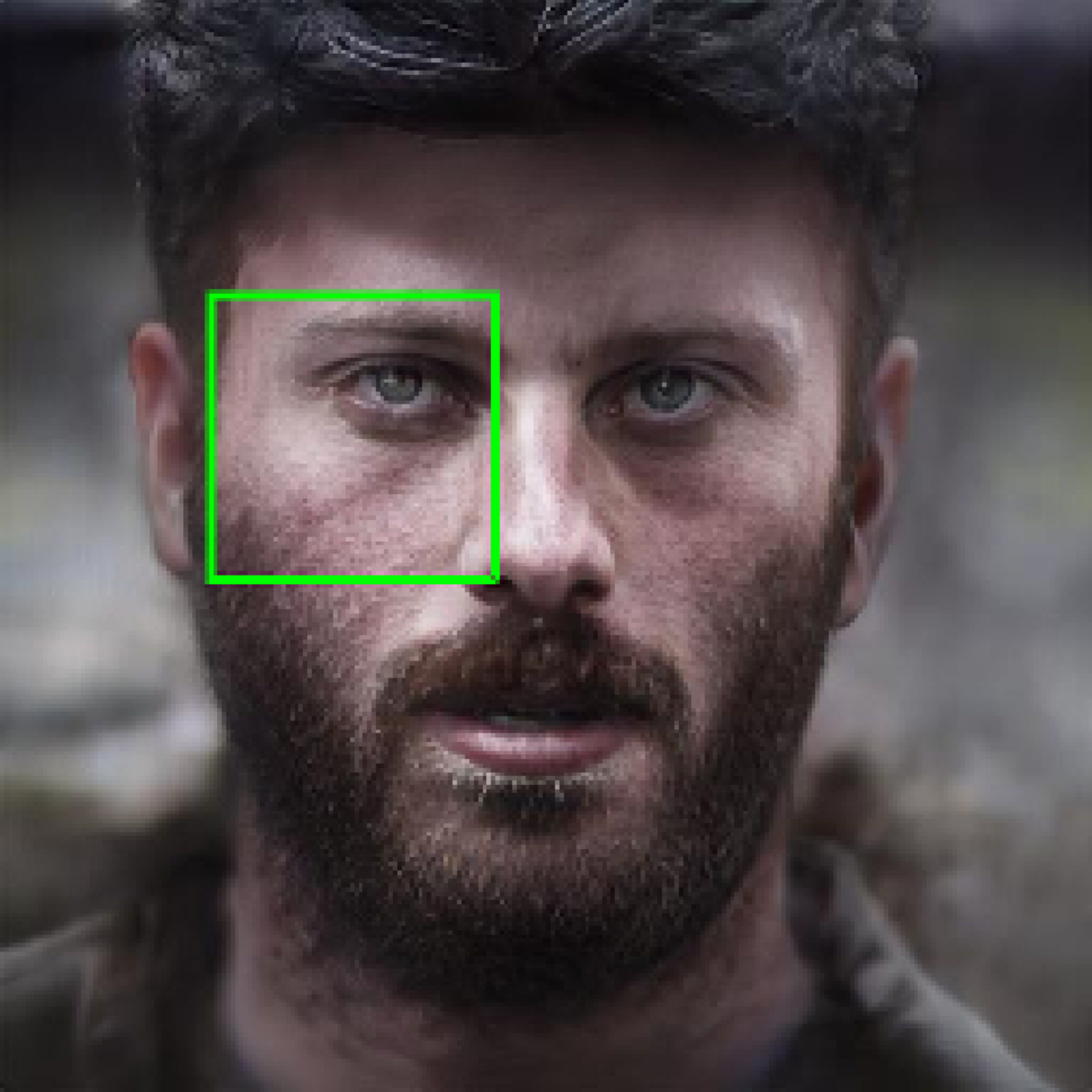}&
    \includegraphics[width=0.19\linewidth]{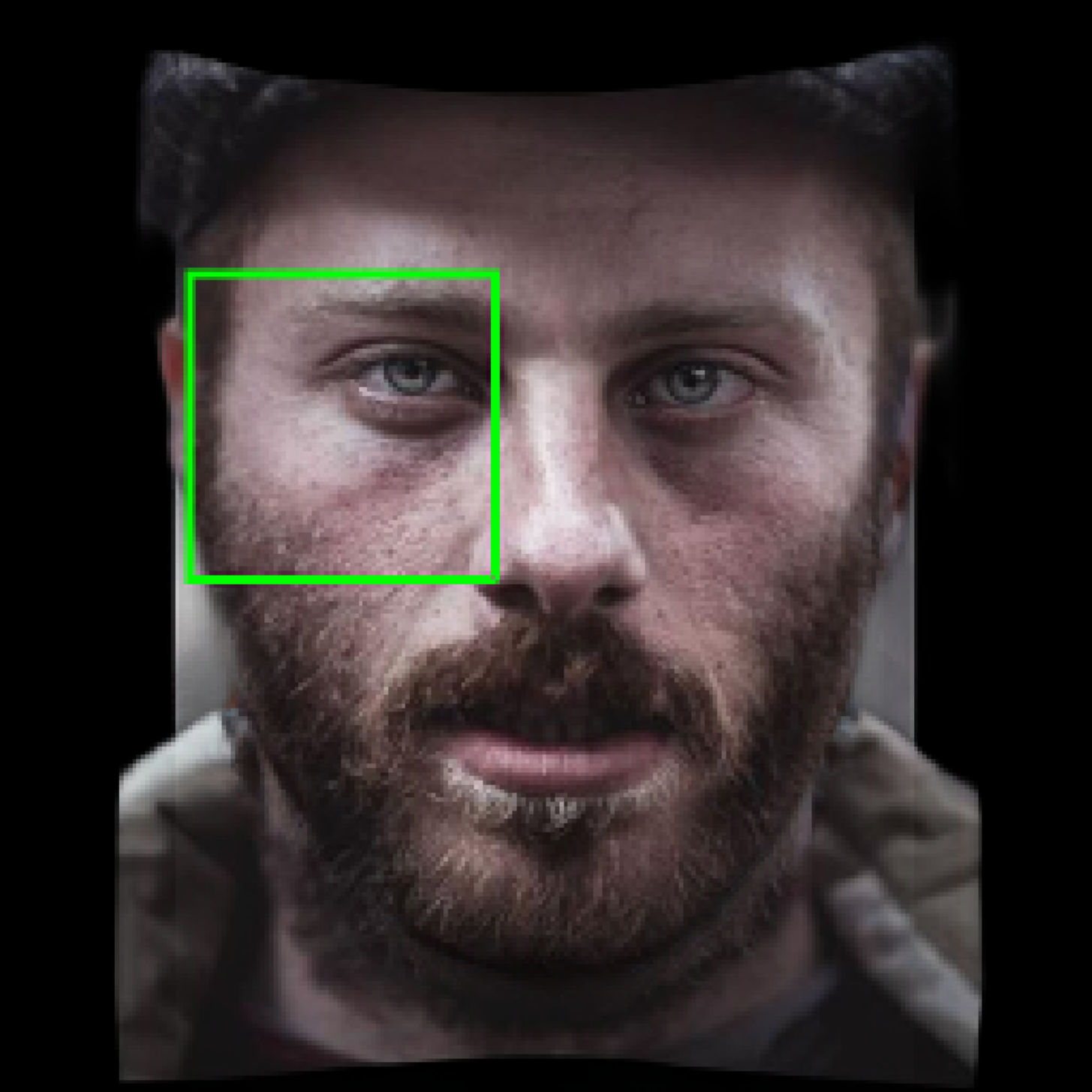}\\
    \vspace{-0.07cm}
    \includegraphics[width=0.19\linewidth]{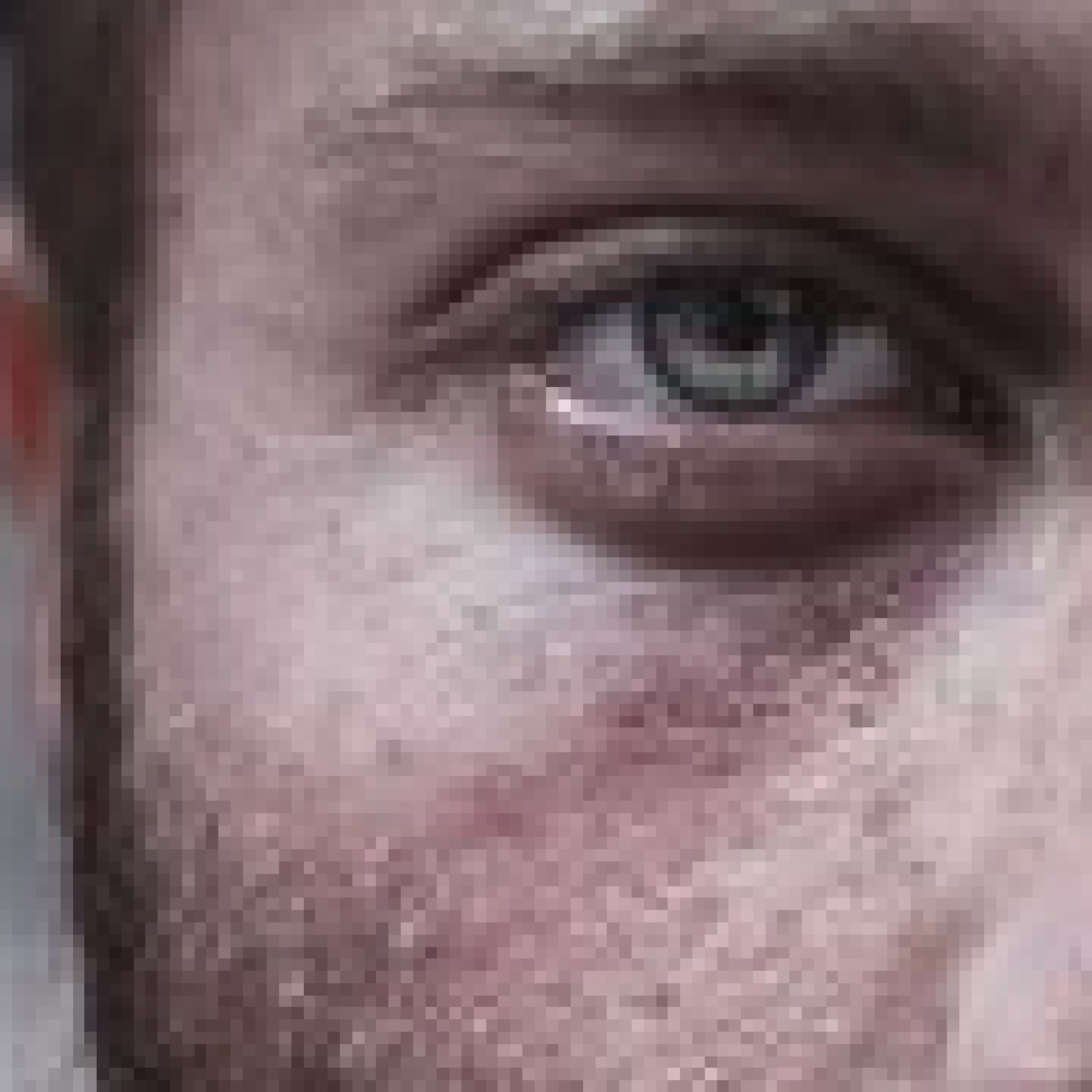}&
    \includegraphics[width=0.19\linewidth]{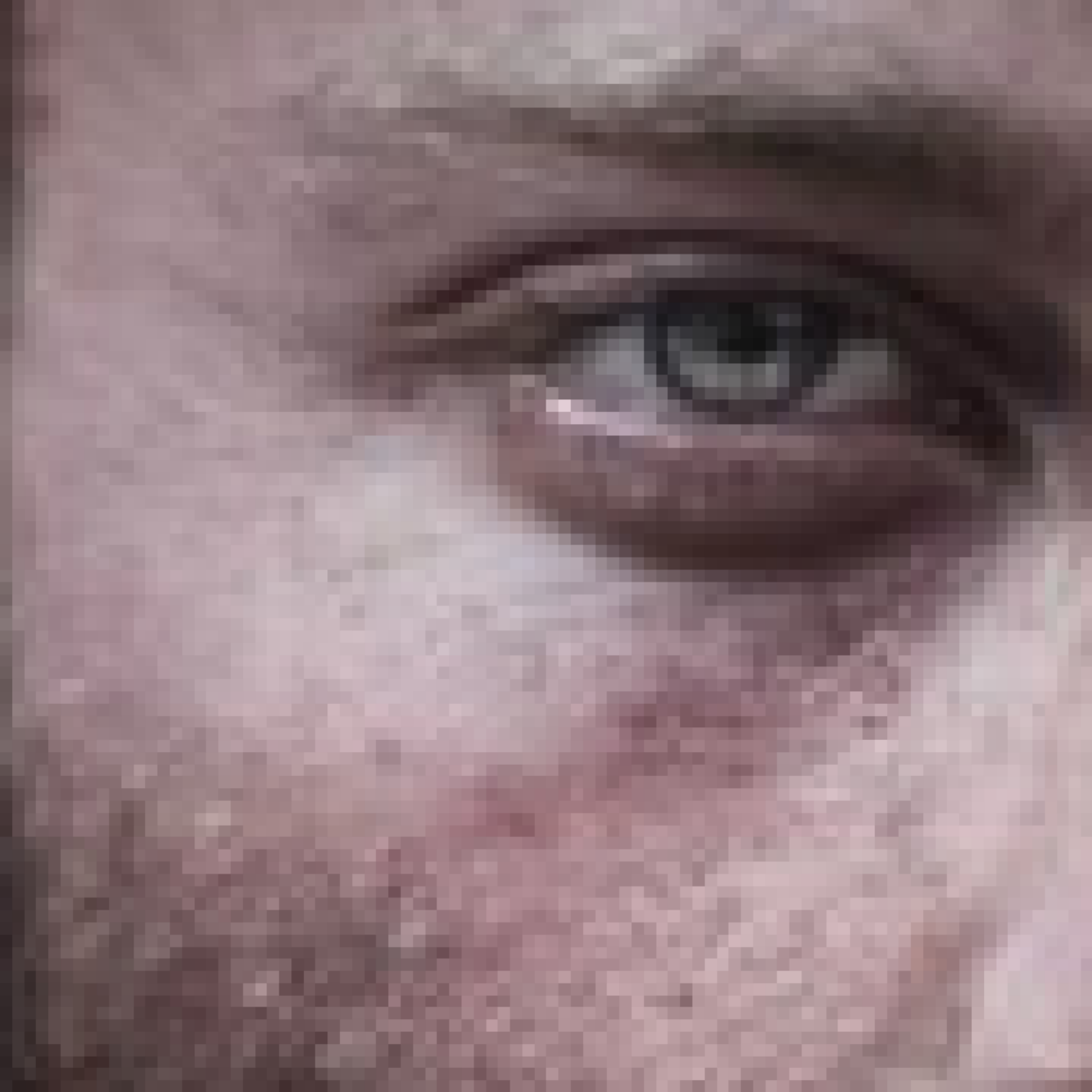}&
    \includegraphics[width=0.19\linewidth]{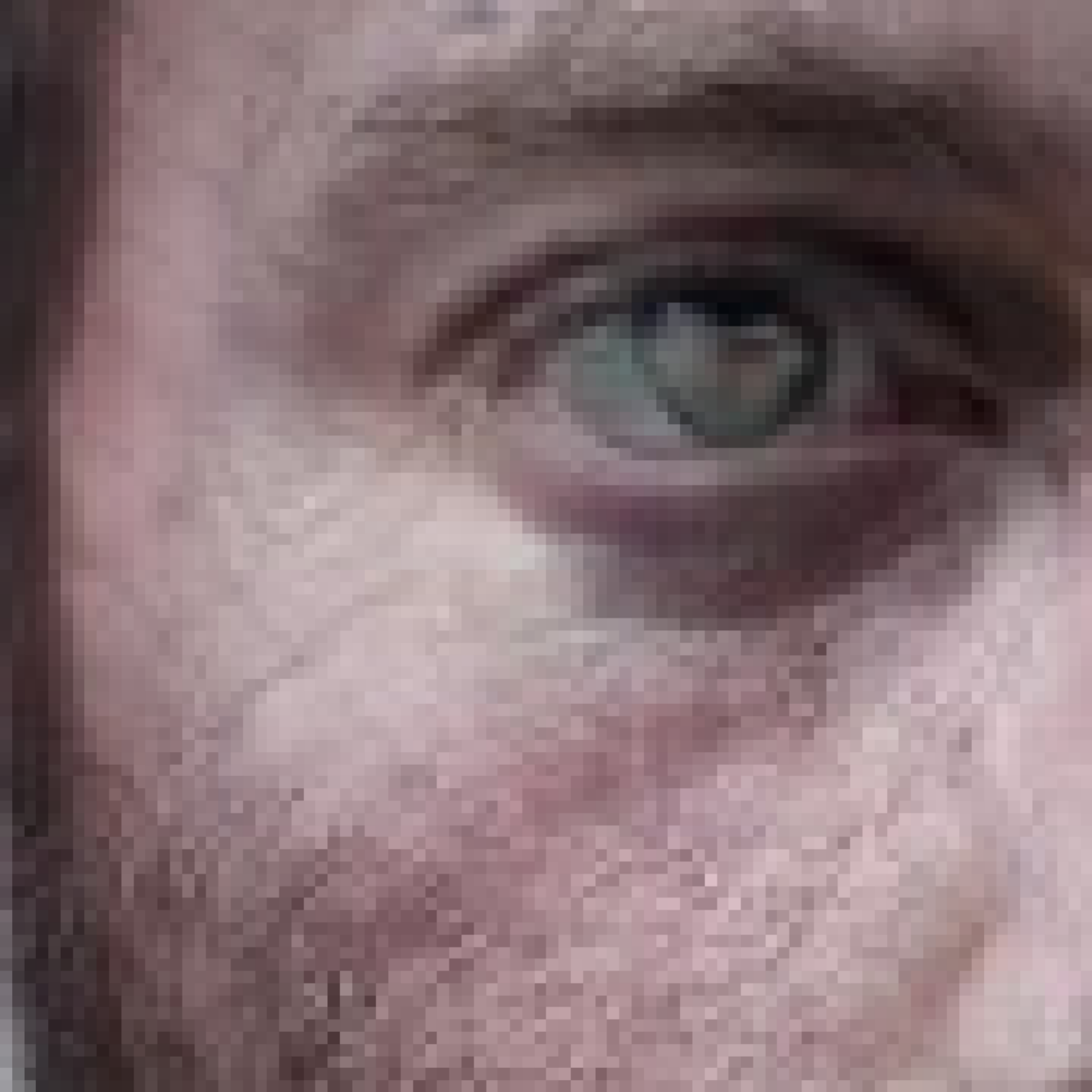}& 
    \includegraphics[width=0.19\linewidth]{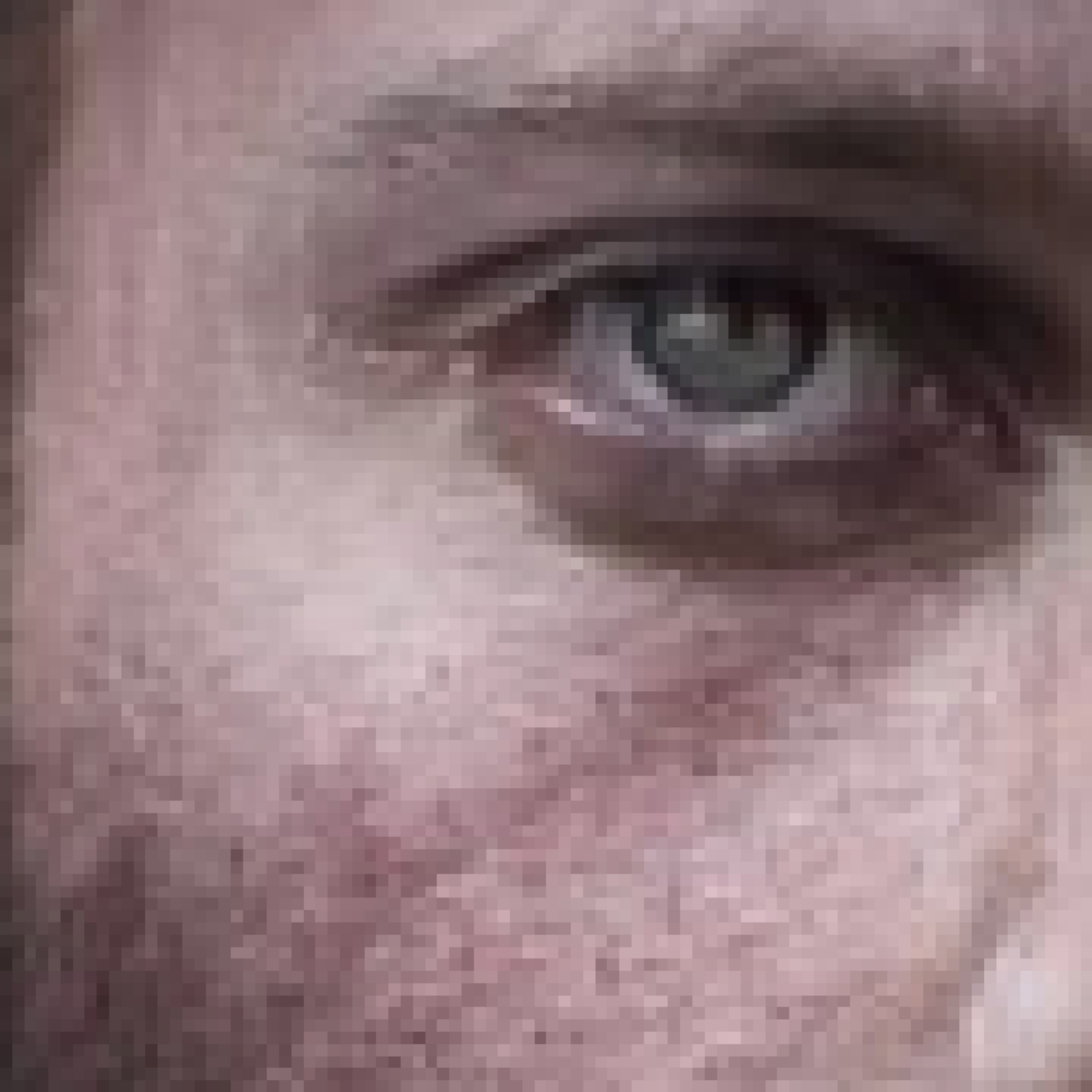}&
    \includegraphics[width=0.19\linewidth]{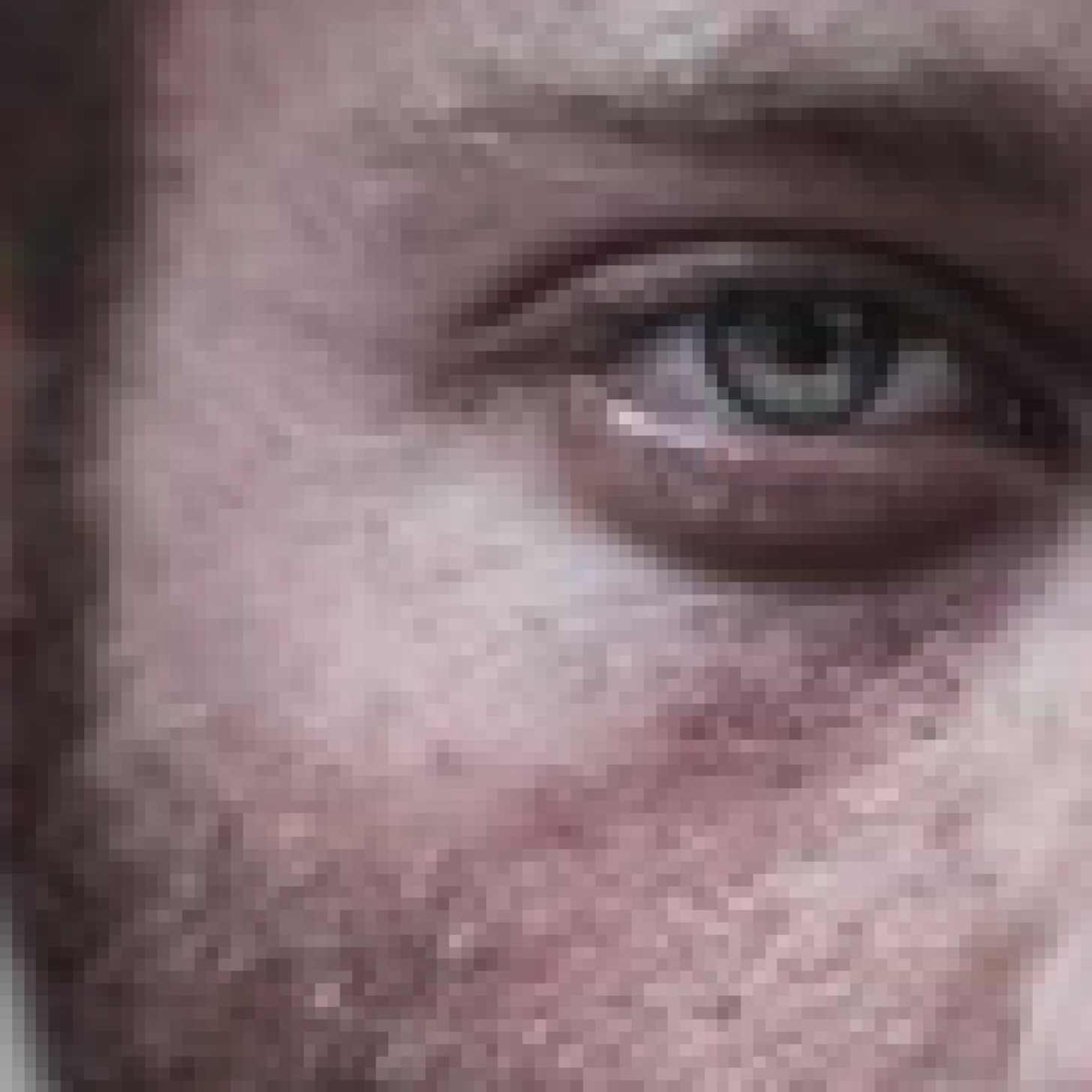}\\
    \vspace{-0.07cm}
    \includegraphics[width=0.19\linewidth]{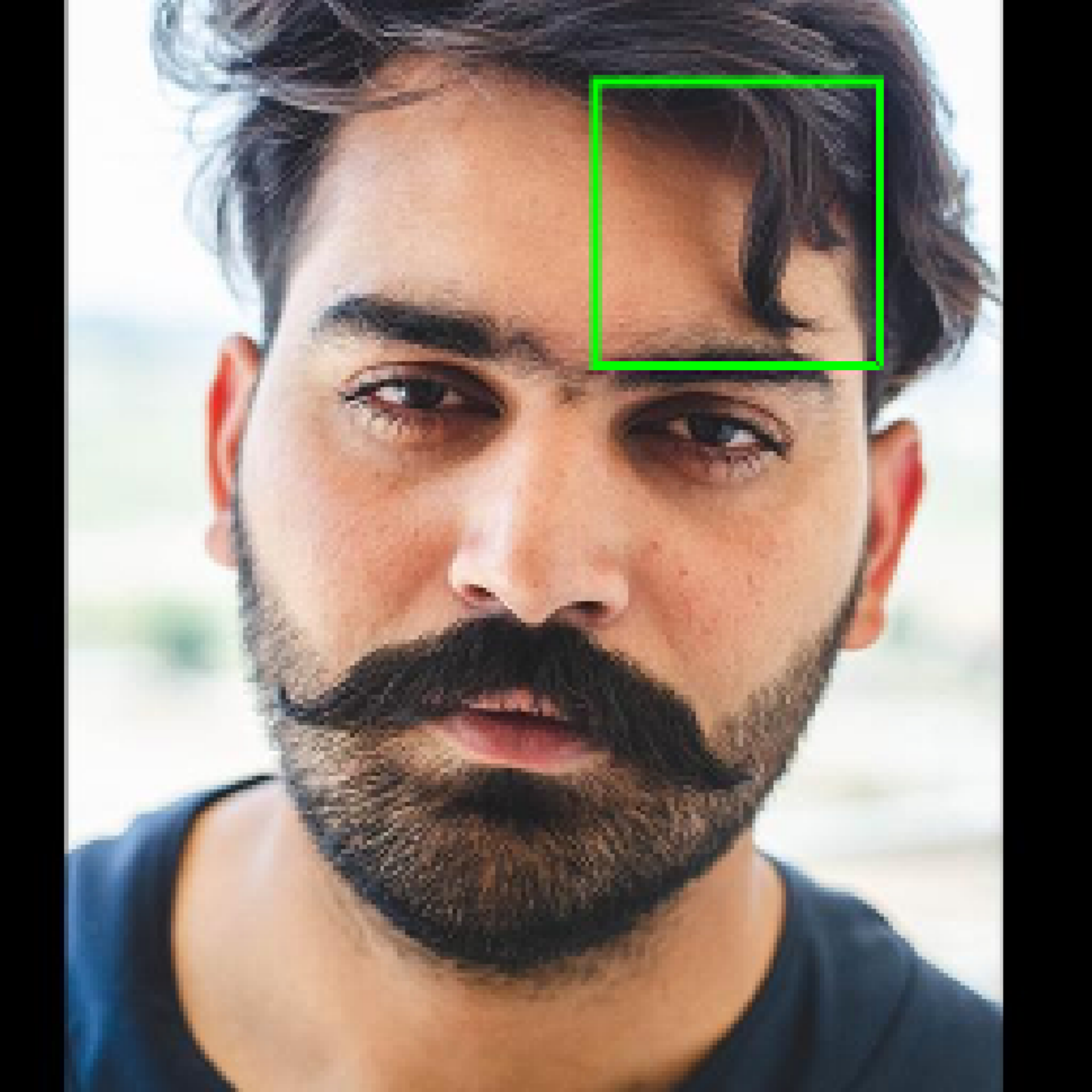}&
    \includegraphics[width=0.19\linewidth]{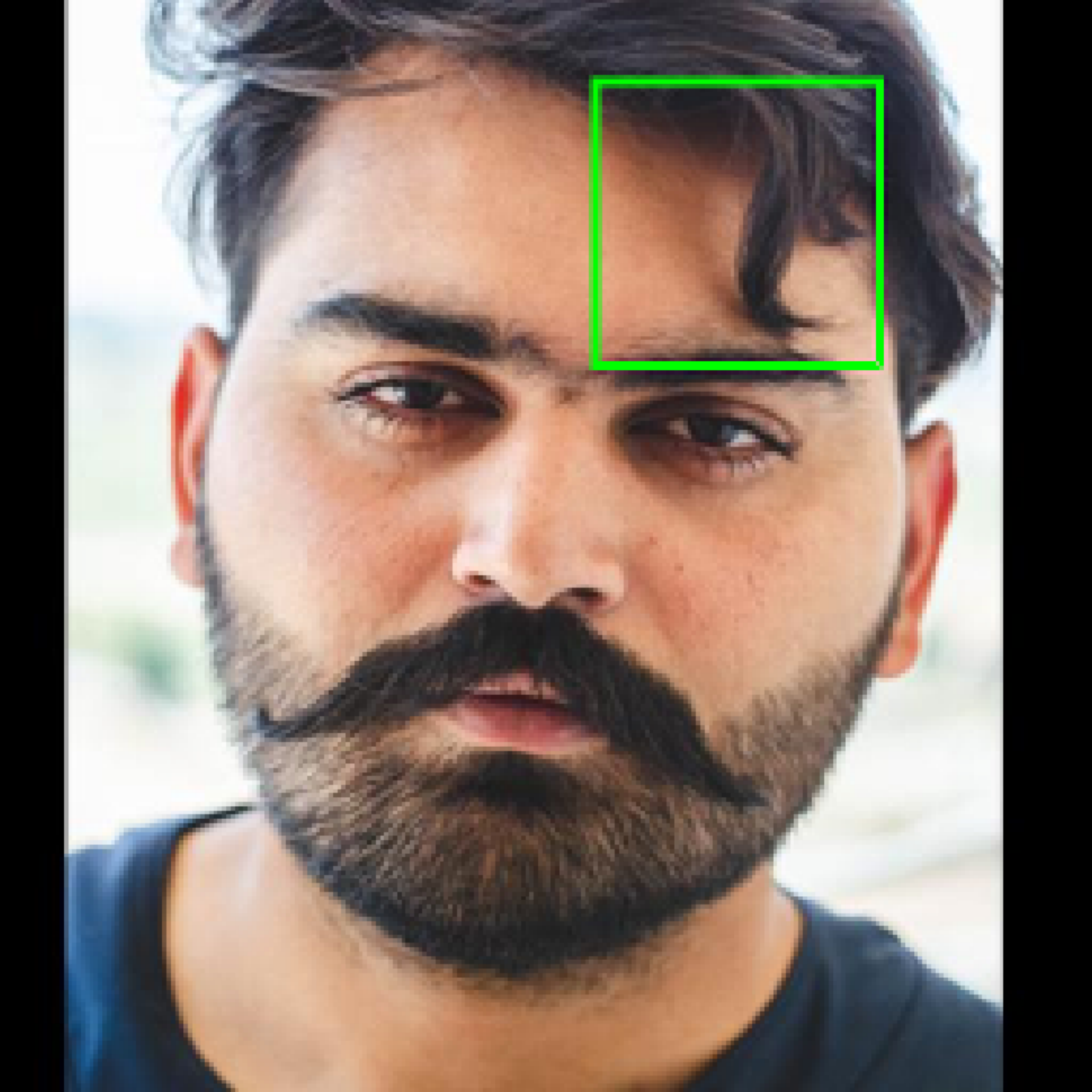}&
    \includegraphics[width=0.19\linewidth]{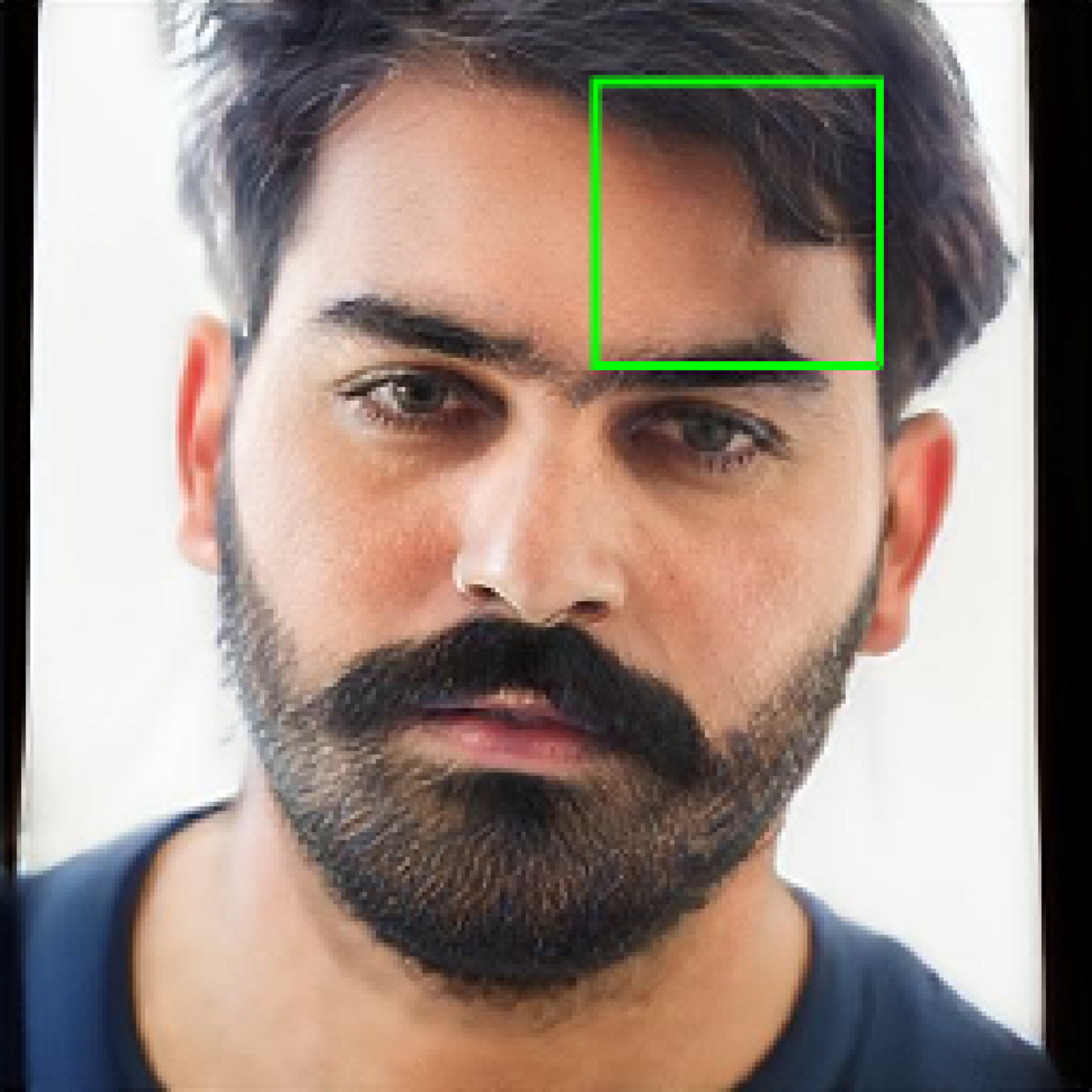}& 
    \includegraphics[width=0.19\linewidth]{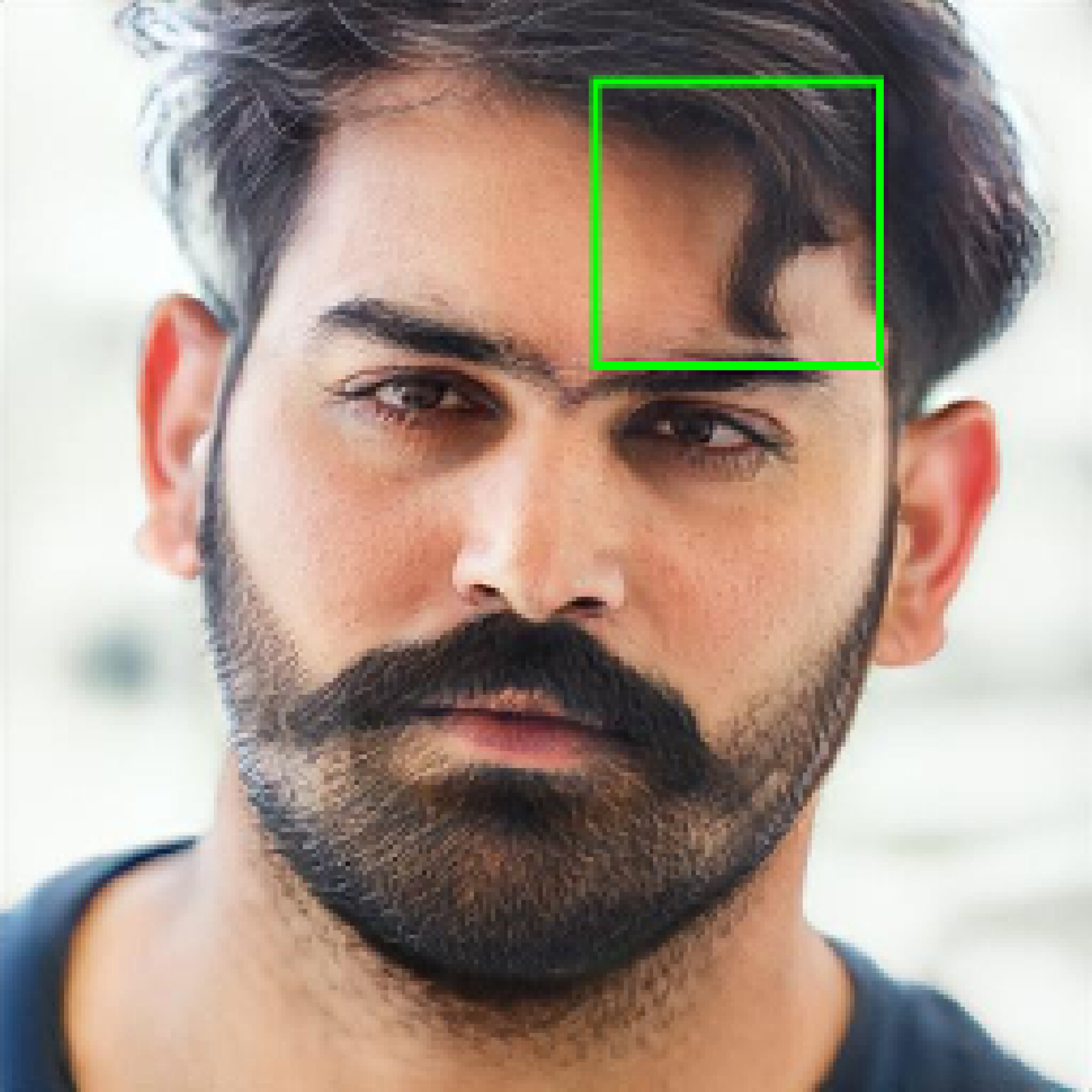}&
    \includegraphics[width=0.19\linewidth]{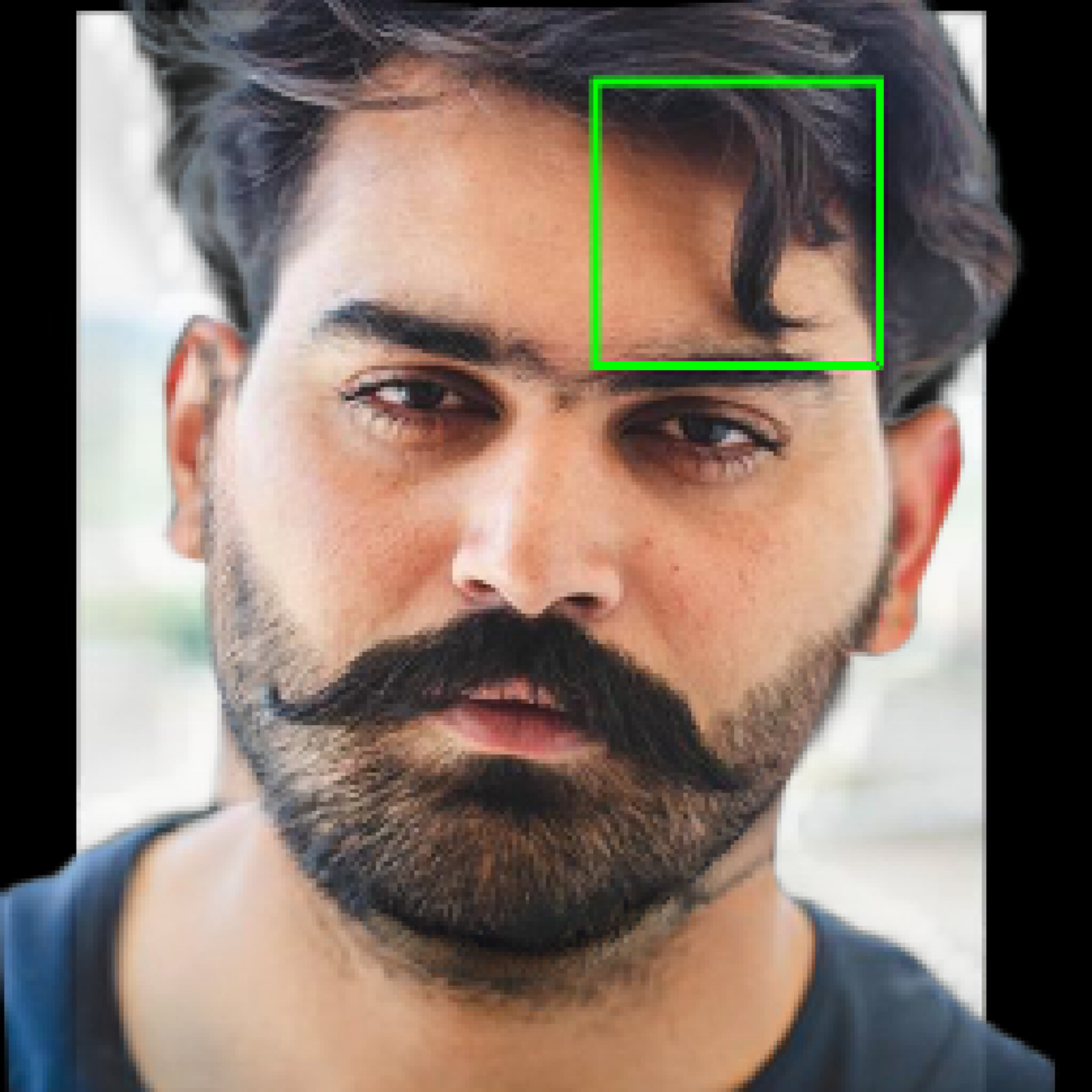}\\
    \vspace{-0.07cm}
    \includegraphics[width=0.19\linewidth]{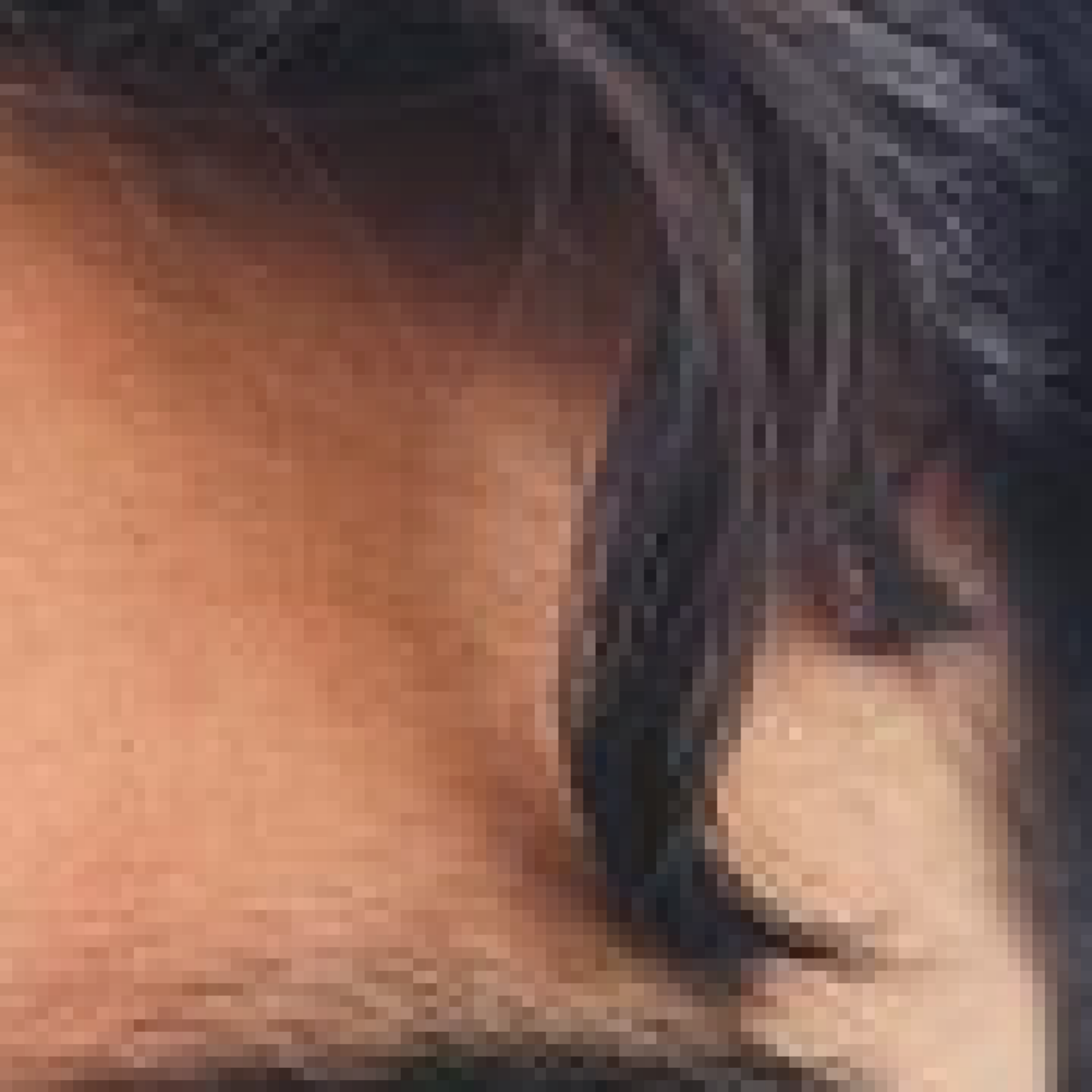}&
    \includegraphics[width=0.19\linewidth]{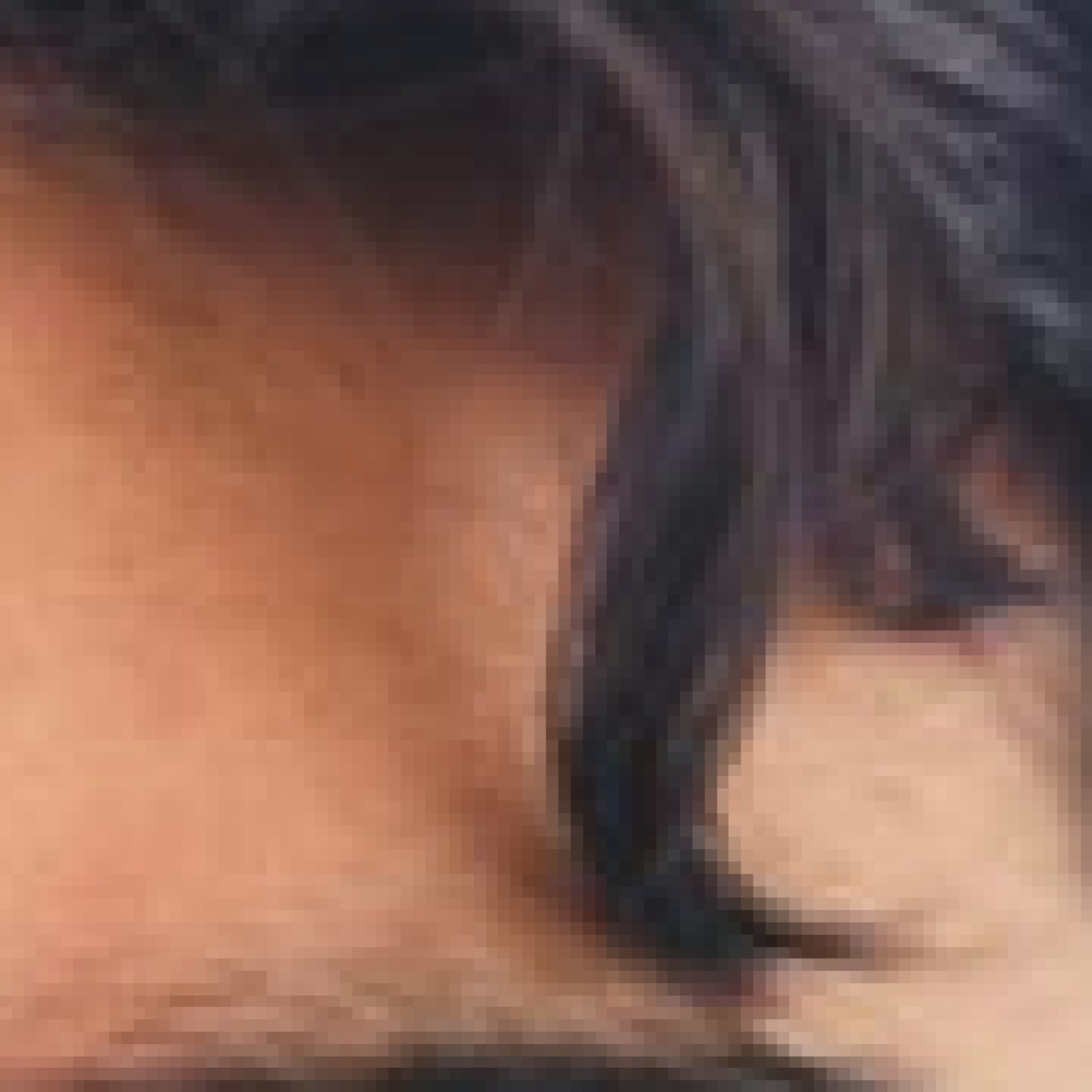}&
    \includegraphics[width=0.19\linewidth]{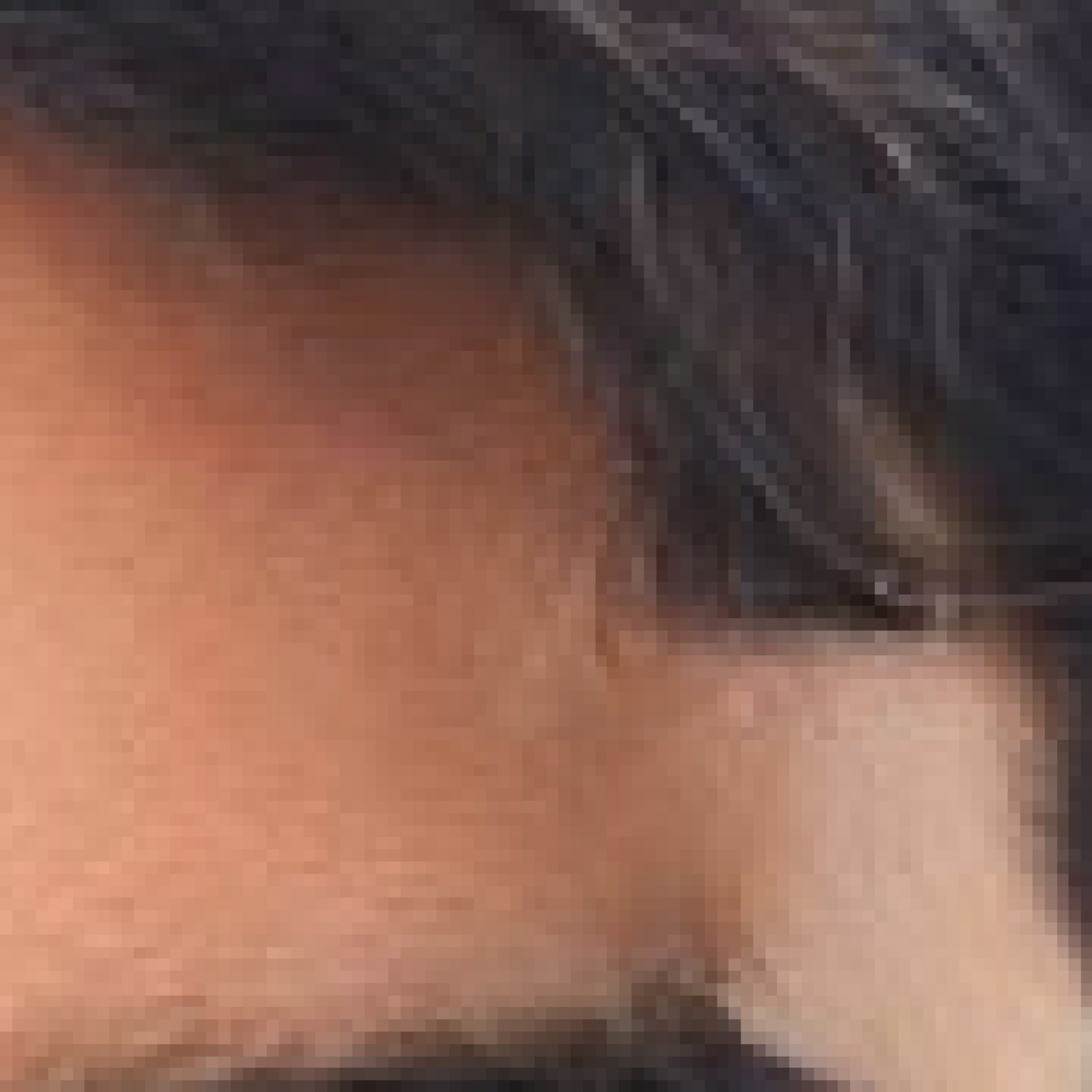}& 
    \includegraphics[width=0.19\linewidth]{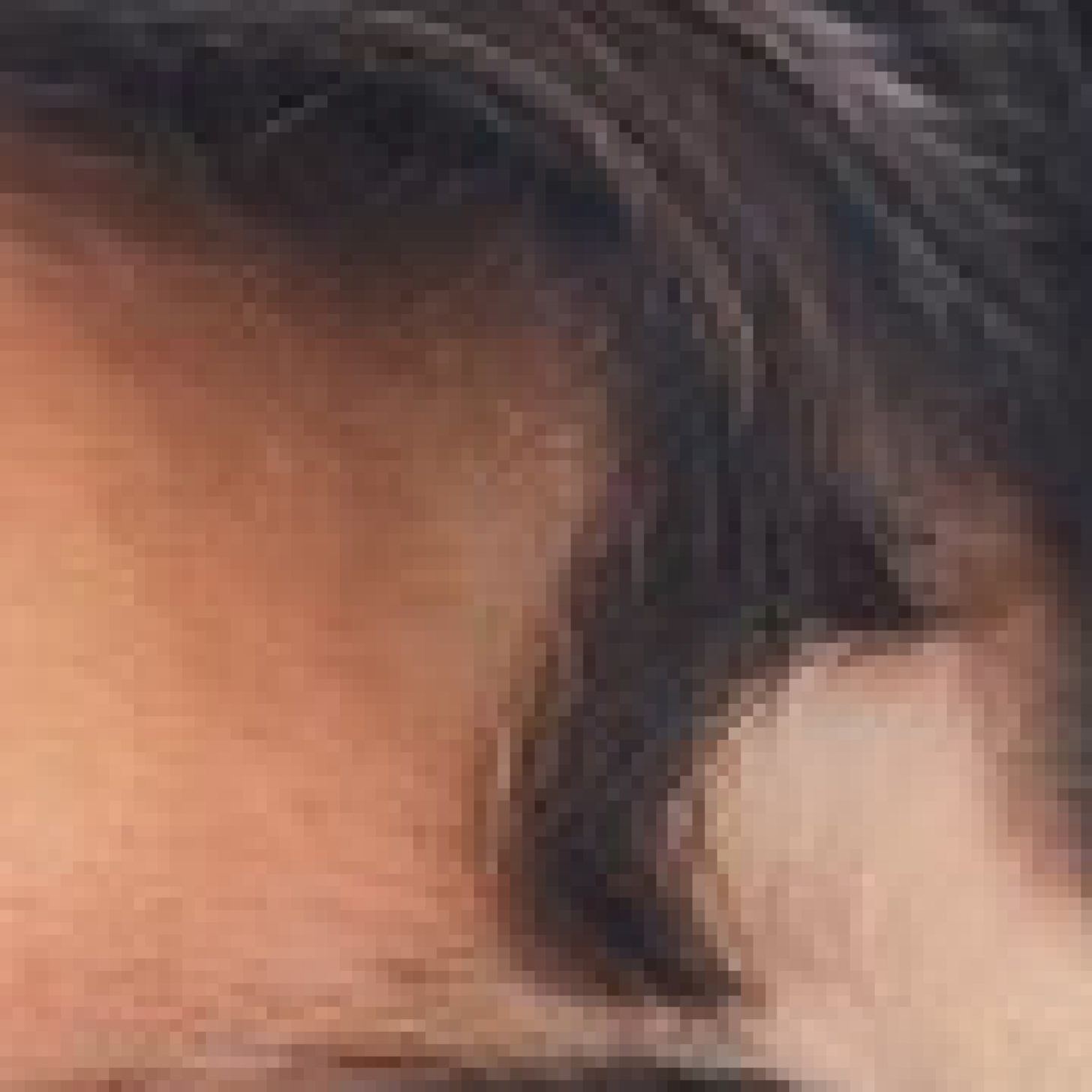}&
    \includegraphics[width=0.19\linewidth]{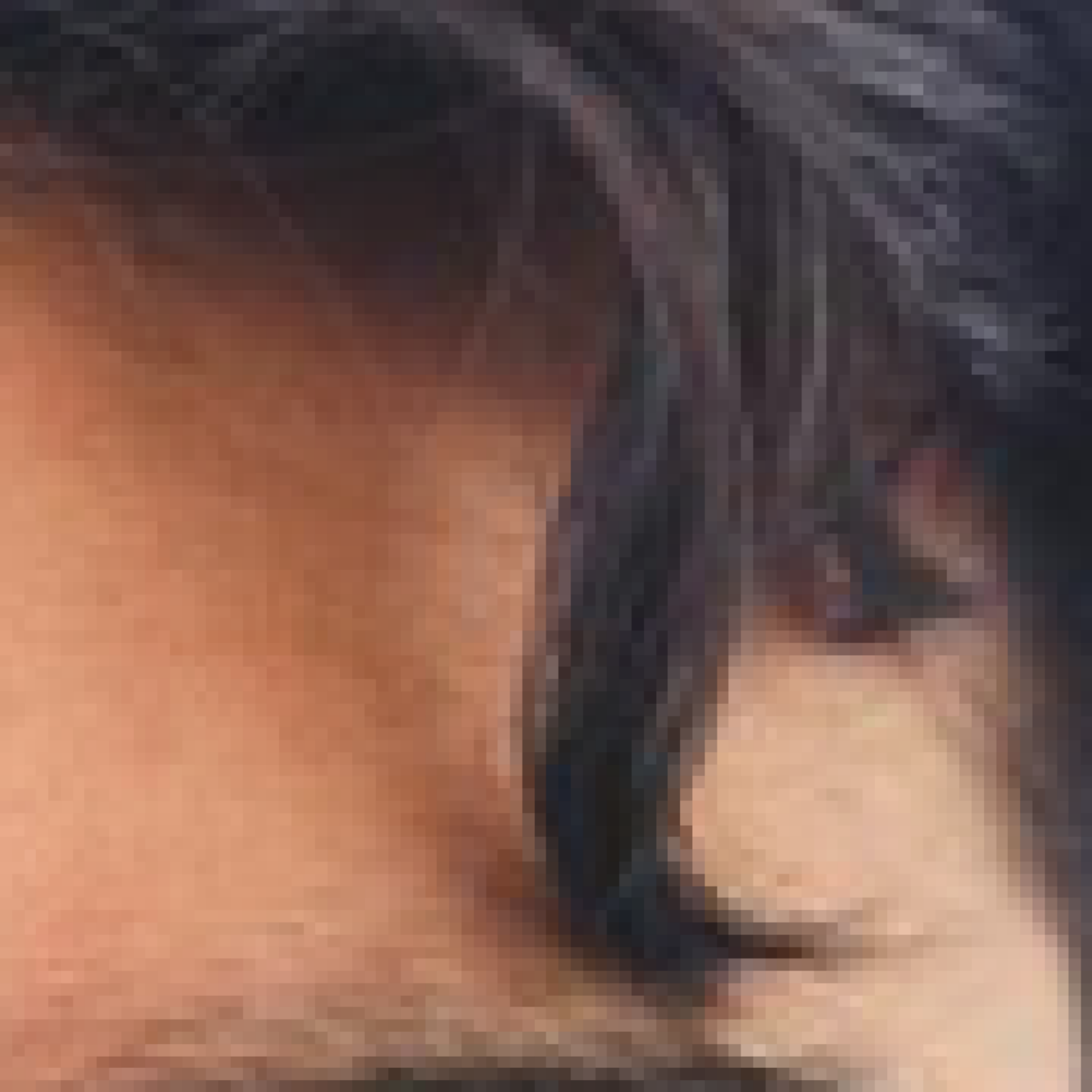}\\
  \end{tabular}
  \caption{ Qualitative comparisons on the in-the-wild images.}
  \label{fig:results-in-the-wild}
\end{figure}

The portraits corrected through a variety of approaches, including \ours, are depicted in Fig.~\ref{fig:cmdp_results}. Warping-based \cite{fried2016perspective} do not seem to have a major effect on inputs. In contrast, 3DP~\cite{shih20203d} introduces noticeable changes, while amplifying distortions, so that the middle part of a face exhibits less distortion, but the head and chin are malformed. Generative TriPlaneNet~\cite{bhattarai2024triplanenet} and DisCo~\cite{wang2023disco} make face look different. HFGI3D~\cite{Xie2022Highfidelity3G} produces recognizable portraits, yet over-smoothed and featuring visual artifacts. Our method generates faces with fewer perspective distortions while maintaining identity (Fig.\ref{fig:results-in-the-wild}), especially in cases when the camera changes notably and the inpainted regions are relatively large.

To identify a necessary and sufficient number of optimization steps, we vary the number of iterations and assess the quality achieved. As can be observed from Fig.~\ref{fig:ablation-iterations}, PSNR peaks after 100 iterations and then decreases slightly as the number of iterations grows. Respectively, we use 200 iterations by default. Note that our baseline DisCo~\cite{wang2023disco} requires approximately 1200 iterations of a comparable complexity.

\begin{table}[h!]
\renewcommand\tabcolsep{3pt}
\centering 
\begin{tabular}{lcccc} \hline
Methods & PSNR$\uparrow$ &
SSIM$\uparrow$ & LPIPS$\downarrow$ & ID$\uparrow$\\
\hline
PTI \cite{roich2022pivotal} & \textbf{20.06} & 0.630 &  \underline{0.138} & 0.614  \\
Ko’s  \cite{ko20223d} & 18.91 & 0.610 & 0.155 & 0.654  \\
HFGI3D \cite{Xie2022Highfidelity3G} & 19.16 & \textbf{0.650} & 0.203 &  \underline{0.780}\\
TriPlaneNet \cite{bhattarai2024triplanenet} & 19.27 & 0.613 & 0.146 & 0.740 \\
\textbf{\ours{}, ours} &  \underline{19.61} &  \underline{0.640} & \textbf{0.131} & \textbf{0.906}   \\ 
\hline
\end{tabular}
\caption{
Quantitative comparison of head pose correction on our HeRo dataset. We report metric averaged across views. 
}
\label{tab:results-rotation}
\end{table}

\subsection{Head Pose correction}
\label{ssec:head-pose-correction}

To assess head pose correction, we transform HeRo photos from a Front camera (reference) into a Left, Right, or Top view and compare them with ground truth photos taken from these views. Quantitative and qualitative results are presented in Table \ref{tab:results-rotation} and Fig.~\ref{fig:results-hero}, respectively. The major advantage of our approach is identity preservation, as reflected in an exceptional ID score. Besides, for small rotation angles, \ours{} successfully handles complicated details as glasses.

\section{CONCLUSION}
\label{sec:conclusion}

In this paper, we introduced \ours{}, a selfie editing method that eliminates face perspective distortion and corrects head pose in a close-up face crop. Our approach enriches 3D warping with the flexibility and expressiveness of a 3D generative model. The resulting image is a blend of a warped image obtained through mesh-based rendering, and another image produced with a 3D GAN. Experiments on face undistortion benchmarks and our novel Head Rotation dataset proved that \ours{} provides more realistic results with finer details and better preserves identity compared to existing techniques, and hence establishes a new state-of-the-art in face undistortion and head pose correction tasks.

\vfill\pagebreak

\bibliographystyle{IEEEbib}
\bibliography{strings,refs}

\begin{thebibliography}{10}

\bibitem{fried2016perspective}
Ohad Fried, Eli Shechtman, Dan~B Goldman, and Adam Finkelstein,
\newblock ``Perspective-aware manipulation of portrait photos,''
\newblock {\em ACM Transactions on Graphics (TOG)}, vol. 35, no. 4, pp. 1--10, 2016.

\bibitem{Zhao2019LearningPU}
Yajie Zhao, Zeng Huang, Tianye Li, Weikai Chen, Chloe LeGendre, Xinglei Ren, Jun Xing, Ari Shapiro, and Hao Li,
\newblock ``Learning perspective undistortion of portraits,''
\newblock {\em International Conference on Computer Vision (ICCV)}, pp. 7848--7858, 2019.

\bibitem{roich2022pivotal}
Daniel Roich, Ron Mokady, Amit~H Bermano, and Daniel Cohen-Or,
\newblock ``Pivotal tuning for latent-based editing of real images,''
\newblock {\em ACM Transactions on Graphics (TOG)}, vol. 42, no. 1, pp. 1--13, 2022.

\bibitem{Chan2021EfficientG3}
Eric Chan, Connor~Z. Lin, Matthew Chan, Koki Nagano, Boxiao Pan, Shalini~De Mello, Orazio Gallo, Leonidas~J. Guibas, Jonathan Tremblay, and et~al.,
\newblock ``Efficient geometry-aware 3d generative adversarial networks,''
\newblock {\em Conference on Computer Vision and Pattern Recognition (CVPR)}, 2021.

\bibitem{ide3d}
Jingxiang Sun, Xuan Wang, Yichun Shi, Lizhen Wang, Jue Wang, and Yebin Liu,
\newblock ``Ide-3d: Interactive disentangled editing for high-resolution 3d-aware portrait synthesis,''
\newblock {\em ACM Transactions on Graphics (TOG)}, vol. 41, no. 6, nov 2022.

\bibitem{yin20223d}
Fei Yin, Yong Zhang, Xuan Wang, Tengfei Wang, Xiaoyu Li, Yuan Gong, Yanbo Fan, Xiaodong Cun, Ying Shan, Cengiz Oztireli, and Yujiu Yang,
\newblock ``3d gan inversion with facial symmetry prior,''
\newblock {\em arXiv preprint arXiv:2211.16927}, 2022.

\bibitem{ko20223d}
Jaehoon Ko, Kyusun Cho, Daewon Choi, Kwang seok Ryoo, and Seung~Wook Kim,
\newblock ``3d gan inversion with pose optimization,''
\newblock {\em Winter Conference on Applications of Computer Vision (WACV)}, 2022.

\bibitem{yuan2023make}
Ziyang Yuan, Yiming Zhu, Yu~Li, Hongyu Liu, and Chun Yuan,
\newblock ``Make encoder great again in 3d gan inversion through geometry and occlusion-aware encoding,''
\newblock {\em International Conference on Computer Vision (ICCV)}, pp. 2437--2447, 2023.

\bibitem{Xie2022Highfidelity3G}
Jiaxin Xie, Hao Ouyang, Jingtan Piao, Chenyang Lei, and Qifeng Chen,
\newblock ``High-fidelity 3d gan inversion by pseudo-multi-view optimization,''
\newblock {\em Conference on Computer Vision and Pattern Recognition (CVPR)}, 2022.

\bibitem{shih20203d}
Meng-Li Shih, Shih-Yang Su, Johannes Kopf, and Jia-Bin Huang,
\newblock ``3d photography using context-aware layered depth inpainting,''
\newblock in {\em Conference on Computer Vision and Pattern Recognition (CVPR)}, 2020.

\bibitem{wang2023disco}
Zhixiang Wang, Yu-Lun Liu, Jia-Bin Huang, Shin'ichi Satoh, Sizhuo Ma, Gurunandan Krishnan, and Jian Wang,
\newblock ``Disco: Portrait distortion correction with perspective-aware 3d gans,''
\newblock {\em arXiv}, 2023.

\bibitem{bazarevsky2019blazeface}
Valentin Bazarevsky, Yury Kartynnik, Andrey Vakunov, Karthik Raveendran, and Matthias Grundmann,
\newblock ``Blazeface: Sub-millisecond neural face detection on mobile gpus,''
\newblock {\em arXiv}, 2019.

\bibitem{MODNet}
Zhanghan Ke, Jiayu Sun, Kaican Li, Qiong Yan, and Rynson~W.H. Lau,
\newblock ``Modnet: Real-time trimap-free portrait matting via objective decomposition,''
\newblock in {\em AAAI Conference on Artificial Intelligence (AAAI)}, 2022.

\bibitem{Deep3DFaceRecon}
Yu~Deng, Jiaolong Yang, Sicheng Xu, Dong Chen, Yunde Jia, and Xin Tong,
\newblock ``Accurate 3d face reconstruction with weakly-supervised learning: From single image to image set,''
\newblock {\em Conference on Computer Vision and Pattern Recognition Workshops (CVPRW)}, 2019.

\bibitem{bhattarai2024triplanenet}
Ananta~R. Bhattarai, Matthias Nie{\ss}ner, and Artem Sevastopolsky,
\newblock ``Triplanenet: An encoder for eg3d inversion,''
\newblock in {\em Winter Conference on Applications of Computer Vision (WACV)}, 2024.

\bibitem{zhang2018perceptual}
Richard Zhang, Phillip Isola, Alexei~A Efros, Eli Shechtman, and Oliver Wang,
\newblock ``The unreasonable effectiveness of deep features as a perceptual metric,''
\newblock in {\em Conference on Computer Vision and Pattern Recognition (CVPR)}, 2018.

\bibitem{mediapipe}
Camillo Lugaresi, Jiuqiang Tang, Hadon Nash, Chris McClanahan, Esha Uboweja, Michael Hays, Fan Zhang, Chuo-Ling Chang, Ming Yong, Juhyun Lee, Wan-Teh Chang, Wei Hua, Manfred Georg, and Matthias Grundmann,
\newblock ``Mediapipe: A framework for perceiving and processing reality,''
\newblock in {\em Conference on Computer Vision and Pattern Recognition Workshops (CVPRW)}, 2019.

\bibitem{Ansari_2019}
Sameer Ansari, Neal Wadhwa, Rahul Garg, and Jiawen Chen,
\newblock ``Wireless software synchronization of multiple distributed cameras,''
\newblock in {\em International Conference on Computational Photography (ICCP)}, 2019.

\bibitem{LPIPS}
Richard Zhang, Phillip Isola, Alexei~A. Efros, Eli Shechtman, and Oliver Wang,
\newblock ``The unreasonable effectiveness of deep features as a perceptual metric,''
\newblock {\em Conference on Computer Vision and Pattern Recognition (CVPR)}, 2018.

\bibitem{Deng2018ArcFaceAA}
Jiankang Deng, J.~Guo, J.~Yang, Niannan Xue, Irene Kotsia, and Stefanos Zafeiriou,
\newblock ``Arcface: Additive angular margin loss for deep face recognition,''
\newblock {\em Transactions on Pattern Analysis and Machine Intelligence (TPAMI)}, vol. 44, pp. 5962--5979, 2018.

\end{thebibliography}

\end{document}